\pdfoutput=1

\documentclass[10pt,letterpaper]{article}
\usepackage[top=0.85in,left=2.75in,footskip=0.75in]{geometry}

\usepackage{amsmath,amssymb}

\usepackage{changepage}

\usepackage[utf8x]{inputenc}

\usepackage{textcomp,marvosym}

\usepackage{cite}

\usepackage{nameref,hyperref}

\usepackage[right]{lineno}

\usepackage{microtype}
\DisableLigatures[f]{encoding = *, family = * }

\usepackage[table]{xcolor}

\usepackage{array}

\newcolumntype{+}{!{\vrule width 2pt}}

\newlength\savedwidth

\raggedright
\usepackage[aboveskip=1pt,labelfont=bf,labelsep=period,justification=raggedright,singlelinecheck=off]{caption}

\makeatletter
\renewcommand{\@biblabel}[1]{\quad#1.}
\makeatother

\date{}

\usepackage{lastpage,fancyhdr,graphicx}
\usepackage{epstopdf}
\fancyheadoffset[L]{2.25in}
\fancyfootoffset[L]{2.25in}
\usepackage[normalem]{ulem}
\usepackage{changepage}
\usepackage{adjustbox}
\usepackage[table]{xcolor}

\usepackage{color}
\usepackage{soul}
\usepackage{booktabs}
\usepackage{mathtools}
\usepackage{array}

\definecolor{Rred}{RGB}{178, 24, 43}
\definecolor{Rorange}{RGB}{214, 96, 77}
\definecolor{Rlrdorange}{RGB}{244, 165, 130}
\definecolor{Rltorange}{RGB}{253, 219, 199}
\definecolor{Reggshell}{RGB}{247, 247, 247}
\definecolor{RCHIsky}{RGB}{209, 229, 240}
\definecolor{RCHIltblue}{RGB}{146, 197, 222}
\definecolor{RCHImedblue}{RGB}{67, 147, 195}
\definecolor{RCHIblue}{RGB}{33,102, 172}

\usepackage{changepage}

\begin{document}
\vspace*{0.2in}

\begin{flushleft}
{\Large
\textbf\newline{
Predicting Acute Kidney Injury at Hospital Re-entry Using High-dimensional Electronic Health Record Data
}
}
\newline
\\
Samuel J. Weisenthal\textsuperscript{1,2},
Caroline Quill\textsuperscript{1,2,4},
Samir Farooq\textsuperscript{1},
Henry Kautz\textsuperscript{5,6},
Martin S. Zand\textsuperscript{1,2,3*}
\\
\bigskip
\textbf{1} Rochester Center for Health Informatics, University of Rochester Medical Center, Rochester, NY, USA
\\
\textbf{2} Clinical Translational Science Institute, University of Rochester Medical Center, Rochester, NY, USA
\\
\textbf{3} Department of Medicine, Division of Nephrology, University of Rochester Medical Center, Rochester, NY, USA
\\
\textbf{4} Department of Medicine, Division of Pulmonary and Critical Care Medicine, University of Rochester Medical Center, Rochester, NY, USA
\\
\textbf{5} Department of Computer Science, University of Rochester, Rochester, NY, USA
\\
\textbf{6} Goergen Institute for Data Science, University of Rochester, Rochester, NY, USA

\bigskip

%
%





* martin\textunderscore zand@urmc.rochester.edu

\end{flushleft}


\section*{Abstract}

Acute Kidney Injury (AKI), a sudden decline in kidney function, is associated with increased mortality, morbidity, length of stay, and hospital cost.  Since AKI is sometimes preventable, there is great interest in prediction.  Most existing studies consider all patients and therefore restrict to features available in the first hours of hospitalization.  Here, the focus is instead on rehospitalized patients, a cohort in which rich longitudinal features from prior hospitalizations can be analyzed.  Our objective is to provide a risk score directly at hospital re-entry.  Gradient boosting, penalized logistic regression (with and without stability selection), and a recurrent neural network are trained on two years of adult inpatient EHR data (3,387 attributes for 34,505 patients who generated 90,013 training samples with 5,618 cases and 84,395 controls).  Predictions are internally evaluated with 50 iterations of 5-fold grouped cross-validation with special emphasis on calibration, an analysis of which is performed at the patient as well as hospitalization level.  Error is assessed with respect to diagnosis, race, age, gender, AKI identification method, and hospital utilization.  In an additional experiment, the regularization penalty is severely increased to induce parsimony and interpretability.  Predictors identified for rehospitalized patients are also reported with a special analysis of medications that might be modifiable risk factors.  Insights from this study might be used to construct a predictive tool for AKI in rehospitalized patients.  An accurate estimate of AKI risk at hospital entry might serve as a prior for an admitting provider or another predictive algorithm.


\section*{Introduction}
Acute kidney injury (AKI) is a sudden decline in kidney function over days, which may be temporary or permanent~\cite{RIFLE2004, Venkataraman2007}.  
AKI is common in hospitalized patients, with an estimated incidence of 13\% 
and, importantly, 
is associated with greatly increased morbidity (e.g., long-term dialysis), mortality, length of stay, and hospital cost~\cite{chertow2005MortLOSCost}. 
Diagnosis of AKI is challenging, as patients are generally asymptomatic and commonly used biomarkers change over a period of days following injury~\cite{bellomo2012acute}. 
Causes of AKI are generally grouped into decreased renal blood flow (e.g., hypotension due to sepsis or heart failure), direct renal toxicity (e.g., due to medications, radiocontrast dye, or  bacterial toxins), and urinary outflow obstruction (e.g., bladder outlet obstruction or kidney stones).

Defining AKI for research purposes, or to assess clinical outcomes, is also challenging.  A variety of definitions exist, primarily based on changes in the concentration of serum creatinine (sCr).  Creatinine is a protein made by muscle and excreted by the kidneys via glomerular filtration.  Serum Cr is inversely proportional to the glomerular filtration rate (GFR), a true indicator of renal function that is not easily measured.  Doubling of sCr at steady state reflects a 50\% decrease in renal function.  Consensus definitions for AKI rely heavily on changes in sCr over time, and include the 2004 RIFLE criteria~\cite{RIFLE2004} (modified in 2007 by the Acute Kidney Injury Network (AKIN)~\cite{AKIN}) and the 2012 Kidney Disease: Improving Global Outcomes (KDIGO)~\cite{KDIGO-AKIdef2012, ERBPpositionState} definitions.  The KDIGO AKI definition, which we use here, combines the RIFLE ``Risk'' definition with the AKIN criterion for absolute increase in sCr.   

Developing a broadly applicable and accurate risk index for AKI in rehospitalized patients could have a major impact on hospital care, particularly if it were practical enough to allow preventive intervention or more intense monitoring from the time of hospital admission~\cite{sutherland2016utilizingAKIADQI}.  
With early risk identification, a variety of preventive strategies can be implemented~\cite{lameire2008prevention}.  For example, AKI from radiocontrast dye, chemotherapy, or aminoglycoside antibiotics can be prevented by altering treatment, administration of fluids, alternate imaging modalities or close monitoring~\cite{TepelPreventContrastRedRenalFunc,SolomonRadioCon,GuoDrugNephro,VogelzangChemoNeph}. 
Given that such interventions can mitigate severity, AKI prediction is an area of active research, with recent emphasis on Electronic Health Record (EHR) data~\cite{koyner2016development,sutherland2016utilizingAKIADQI,calDavis,CroninVetStrat}.  
Existing studies generally focus on AKI in the context of cardiac procedures~\cite{c8,c12,c15,c18}, critical illness~\cite{i1,i2,i3,i4,i5,i6}, the elderly~\cite{Kate2016}, transplants of the liver~\cite{livt2} and lung~\cite{lungT1}, and extensive muscle injury (rhabdomyolysis)~\cite{R1,R2}.  Recently, we see predictive systems~\cite{CroninVetStrat,Kate2016,calDavis} that exploit numerous features from the EHR rather than a small number of manually picked variables. 
In existing studies, predictions of AKI risk are made for \textit{all} hospitalized patients, many of whom do not have previous hospitalizations.  They are hence restricted to features from the current hospitalization, even when a patient has more extensive information in the EHR.  

To our knowledge, there are no published studies focused explicitly and exclusively on a large cohort of \textit{rehospitalized} patients.  Focus on this group allows analysis of longitudinal information from prior hospitalizations (e.g., the number of previous episodes of AKI, the number of abnormal urea nitrogen (UN) readings, or the number of loop diuretics administered).  Although the subset of rehospitalized patients is a specific cohort, such an analysis is general as it pertains to \textit{all} rehospitalized patients.  Since rehospitalized patients have not been studied explicitly in the literature, all available features from all available time points were analyzed.

In this framework, prior hospitalizations might be considered surrogate ``renal stress tests,'' reflecting renal resiliency to injurious events.  Conversely, prior hospitalizations might be renal stressors, diminishing renal reserve. Most previous studies on AKI posit data models, although some more recent work~\cite{CroninVetStrat,Kate2016} explores predictive algorithms, distinct from data models~\cite{Breiman_alg_v_mod}, as done here.  Penalized regression and ensemble methods were employed to mitigate overfitting.  In particular, a decision tree ensemble classifier constructed with gradient boosting (GBC), which is highly robust to outliers and well suited to high-dimensional, noisy data~\cite{friedman2001greedy}, was explored along with a recurrent neural network~\cite{hochreiter1997long} (LSTM) for time series analysis, and penalized logistic regression (LR1) for high-dimensional data where it is believed that only a few features are relevant~\cite{tibshirani1996regression}.  In an additional experiment with the latter, the penalty was increased severely to induce parsimony and interpretability.  
New AKI predictors specific to rehospitalized patients were identified; a special analysis of medication-related predictors is presented as they may be of interest as potentially modifiable risk factors.

\section*{Materials and Methods}

\subsection*{Dataset}

The research protocol was approved by the University of Rochester Research Subjects Review Board (RSRB00056930).  Research data were coded such that patients could not be directly identified in compliance with the Department of Health and Human Services Regulations for the Protection of Human Subjects (45 CFR 46.101(b)(4)).  This dataset is a comprehensive 2-year window into the EHR for informatics and population-health studies.  For this work we excluded all hospitalizations with age at admission $<$ 18 years, all hospitalizations following a prior hospitalization in which the ICD-9 code for end stage renal disease (ESRD; 585.9) was assigned.  Hospitalizations following a transplant for ESRD were included.  Patients who had undergone dialysis were included, as dialysis is sometimes performed in the setting of transient AKI, and therefore presence of dialysis does not indicate permanent renal dysfunction.  Multiple hospitalizations were available for roughly 32\% of patients.

The dataset consisted of tables containing administrative, laboratory, and medication data that was queried respectively from separate billing (Flowcast, IDX Systems), eRecord (Epic), and pharmacy databases which could be joined on admit id, which were linked during de-identification.  The administrative dataset included International Classification of Diseases, 9\textsuperscript{th} Revision (ICD-9) diagnosis and procedure codes, Current Procedural Terminology 4th Edition (CPT-4) procedure codes, Diagnosis-Related Groupings (DRG) codes, bed locations during hospitalization, discharge disposition, discharge and admission days, insurance (primary, secondary, and other), marital status, gender, age, race, and total length of stay.  The laboratory dataset included direct bilirubin, point-of-care creatinine, bicarbonate, chloride, calcium, anion gap, phosphate, glomerular filtration rate, sCr, urea nitrogen (UN), albumin, total protein, aspartate and alanine transaminase, hemoglobin, glucose, and glycated hemoglobin.  The pharmacy dataset included, for each medication, description, pharmacologic class and subclass, and therapeutic class. Table~\ref{tab:notation} contains abbreviations.

\subsection*{Definitions} 
\begin{table}[!ht]
\small
\centering
\caption{{\bf Abbreviations and Notation}}
\begin{tabular}{l|l}
\hline
Abbreviation & Description \\ 
\hline
AKI & Acute kidney injury \\
ALR1 & Anscombe LR1\\
AUC & Area under the curve\\
CKD & Chronic kidney disease \\
CLR & Clinical LR\\
CV & Cross validation\\
Dx & Diagnosis \\
EHR & Electronic health record\\
ESRD & End-stage renal disease\\
GBC & Gradient boosting classifier\\
GFR & Glomerular filtration rate\\
RGBC & Recent GBC \\
RHPLR1 & Randomized HPLR1\\
RLR1 & Randomized LR1\\
$H_C$ & Current hospitalization\\
$H_P$ & Prior hospitalizations\\
HP  & Hyperparameter \\
HPLR1 & Highly penalized LR1\\
LR1 & Logistic regression with $l1$-norm penalty\\
LSTM & Long short-term memory\\
MGBC & Medication GBC \\
MLR1 & Medication LR1\\
PPV & Positive predictive value\\
$P_P$ & Predicted probability \\
$\overline{P_P}$ & Mean $P_P$ \\
$P_O$ & Observed probability \\
$\overline{P_O}$ & Mean $P_O$ \\
PR & Precision recall \\
Px & Procedure \\
ROC & Receiver operating characteristic\\
sCr & Serum creatinine\\
STD & Standard deviation\\
UN & Urea nitrogen\\
\hline 
\end{tabular}
\label{tab:notation}
\end{table}

\textit{Hospitalization:} hospitalization was defined as an admission (inpatient) or administrative status ``under observation'' (e.g., in the emergency department, but not admitted to inpatient care).  \textit{AKI:}  AKI was defined as the presence of either an administrative diagnosis code or sCr delta.  Administrative ICD9 codes included 584.5 (AKF with lesion of tubular necrosis), 584.6 (AKF with lesion of renal cortical necrosis), 584.7 (AKF with lesion of renal medullary (papillary) necrosis), 584.8 (AKF with other specified pathological lesion in kidney), or 584.9 (AKF, unspecified).  As diagnosis codes are believed to be specific but not sensitive for AKI~\cite{waikar2006validity}, they were supplemented with sCr for patients with available laboratory values. Using KDIGO guidelines~\cite{KDIGO-AKIdef2012}, diagnosis was made with a 1.5-fold or greater increase in sCr from baseline within 7 days or 0.3 mg/dL or greater increase in sCr within 48 hours.  Baseline sCr for an individual hospital stay was defined as the first documented inpatient sCr, as recommended by~\cite{ERBPpositionState}, and then as a sliding baseline.  All such diagnoses were made within a single hospital stay (e.g., a case in which the rise in sCr occurred over two rapidly successive hospital stays was ignored).  It is possible that some patients who did have AKI were neither assigned a code nor had their sCr measured, and would thus be invisible with respect to AKI diagnosis. 

\subsection*{Preprocessing}
Medication descriptions were stripped of dosage information and treated as categorical variables.  Abnormal lab flags were constructed by combining EHR-generated flags (an automated indication of, e.g., hyperkalemia) with test name.  For features such as diagnoses and procedures, each admission contained a list.  The hospitalization with the highest number, $D$, of diagnoses was identified.  Given any other hospitalization with a list of $D'$ diagnoses, $D-D'$ ``non-diagnoses,'' the number of diagnoses not assigned relative to its peers, were added.  This was done to enhance predictive performance, as missingness patterns were useful for the related task of phenotyping~\cite{lipton2016directly}.  Top-level binary representation~\cite{singhLeveHier} was used for the hierarchical ICD-9, CPT-4, and DRG codes and ``code precision'' is defined as the level at which the tree was accessed.  For example, for chronic kidney disease (CKD), precision 3 produces a single feature, 581, that contains any occurrence of 585.1-6 (CKD, Stages 1-6) or 585.9 (CKD, unspecified), essentially grouping these codes.  Alternatively, precision 4 ungroups the subcodes, producing 7 different features, one for each administrative classification of CKD.  An exploratory grid search for precision of ICD-9, CPT-4, and DRG codes was performed and discrimination found to be relatively insensitive, so code precision was fixed at 3, a level at which different subgroups of AKI and CKD were aggregated. 

All extremely sparse features (with fewer than 100 non-zero or non-missing elements) were removed.  Hence, for continuous values, features that were unobserved frequently or frequently zero (continuous lab and demographic values in this dataset should generally be nonzero) were removed; categorical variables that were rarely observed were removed.  Besides reducing training time, removing rarely-present features, which can be difficult to gather, improves clinical applicability.  This step did not need to be incorporated into the pipeline as it was a form of response-independent dimensionality reduction.  

\subsection*{Feature Extraction}
\begin{figure}[!ht]
\includegraphics[scale=0.7]{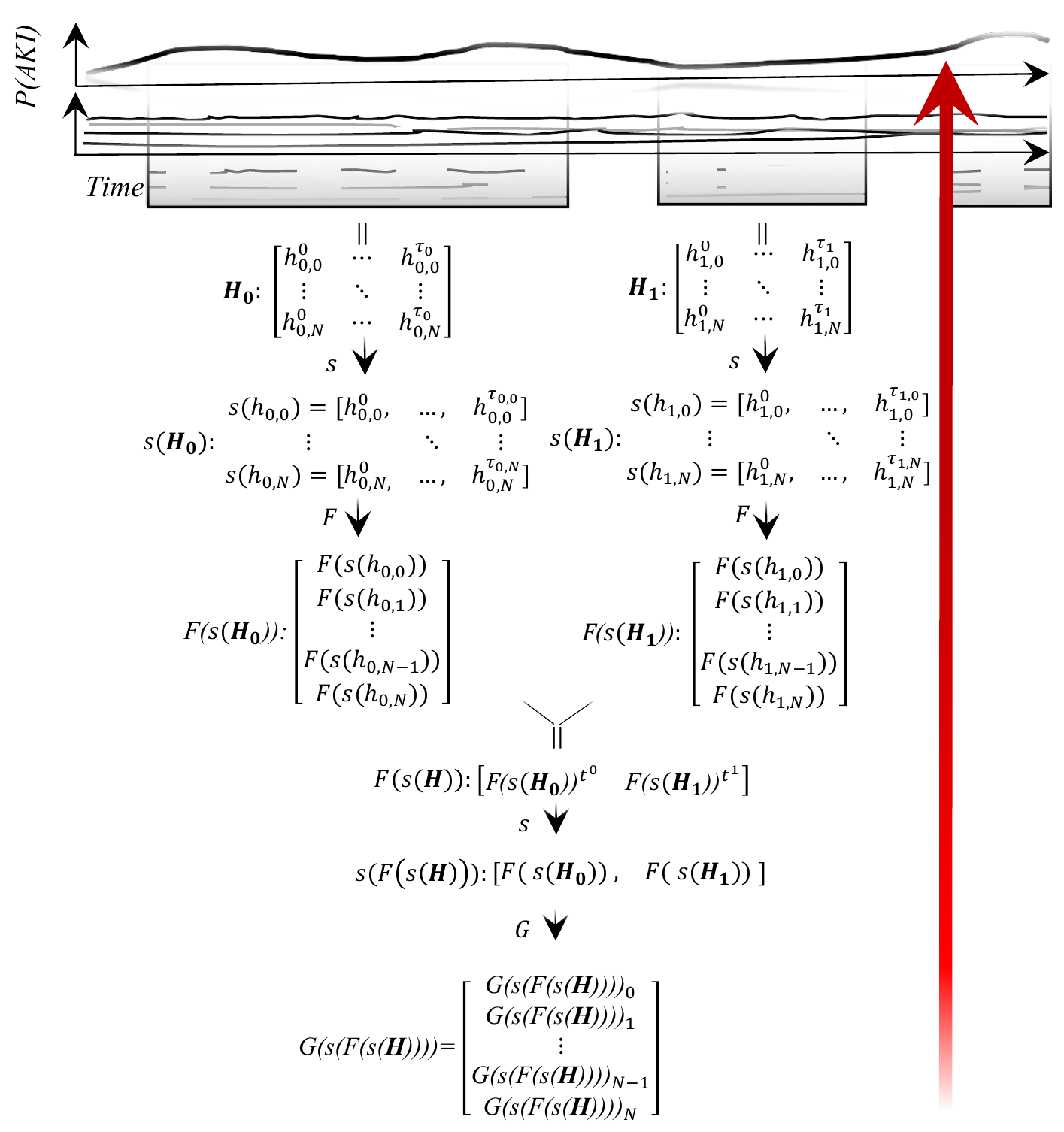}
\caption{{\bf Feature extraction pipeline for estimating $P(AKI)$ during rehospitalization.} We sought to estimate the probability of AKI during rehospitalization given all of a patient's previous hospitalizations, P(AKI$_{H_{C}}|{H_{P}}$), shown by the red arrow. An example of three hospitalizations ($H_1, H_2, H_3$) is shown.  Here $H_1$ and $H_2$ are used to estimate $P(AKI)$ during $H_3$.  The EHR captures raw data (shown in boxes closest to the time series tracings) of which our dataset contains $N+1$ features.  At each level, data is aggregated via domain-expertise-informed functions $F$ and $G$.  The pipeline produces a single, fixed-length representation of all previous hospitalizations to serve as input to a learning algorithm. Measurements from each hospitalization, and series of hospitalizations, are treated as sequences, denoted with operator $s$.}
\label{feat_ext}
\end{figure}

Over time, all patients have some continuous risk, $P(AKI)$. Using data from all previous hospitalizations, $H_{P}$, we hope to estimate the probability of AKI during the current hospitalization, $H_{C}$, at the time of hospital re-entry,  P(AKI$_{H_{C}}|{H_{P}}$).  An example case illustrating the feature extraction procedure used throughout this study is diagrammed in Fig~\ref{feat_ext}.  It was designed to compress longitudinal, irregular and misaligned observations into a fixed-length representation.  Summary statistics are used for repeated measures since they are interpretable and shown by~\cite{goldstein2017comparison} to be effective for some risk prediction problems.  Note that a patient with $n$ hospitalizations generates $n-1$ training samples.

EHR data is an irregularly sampled (e.g., sCr is \textit{not} measured at an hourly frequency), misaligned (e.g., sCr and hemoglobin are \textit{not} consistently simultaneously sampled) window into a patient's renal health.  The $j^{th}$ hospitalization can be conceptualized as a matrix $\boldsymbol{H_{j}}$ where each row is one of $N+1$ features and each column corresponds to some time step  $\leq$ $\tau_{j}+1$, the end time of hospitalization $j$.  Since observations are irregularly sampled and misaligned, it is convenient to transform the time-indexed hospitalization matrix into a collection of $N+1$ sequences $s(\bf{H})$, where $s$ is a function that converts a time series to a sequence.  The sequence $s(h_{i,j})$ corresponds to feature $j$ of hospitalization $i$ and has its own number of entries, $\tau_{i,j}+1$, corresponding to the number of times feature $i$ was recorded during hospitalization $j$.  Such sequences are useful because they can be summarized or transformed via some function $F$ without \textit{explicit} imputation, albeit with information loss.  

Summary functions can encode relevant characteristics of the generative process; e.g., a sum provides a sense of the number of tests along with some information about the results of the tests.  A higher number of sCr tests may reflect a heightened concern for, or closer monitoring of, AKI and its metabolic consequences.  $F$ takes as input a sequence and outputs the minimum, maximum, mean, variance, and sum for continuous variables and sum for categorical variables.  $F(s({\boldsymbol{H}}_{i}))$ is now a $M$x1 fixed-length representation of the $N+1$ sequences in the $i^{th}$ hospitalization, where we have increased the dimension $M > (N+1)$ by concatenating any vector outputs of F.  

Let $\textbf{H}$ refer to all hospitalizations before the rehospitalization for which $P(AKI)$ is to be estimated (e.g., the third in Fig~\ref{feat_ext}).  As with the sequences of laboratory measurements, $F(s({\boldsymbol{H}}))$ is represented as a matrix whose entries are indexed by time of admission. However, although the observations are now aligned, they are still irregularly sampled.  $F(s(\boldsymbol{H}))$ can then again be converted to a sequence $s(F(s(\boldsymbol{H})))$.  Finally, the sequence of hospitalizations $s(F(s(\boldsymbol{H})))$ can be summarized, or aggregated, using $G$ to yield $G(s(F(s(\boldsymbol{H}))))$, a $P$-dimensional vector, where any vector outputs of $G$ have been concatenated as with the hospitalization representations, so $P>M$.  The full set of hospitalization-level aggregation functions for G is:
\textit{Administrative:} Max (age), First (race), 
		Last (marital status, gender, insurance), 
		Sum (DRG, discharge disposition, length of stay, locations visited, diagnoses, CPT4/ICD9 procedure);
\textit{Medications:}
Sum (administration of medication by description, class, and subclass); 
\textit{Labs:}
Minimum, Mean, Maximum, Sum, and Variance (labs and abnormal lab flags).  
By aggregating over hospitalizations with $G$, most (not those that were ``first'' or ``last'') categorical variables are rendered continuous, so standard rather than minimum-maximum scaling is used where necessary for regression.

There are benefits and drawbacks of this aggregation-based approach for time series data. Benefits include easy determination of features since $F$ and $G$ (chosen by the analyst) are known.  The \textit{sum} of sCr from prior hospitalizations is easy to understand; a more complicated function learned from the data in a time-dependent algorithm might not be so.  A mirroring limitation is that $F$ and $G$, invented by humans, are probably not optimal for the task at hand (e.g., the optimal hospital aggregator is probably a function with some weight decay over time, allowing distant events to be ``forgotten'').  Another drawback is loss of information on time between events (e.g., very frequent testing might be informative) or recency of events (e.g., a very distant nephrotoxic medication might be less important than a very recent one).  We therefore implemented a recurrent neural network as well, but this work would likely benefit from further exploration of clinical time series methods, an active area of research~\cite{choiHF,lipton2016directly,colopy2017bayesian,alaa2016hidden}. 

\subsection*{Training}
  
The algorithms used have hyperparameters (HP) (e.g., the number of estimators in GBC or scale of the Laplace distribution in LR1) that must be set in addition to the parameters.  For learning algorithms with HP, ``nested'' cross-validation (CV) is recommended~\cite{cawley2010over, varma2006bias} to provide an un-optimistic performance estimate.  Pure nested CV requires that both choosing the HP search space and conducting the HP search be executed independently and identically within every fold.  This is computationally expensive because it allows for high complexity HP (e.g., GBC with a large number of estimators or LR1 with a very small penalty) that lead to overfitting and slow training.  HP were therefore fixed at values found in preliminary experiments (manually or with grid or random search~\cite{bergstra2012random}) to not overfit the data (as determined by a validation set distinct from the test set) and to produce reasonable features per domain expertise.  It was also confirmed that performance on the test set of the fold used to determine HP did not differ substantially from performance on the test sets of the other four folds.  

In greater detail, our HP selection method was as follows: create splits for 5-fold CV. Hence, we have folds 1-5, which consist of Train1, Train2, ..., Train5, and Test1, Test2, ..., Test5.  To set HP, take Train1 and split it into a (sub)train set Train1Train and validation set Train1Val.  Fit different $HP$ on Train1Train and see which HP makes performance (per) for system (sys) per(sys$_{HP}$(Train1Train)) $\approx$ per(sys$_{HP}$(Train1Val)).  Note that we do not try a large number of different HP and select the one with best performance on the validation set; we select the HP for which training and validation errors are most similar.  Also, we examine importances/coefficients from sys$_{HP}$(Train1Train) to ensure that they are reasonably related to renal function. If not, increase regularization.  With this process, choose $HP$, which were found by analyzing Train1, so call them $HPTrain1$.   Fix $HPTrain1$.  Evaluate sys$_{HPTrain1}$(Train1) on Test1, where Train1Train+Train1Val = Train1. This is a pure estimate of generalization performance.  Now, keeping $HPTrain1$ fixed, evaluate sys$_{HPTrain1}$(Train2) on Test2, evaluate sys$_{HPTrain1}$(Train3) on Test3, and so on.  Note that there is necessarily overlap between Train1 and Test2, ..., Test5.  Hence, there is potential leakage from $HPTrain1$ into the performance estimates of Test2, ..., Test5 (but again not into Test1). During training, we therefore checked that performance on Test1 was roughly similar to performance on Test2, ..., Test5.   

We name this process ``pseudo''-nested CV because HP selection was not performed independently in each fold as is required for pure nested CV.  In pure nested CV, we would have specified a search region for HP and allowed HP to be selected in every fold, selecting $HPTrain1$ to be tested on Test1, $HPTrain2$ to be tested on Test2, and so on.  Knowing that choosing our HP manually using the data put us in danger of overfitting, we purposely tried to choose $HPTrain1$ that would yield systems with lower capacity.  Also, note that manual choice of HP precludes comparison of algorithms because HP choice is a confounder; our comparison is therefore over trained systems, not training algorithms.     

Fixed HP for GBC included maximum depth = 2, minimum samples per split = 150, and minimum samples per leaf = 100 and for LR1 C = 2 x $10^{-3}$.  To produce a parsimonious, highly penalized LR1 (HPLR1), C was decreased to 2 x $10^{-4}$ (aiming for $\approx{12}$ features).  For LR1, classes were weighted according to prevalence. 
Between GBC and LR1, choice of learning algorithm was also an HP, but was wrapped into the inner folds of the nested CV as a grid search.  In preliminary experiments, LR1, Ridge~\cite{hoerl1970ridge}, random forest~\cite{Breiman_RF}, multilayer perceptron~\cite{rumelhart1985learning}, and GBC were explored manually.  Ultimately, LR1 and GBC were chosen as candidates for the search since LR1 was close enough to ridge (the problem was expected to be sparse) and GBC close enough to random forest.  As recommended in~\cite{harrell2015regression}, log loss, rather than a binary metric, was optimized in the searches.  

Although a search was performed over learning algorithms, there was no intention of comparing them outright (there are many confounders, e.g., HP choice).  Rather, they were intended for use in concert since both have benefits and drawbacks.  A major difference is that LR1 is linear in its parameters and therefore quite interpretable while GBC is nonlinear and sometimes gives better off-the-shelf predictions (LR1 could be enhanced with basis functions to rival, but this was not done here).  Besides manual setting of HP, all other steps were performed in a pipeline within each fold.  Pipelines were constructed to successively impute (using the most frequent value), scale (using standard scaling; only for LR1), fit, and calibrate (using Platt's scaling~\cite{Platt}).  Training data were split such that, in each fold, 75\% of the observations were used to fit and select classifiers, and the remaining 25\% were held out and used to calibrate the estimator with the lowest log loss.  For HPLR1, there was no search over GBC. 

In an additional experiment, we implemented a variance stabilizing Anscombe transform for LR1 (ALR1) for the count and categorical variables. GBC seemed to be unaffected by this transform because it is a tree-based system invariant to monotone transformations of the input.  Since $l1$-norm penalty is known to select one and discard $x-1$ of $x$ highly correlated features, for the purposes of reporting features, Both LR1 and HPLR1 were rerun with stability selection~\cite{stabselect} (these were named RLR1 and RHPLR1, respectively), which is less likely to discard the remaining $x-1$ features.  In this case, the penalty weight, $C_2$, on the final classifier was roughly nonexistent (vanilla logistic regression) because the feature selection step with penalty weight $C_1$ regularized.  RLR1 randomized selection had $C_1=0.5$ and $C_2=1$; RHPLR1 had $C_1=0.2$ and $C_2=1$. For both RLR1 and RHPLR1, the stability selection sampling fraction was 0.75 with 50 resamples.  

To explore alternative strategies for repeated measures, the first-described experiment was redone exactly, but repeated samples were weighted such that each patient received equal total representation (e.g., a patient with 3 samples was weighted by 1/3; a patient with 2 by 1/2), producing weighted GBC (WGBC), weighted LR1 (WLR1), and weighted HPLR1 (WHPLR1);  alternatively, one sample per patient was randomly selected to produce independent data, producing sampled GBC (SGBC), sampled LR1 (SLR1), and sampled HPLR1 (SHPLR1). Also, for repeated measures, we implemented a recurrent neural network with long short-term memory (LSTM) cells~\cite{hochreiter1997long} that processed the two most recent hospitalizations in sequence.  This recurrent system obviated the need for the hospital aggregator $\textit{G}$. We set the number of hidden layers and units \textit{a priori} and searched over levels of dropout.  Thus in contrast to GBC and LR, we did not set HP for LSTM by using features in one fold of cross validation, and therefore the LSTM was trained with pure nested cross validation. For insertion of LSTM into a pipeline, the Scikit-learn scaler and imputer were decorated to process tensors.

To explore the effect of previous hospitalizations, the first-described experiment was redone exactly, but using only the most recent hospitalization as input (results are reported for ``recent'' GBC: RGBC).  The original decision to include data from all previous hospitalizations was based on the premise that it is better to provide more rather than less information to a learning algorithm (although this requires an extra step of aggregation over hospitalizations).  As medications are potentially modifiable risk factors, GBC and LR1 were also refit exactly as in the first-described experiment, but with only medications as features (MGBC and MLR1). We also implemented a system, CGBC, with only a handful of clinically known risk factors from~\cite{LeblancRiskFact}: age, underlying renal insufficiency (prior AKI or CKD), diabetes, and heart failure.  Logistic regression was used
by setting C=1000 with ridge regression (in order to remain in the scikit-learn ecosystem where all logistic regressions are penalized).  We also randomly permuted the response variable and refit exactly as in the first-described experiment to produce noise GBC (NGBC).

\subsection*{Assumptions}

It is assumed that the majority of patients in the dataset who have an episode of AKI are, by medical history, high risk for AKI.  Conversely, we assume that the majority of patients without AKI have past medical histories that are low risk for AKI.  This is paradoxically a strong assumption.  To see why, consider a patient with high risk for AKI.  We hope to associate this patient's prior hospitalizations with high risk.  Upon rehospitalization, however, suppose that an admitting provider, evaluating the risk as high, decides to administer extra fluids.  Ultimately, and fortunately for the patient, this effort may prevent AKI.  However, the training set now contains a high risk history coupled to a hospitalization in which \textit{AKI did not occur}.  Hence, this patient's high-risk history will incorrectly be associated with a flipped label of non-AKI.  Conversely, a patient with low AKI risk might receive a medication with the potential for causing AKI, resulting in a similar mismatch.  It is therefore assumed that the modifications of disease course just described contribute negligible bias to our predictions, but recognized that this bias \textit{is not detectable via internal or external validation}.  If this assumption is false, it would invalidate our approach, and future work will focus on developing methods to test this assumption.  Notably, this assumption has been shown to fail in a previous study on pneumonia where patients with risk-increasing asthma were given systematic, preferential treatment, effectively flipping their labels~\cite{caruana2015intelligible}.  Bias resulting from interventions could be removed by incorporating events that occur during rehospitalization as predictors.  However, this is precluded because an intervention could occur all the way up to AKI (e.g., a provider might discontinue intravenous fluids and increase the risk of AKI).  Many of our labels are diagnosis codes assigned at the end of the hospitalization, so we do not know when AKI occurred.  With the interpretable HPLR1, it is at least possible to confirm that the features are reasonable and appear not to be subject to this bias.

It is also assumed that a time-based (2-year) sample approximates an ideal patient-based sample.  Repeating training on a patient-based sample would be a useful complement to this study, and if implemented in the EHR should be formulated as such, since a patient may have a previous hospitalization or rehospitalization outside of the sample.  Similarly, it is assumed that our dataset sampled from only one hospital network is representative enough for learning local patterns.  We strongly recommend retraining if the model is to be used outside of the population that generated the training data.  Finally, it is assumed that undetected AKI from lack of sCr measuremenst or no assignment of a diagnosis code is a rare event.

\subsection*{Evaluation}

For evaluation, 50 iterations of nested (except HP determination, as described above) 5-fold CV were performed.  Since any two hospitalizations from the same patient were correlated, CV sampling was ``grouped'' at the patient level.  Micro (over all 250 outer folds) and macro (over 50 iterations) mean and standard deviation of all metrics are reported.  As recommended in~\cite{harrell2015regression}, a probability estimate rather than binary output is provided so the final decision can be made with maximal information at the point of care (e.g., if one patient has 0.499 risk and another has 0.501, these should not be converted to 0 and 1 by an algorithm, but by a provider in clinical context).  Although calibration is primarily assessed, discrimination is also described, as is standard practice, with receiver operating characteristic (ROC) and precision-recall (PR) curves and corresponding areas.  For calibration, curves are shown with Brier score.  \textit{Every} calibration curve shown contains 10 bins.  In addition to hospitalization-level performance of GBC, patient-level performance is also analyzed.  This is conveyed via scatter plots of the average risk per patient (e.g., a patient with two hospitalizations, one of which had AKI and the other of which did not has 0.5 observed risk) by the average predicted risk.  Calibration curves are superimposed for the cases that had 0 or 1 observed risk (all of the hospitalizations and a subset of the patients).

Since algorithms have potential to harm certain subgroups, algorithmic fairness is an active area of research~\cite{KCraw,corbett2017algorithmic}.  Here, an error analysis is performed with special focus on the black box GBC, to detect subgroups for which this might be the case.  After stratifying by outcome, the same iterated, semi-nested CV procedure described above was used to fit an $l1$-penalized linear regression with either diagnosis, race, gender, or age alone as features and the absolute magnitude of the error as the response (minimum 0, maximum 1).  To analyze error by utilization, patients were binned based on the number of hospitalizations that they generated and average error was plotted for each bin.   The relationship between number of hospitalizations from a patient and that patient's impact on coefficients was assessed by removing all hospitalizations from each patient and fitting HPLR1 and then comparing to the coefficients of HPLR1 fit on the full dataset.  The comparison was made using $l1$ norm because the coefficient vectors were low dimensional for HPLR1. Error was also assessed as it related to method of diagnosis (code or sCr) and variance of predicted risk.

\subsection*{Computing environment}

All computational work was performed in Python 2.7.14.  Libraries in scikit-learn~\cite{pedregosa2011scikit} and the SciPy ecosystem~\cite{scipy,mckinney2010pandas,perez2007ipython,Hunter:2007,behnel2010cython,walt2011numpy} were used throughout.  Code was run on a linux-based cluster.  Each experiment was run via an sbatch script requesting roughly 1 node and 100 to 200 GB of random-access memory. All iterations were distributed using job arrays. Code will be made available upon publication on \url{github.com}.
\section*{Results}

\subsection*{AKI Cohort Selection}
\begin{figure}[!ht]
\centering
\includegraphics[scale=0.30]{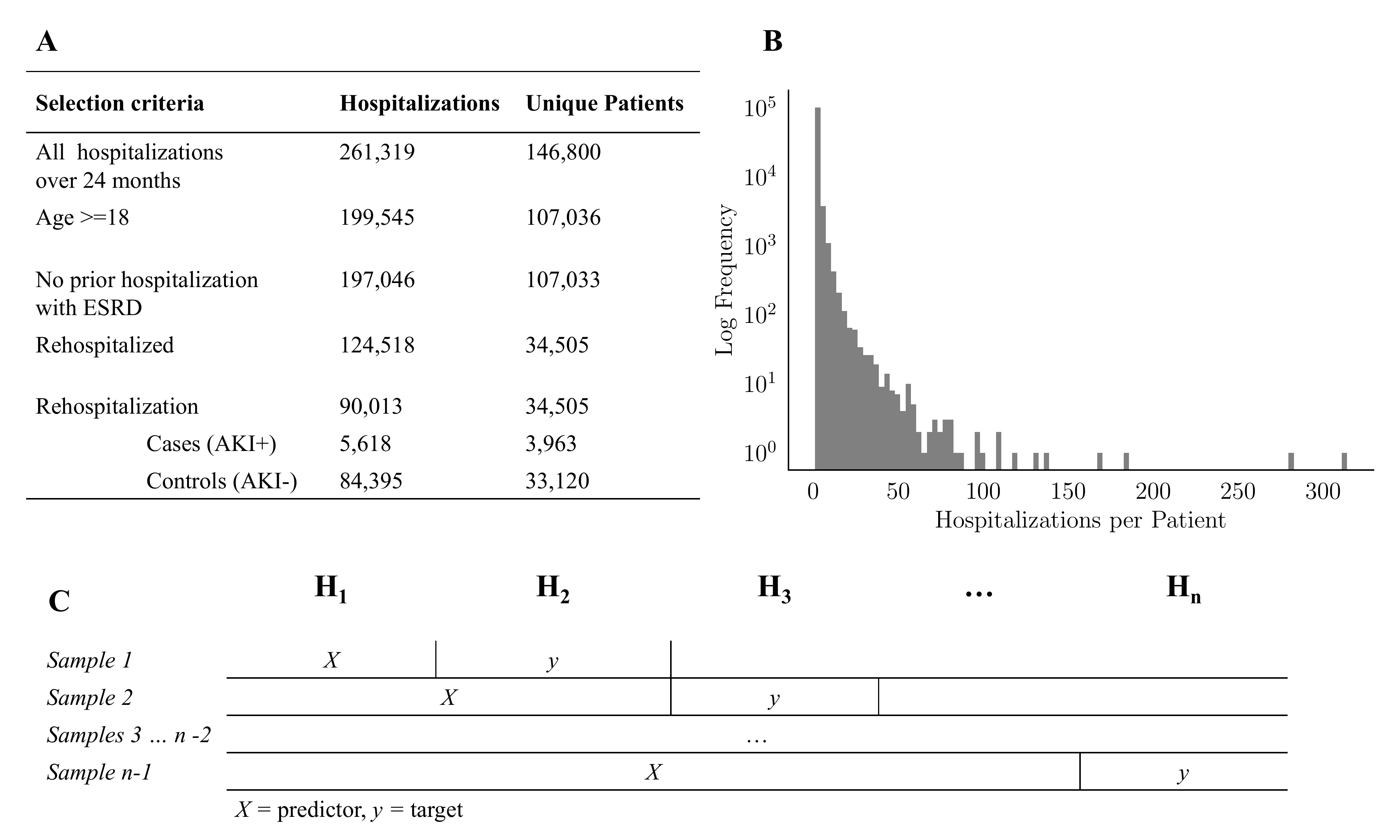}
\caption{{\bf Cohort selection.} On the top left, the selection procedure used to obtain the rehospitalization cohort is shown.  On the top right, the distribution of the 197,046 hospitalizations not preceded by a diagnosis of ESRD is shown.  On the bottom, a schematic of predictor/target generation is shown for an example patient with $n$ hospitalizations from which $n-1$ training cases were derived.  For each target rehospitalization, $y$, data from all prior hospitalizations, $X$, are used as predictors.  Multiple prior hospitalizations were aggregated using $G$ as described above.}
\label{cohor_sele}
\end{figure}

A cohort selection schema and results are shown in Fig~\ref{cohor_sele} along with a histogram of the number of hospitalizations per patient. During the two-year window, 146,800 patients generated 261,319 hospitalizations; after excluding hospitalizations with age at admission $<$ 18, 107,036 patients generated 199,545 hospitalizations. Excluding hospitalizations preceded by diagnosis of ESRD, but not preceded by a renal transplant, yielded 197,046 hospitalizations for 107,033 patients.  Of these patients, 34,505 (32.2\%) were rehospitalized at least once during the two-year period, accounting for 123,828 (62.8\%) of total hospitalizations.  Within hospitalizations generated by these patients, 90,013 were rehospitalizations (i.e., not the first hospitalization from that patient in our dataset).  There were 5,618 (6.2\%) cases of AKI.  The hospitalization:patient ratio was 1.4 for the cases, and 2.5 for controls.  Hence the cases showed more patient-level diversity than the controls, which were generated by patients who returned more often.  

\subsection*{AKI Diagnosis}
AKI was identified by both diagnosis code and sCr, shown in Table~\ref{dx_dist}.  Of all 197,046 hospitalizations in our cohort (not by a patient with previous diagnosed ESRD), 11,166 (5.7$\%$) involved AKI; 4,135 were diagnosed by sCr but not code, 2,747 by sCr and code, and 4,284 by code but not sCr.
\begin{table}[!ht] 
\centering 
\footnotesize
\caption{{\bf AKI diagnosis distribution}}
\begin{tabular}{cc rrr r }
\toprule
\multicolumn{3}{c}{} & \multicolumn{2}{c}{\bf Lab Diagnosis} & \multicolumn{1}{c}{\bf Total}\\ 
\multicolumn{3}{c}{} & \multicolumn{1}{c}{\bf +} & \multicolumn{1}{c}{\bf -} &\multicolumn{1}{c}{}\\ 
\cmidrule(lr){4-5} \cmidrule{6-6}\\
\textbf{Coding Diagnosis} & \textbf{+} & & 2,747 & 4,284 &  7,031 \\ 
{} & \textbf{-} &  & 4,135 & 185,880  &  190,015 \\[8 pt]
\cmidrule(lr){1-2} \cmidrule(lr){4-5} \cmidrule{6-6}\\
\textbf{Total} &  &  & 6,882	 & 190,164 &  197,046 \\ 
\bottomrule
\end{tabular}
\begin{flushleft} 
\end{flushleft}
\label{dx_dist}
\end{table}

Cohort demographics for all 124,518 adult hospitalizations (after exclusion of cases following a diagnosis of ESRD) generated by patients who were rehospitalized at some time are shown in Table~\ref{cohort_dem}.  This corresponds to the fourth cohort shown in Fig~\ref{cohor_sele}.  These summary statistics are by hospitalization, and not patient, and therefore some patients are represented multiple times.  Note that ESRD is present since a hospitalization can contain a diagnosis of ESRD (i.e., permanent kidney failure) even though it does not follow a hospitalization with diagnosis of ESRD.  General cohort demographics corresponded to known findings.  As expected, hospitalizations in which AKI occurred had higher age on admission~\cite{i5} and longer duration~\cite{chertow2005MortLOSCost}.  A higher proportion of AKI+ subjects were male and white.  Also more prevalent in the AKI+ hospitalizations were previously identified risk factors~\cite{koyner2016development, bellomo2012acute, ShustermanHospAcq} including prior CKD diagnosis~\cite{chawla2014acute}, prior dialysis procedures without ESRD~\cite{chawla2014acute}, congestive heart failure~\cite{ShustermanHospAcq, ronco2008cardiorenal}, diabetes~\cite{ShustermanHospAcq}, shock~\cite{LeblancRiskFact}, and liver failure~\cite{garcia2008acute, fede2012renal}.  

\begin{table}[!ht]
\begin{adjustwidth}{-0.05 in}{0 in} 
\tiny
\caption{{\bf Cohort demographics} Statistics are computed per hospitalization.  There are a total of 124,518 hospitalizations from 34,505 patients, each with more than one hospitalization.  These are therefore all hospitalizations generated by patients in the final cohort (including the first hospitalization from each patient, for which AKI is not predicted).}
\footnotesize 
\begin{tabular}{lrrrr}
\toprule
{} & \textbf{AKI+}  &(n=7,762)& \textbf{AKI-} & (n=116,756)  \\
\toprule
{} & Mean $\pm$ STD& Median & Mean $\pm$ STD &Median \\
\midrule
\textbf{Age}& 62.06 $\pm$ 17.23& 63.00 & 44.01 $\pm$ 18.86 &42.00\\
\textbf{Length of Stay}&14.22 $\pm$ 22.59 &7.89&1.89 $\pm$ 4.89& 0.33\\
\toprule
{} & \textbf{Count} & \% &  \textbf{Count} & \%\\
\midrule
\textbf{Female} &  3,497& 44.0 &  67,811 &56.0 \\
\textbf{American Indian} &      5& 0.0 &     205& 0.0 \\
\textbf{Asian} &     61& 1.0 &    1,364& 1.0 \\
\textbf{Black} &  1,822 &23.0 &  39,038 &28.0 \\
\textbf{White} &  5,557& 73.0 &  68,003& 64.0 \\
\textbf{Other} &    322& 4.0 &    8351& 7.0 \\
\midrule
\textbf{Chronic Kidney Disease (CKD)} &  2,574& 37.0 &    2,349& 2.0 \\
\textbf{ESRD} &   246& 10.0 &     213& 1.0 \\
\textbf{Dialysis} &   538 &15.0 &     217& 1.0 \\
\textbf{Renal Transplant} &     22& 0.0 &      29& 0.0 \\
\textbf{Unspecified Renal Failure} &    238 &3.0 &      72 & 0.0 \\
\midrule
\textbf{Congestive Heart Failure} &  2,227& 29.0 &    3,218 &2.0 \\
\textbf{Diabetes} &  2,651& 34.0 &    9,031& 7.0 \\
\textbf{Shock} &   997& 14.0 &    2,700& 2.0 \\
\textbf{Liver Failure} &    720 &9.0 &    1,076& 1.0 \\
\textbf{Rhabdomyolysis} &    189& 2.0 &     225 &0.0 \\
\bottomrule
\end{tabular}
\begin{flushleft}
\vspace{5 pt}
\end{flushleft}
\label{cohort_dem}
\end{adjustwidth}
\end{table}

\subsection*{Evaluation}

\begin{figure}[!ht]
\includegraphics[width=\linewidth]{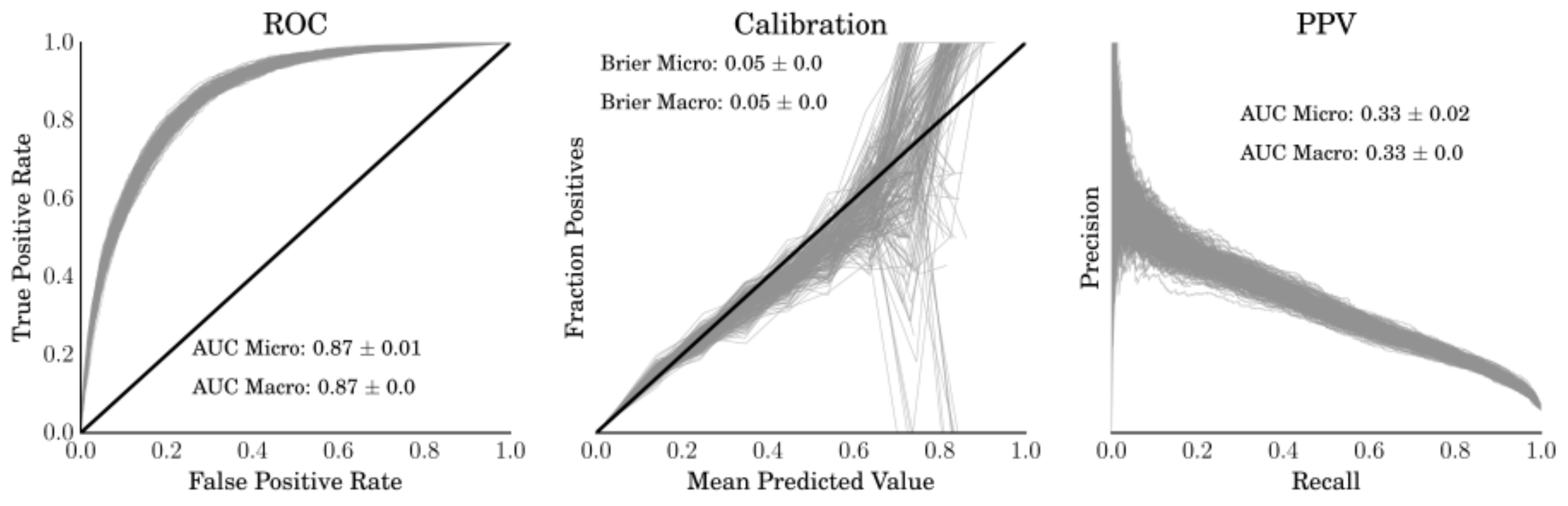}
\caption{\textbf{GBC evaluation.} ROC, Calibration, and PR curves for 50 iterations of 5-fold CV (250 lines shown; each of 50 iterations has 5 lines corresponding to the 5 outer folds of CV).  The black diagonal line represents chance for the ROC curve and ideal for the calibration curve.  Results are reported per hospitalization, not patient. Alpha level=0.5, line weight=0.5.}
\label{gbc_ev}
\end{figure}

The final dataset had 5,308 features at a code precision of 3 digits.  After removing features that were observed in fewer than 100 of the samples, 3,387 (63.8 \%) remained.  HP are detailed in Supplement~\nameref{alg_spec}.  All performance metrics are reported in Table~\ref{tab_perf}; since the distributions of these individual metrics were approximately normal (Supplement~\nameref{met_dist}), standard deviation is reported.  Also because of approximate normality, the Bayesian correlated t-test~\cite{BayesianTimeForBenavoli} was used to compare systems (Table~\ref{comp}).
We specified \textit{a priori} the regions of practical equivalence (ROPE) for ROC AUC, Brier Score, and PR AUC as, respectively, (0.01, 0.001, 0.01). For metric $m$ with ROPE $r$ and systems in row $i$ and column $j$, tuples in the table correspond to 
$(P(m(i)-m(j))>0.5r$, $P(m(i)-m(j)) \in r,$ $P(m(i)-m(j))<-0.5r$
or, informally, 
(P($i$ higher score than $j$), P($i$ and $j$ practically equivalent), P($j$  higher score than $i$)).  
Note that ROC and PR are both ideal if 1 and Brier score is ideal if 0, so the Brier table is opposite the other two.  We again emphasize that this is a comparison of trained systems, not of the training algorithms, because HP is a confounder. GBC curves are displayed in Figure~\ref{gbc_ev}.  In the low range, GBC has transposed-sigmoidal tendency suggesting overconfidence (predicting low probabilities as too low and high probabilities as too high).  This may be due to dependencies or perhaps a relatively small ratio of cases to features.  In contrast to high ROC AUC, precision suffers greatly when the threshold is lowered.  PPV is dependent on the prevalence of AKI; even a small false positive rate (FPR) might lead to a high false positive (FP) count if the controls outnumber the cases, as is the case with AKI.  Thus even with a low FPR, it can be expected that detecting a TP would cost many FP.  FP in AKI, relative to other diseases, are most often fairly innocuous.  Preventative measures consist mainly of hydration and medication review.  In some cases, however, a FP might result in withholding necessary treatment (e.g., imaging or medication) or unnecessary Nephrology consults~\cite{PalevskyCommentOnKDIGONephConsult}.  This shortcoming is therefore notable.  Ultimately, however, we recommend that a decision based on some threshold \textit{never} be provided to a user in place of a probability estimate~\cite{harrell2015regression}.

\newcommand\acs{ROC = Receiver Operating Characteristic, PR = Precision Recall, ALR1 = Anscombe LR1, GBC = Gradient Boosting Classifier, LR1 = $l1$-Penalized Logistic Regression, LSTM = Long Short-term Memory, HP = Highly Penalized, W = Weighted, S = Sampled, R = Recent (for GBC) or Randomized (for LR1, HPLR1), M = Medication, N = Noise.}

\begin{table}[!ht]
\caption{{\bf Predictive performance. }\small \acs}
\centering
\tiny
\begin{tabular}{p{1cm}|p{1cm}p{2.5cm}p{2.5cm}p{2.5cm}}
\toprule
{}  & & \textbf{ROC AUC}                        &  \textbf{Brier Score}                    &   \textbf{PR} \\
\midrule
\textbf{GBC} &Micro: &\textbf{0.86737  $\pm\ $0.00566} & \textbf{0.04901 $\pm\ $0.00179} & 0.32568 $\pm\ $0.01502\\
             &Macro: &\textbf{0.86737 $\pm\ $0.00045}  & \textbf{0.04901 $\pm\ $ 8e-05} & 0.32568 $\pm\ $0.0026\\
\midrule             
\textbf{LR1} &Micro: &0.86012 $\pm\ $0.00602 &  0.05038 $\pm\ $0.00187 & 0.30068 $\pm\ $0.01533\\
             &Macro: &0.86012 $\pm\ $0.00041  & 0.05038 $\pm\ $0.00011 & 0.30068 $\pm\ $0.00182\\

\midrule   

\textbf{ALR1} & Micro: &0.86188 $\pm\ $0.00606 & 0.05019 $\pm\ $0.00187 & 0.30445 $\pm\ $0.01571\\
                               & Macro: &0.86188 $\pm\ $0.00113  & 0.05019 $\pm\ $0.00025 & 0.30445 $\pm\ $0.00411\\

\midrule   

\textbf{RLR1} & Micro: &0.85312 $\pm\ $0.00621 & 0.05068 $\pm\ $0.0019 & 0.30227 $\pm\ $0.01453\\
             &Macro: &0.85312 $\pm\ $0.00055  & 0.05068 $\pm\ $9e-05 & 0.30227 $\pm\ $0.00159\\

\midrule  
\textbf{HPLR1}&Micro: &0.84545 $\pm\ $0.0064 & 0.05158 $\pm\ $0.00191 & 0.29002 $\pm\ $0.01361\\
              &Macro: &0.84545 $\pm\ $0.00037 & 0.05158 $\pm\ $5e-05 & 0.29002 $\pm\ $0.00091\\
\midrule

\textbf{RHPLR1}&Micro: &0.848 $\pm\ $0.00651 & 0.05102 $\pm\ $0.00192 &0.29869 $\pm\ $0.0142\\
              &Macro: &0.848 $\pm\ $0.05102 &0.05102 $\pm\ $8e-05     & 0.29869 $\pm\ $0.00122\\

\midrule

\textbf{WGBC} &Micro: &0.86328 $\pm\ $0.00568 & 0.04932 $\pm\ $0.00178 & 0.31572 $\pm\ $0.01541\\
             &Macro: &0.86328 $\pm\ $0.00059  & 0.04932 $\pm\ $0.0001 & 0.31572 $\pm\ $0.00261\\

\midrule   
\textbf{WLR1} &Micro: &0.84965 $\pm\ $0.00606 &  0.0507 $\pm\ $0.00188 & 0.29564 $\pm\ $0.01425\\
             &Macro: &0.84965 $\pm\ $0.00046  & 0.0507 $\pm\ $6e-05 & 0.29564 $\pm\ $0.00091\\

\midrule             
\textbf{WHPLR1}&Micro: &0.77923 $\pm\ $0.01208 & 0.05387 $\pm\ $0.00209 & 0.25742 $\pm\ $0.01609\\
              &Macro: &0.77923 $\pm\ $0.00308 & 0.05387 $\pm\ $0.00013 & 0.25742 $\pm\ $0.00385\\
\midrule              
\textbf{SGBC} &Micro: &0.85962 $\pm\ $0.00744 & 0.05326 $\pm\ $0.00195 & \textbf{0.33161 $\pm\ $0.02153}\\
             &Macro: &0.85962 $\pm\ $0.00226  & 0.05326 $\pm\ $0.00045 & \textbf{0.33161 $\pm\ $0.00631}\\
             
\midrule             
\textbf{SLR1} &Micro: &0.84752 $\pm\ $0.00792 &  0.05486 $\pm\ $0.00206 & 0.30596 $\pm\ $0.02022\\
             &Macro: &0.84752 $\pm\ $0.00214  &  0.05486 $\pm\ $0.00049 & 0.30596 $\pm\ $0.00548\\

\midrule             
\textbf{SHPLR1}&Micro: &0.7706 $\pm\ $0.01157 & 0.05754 $\pm\ $0.00229 & 0.28547 $\pm\ $0.01992\\
              &Macro: &0.7706 $\pm\ $0.00366 & 0.05754 $\pm\ $0.0005 & 0.28547 $\pm\ $0.0054\\

\midrule               
\textbf{RGBC} &Micro: &0.86306 $\pm\ $0.00572 & 0.04927 $\pm\ $0.00178 & 0.32198 $\pm\ $0.01526\\
              &Macro: &0.86306 $\pm\ $0.00039 & 0.04927 $\pm\ $6e-05 & 0.32198 $\pm\ $0.00185\\     
\midrule               
\textbf{MGBC} &Micro: &0.82635 $\pm\ $0.00693 & 0.05161 $\pm\ $0.00189 & 0.27079 $\pm\ $0.01484\\
              &Macro: &0.82635 $\pm\ $0.00075 & 0.05161 $\pm\ $8e-05 & 0.27079 $\pm\ $0.00172\\         
\midrule               
\textbf{MLR1} &Micro: &0.80671 $\pm\ $0.00764 & 0.0564 $\pm\ $0.0022 & 0.22051 $\pm\ $0.01397\\
              &Macro: &0.80671 $\pm\ $0.00137 & 0.0564 $\pm\ $0.00019 & 0.22051 $\pm\ $ 0.00212\\ 

\midrule
\textbf{LSTM} &Micro: &0.85744 $\pm\ $0.00592 & 0.05027 $\pm\ $0.0018 & 0.28209 $\pm\ $0.01547\\
              &Macro: &0.85744 $\pm\ $0.0008 & 0.05027 $\pm\ $0.00012 & 0.28209 $\pm\ $0.00526\\

\midrule
\textbf{CLR} &Micro: &0.80149 $\pm\ $0.00785 & 0.05356 $\pm\ $0.00204 & 0.22926 $\pm\ $0.01467\\
                              &Macro: &0.80149 $\pm\ $0.00034 & 0.05356 $\pm\ $6e-05   & 0.22926 $\pm\ $0.0009\\

\midrule
\textbf{NGBC} &Micro: &0.49938 $\pm\ $0.00837 & 0.05853 $\pm\ $0.0015 & 0.06251 $\pm\ $0.00242\\
              &Macro: &0.49938 $\pm\ $0.00399 & 0.05853 $\pm\ $2e-05 & 0.06251 $\pm\ $0.00094\\ 
\bottomrule
\end{tabular}
\label{tab_perf}
\end{table}

The distributions of errors by method of diagnosis (i.e., by code or sCr) are shown in Supplement~\nameref{error_by_dx}.  Without rigorous analysis, it appears that, expectedly, cases detected by both methods have lower mean error than cases detected by one or the other.  Notably, cases detected by sCr but not administrative code appear to have higher errors than cases detected by both or cases detected by code but not sCr; this is also to be expected since many cases detected by sCr but not code were likely subtle AKI episodes, or perhaps even correspond to variation in sCr for reasons impossible to discern from the data, but not due to AKI.  Gross visual differences between the distributions are not noted, but the slight differences could be an interesting future investigation.  
\newcommand\tw{0.32}
\newcommand\lenbox{18cm}
\newcommand\heightbox{1.8cm}
\begin{table}
\caption{
{\bf Predictive performance comparison. }\small \acs
}
\begin{minipage}[t]{\tw\textwidth}
\begin{adjustbox}{angle=-90}
\tiny
\resizebox{\lenbox}{\heightbox}{
\begin{tabular}{llllllllllllllllll}
\toprule
{PR} &            LR1 &             ALR1 &             RLR1 &            HPLR1 &           RHPLR1 &             WGBC &             WLR1 &         WHPLR1 &             SGBC &              SLR1 &            SHPLR1 &              RGBG &              MGBC &           MLR1 &              LSTM &              CLR &           NGBC \\
\midrule
GBC    &  1.0, 0.0, 0.0 &  0.99, 0.01, 0.0 &    1.0, 0.0, 0.0 &    1.0, 0.0, 0.0 &    1.0, 0.0, 0.0 &  0.49, 0.51, 0.0 &    1.0, 0.0, 0.0 &  1.0, 0.0, 0.0 &  0.06, 0.6, 0.34 &   0.84, 0.16, 0.0 &     1.0, 0.0, 0.0 &   0.07, 0.93, 0.0 &     1.0, 0.0, 0.0 &  1.0, 0.0, 0.0 &     1.0, 0.0, 0.0 &    1.0, 0.0, 0.0 &  1.0, 0.0, 0.0 \\
LR1    &---------------&    0.0, 1.0, 0.0 &  0.0, 0.99, 0.01 &  0.56, 0.44, 0.0 &  0.01, 0.99, 0.0 &  0.0, 0.16, 0.84 &  0.05, 0.95, 0.0 &  1.0, 0.0, 0.0 &  0.0, 0.03, 0.97 &  0.05, 0.64, 0.31 &   0.71, 0.29, 0.0 &   0.0, 0.02, 0.98 &     1.0, 0.0, 0.0 &  1.0, 0.0, 0.0 &   0.91, 0.09, 0.0 &    1.0, 0.0, 0.0 &  1.0, 0.0, 0.0 \\
ALR1   &---------------&  ---------------&  0.01, 0.99, 0.0 &  0.83, 0.17, 0.0 &  0.14, 0.86, 0.0 &    0.0, 0.4, 0.6 &  0.36, 0.64, 0.0 &  1.0, 0.0, 0.0 &  0.0, 0.05, 0.95 &  0.11, 0.71, 0.18 &   0.82, 0.17, 0.0 &   0.0, 0.09, 0.91 &     1.0, 0.0, 0.0 &  1.0, 0.0, 0.0 &   0.97, 0.03, 0.0 &    1.0, 0.0, 0.0 &  1.0, 0.0, 0.0 \\
RLR1   &---------------&  ---------------&  ---------------&  0.73, 0.27, 0.0 &  0.01, 0.99, 0.0 &  0.0, 0.21, 0.79 &  0.08, 0.92, 0.0 &  1.0, 0.0, 0.0 &  0.0, 0.03, 0.97 &  0.07, 0.68, 0.25 &   0.77, 0.23, 0.0 &   0.0, 0.02, 0.98 &     1.0, 0.0, 0.0 &  1.0, 0.0, 0.0 &   0.94, 0.06, 0.0 &    1.0, 0.0, 0.0 &  1.0, 0.0, 0.0 \\
HPLR1  &---------------&  ---------------&  ---------------&  ---------------&    0.0, 0.7, 0.3 &    0.0, 0.0, 1.0 &    0.0, 0.9, 0.1 &  1.0, 0.0, 0.0 &    0.0, 0.0, 1.0 &   0.0, 0.25, 0.75 &   0.26, 0.7, 0.04 &     0.0, 0.0, 1.0 &   0.96, 0.04, 0.0 &  1.0, 0.0, 0.0 &   0.36, 0.64, 0.0 &    1.0, 0.0, 0.0 &  1.0, 0.0, 0.0 \\
RHPLR1 &---------------&  ---------------&  ---------------&  ---------------&  ---------------&  0.0, 0.05, 0.95 &    0.0, 1.0, 0.0 &  1.0, 0.0, 0.0 &  0.0, 0.01, 0.99 &  0.03, 0.59, 0.38 &   0.64, 0.35, 0.0 &     0.0, 0.0, 1.0 &     1.0, 0.0, 0.0 &  1.0, 0.0, 0.0 &   0.85, 0.15, 0.0 &    1.0, 0.0, 0.0 &  1.0, 0.0, 0.0 \\
WGBC   &---------------&  ---------------&  ---------------&  ---------------&  ---------------&  ---------------&  0.99, 0.01, 0.0 &  1.0, 0.0, 0.0 &  0.0, 0.27, 0.72 &  0.49, 0.49, 0.02 &   0.98, 0.02, 0.0 &     0.0, 0.8, 0.2 &     1.0, 0.0, 0.0 &  1.0, 0.0, 0.0 &     1.0, 0.0, 0.0 &    1.0, 0.0, 0.0 &  1.0, 0.0, 0.0 \\
WLR1   &---------------&  ---------------&  ---------------&  ---------------&  ---------------&  ---------------&  ---------------&  1.0, 0.0, 0.0 &  0.0, 0.01, 0.99 &  0.01, 0.47, 0.51 &  0.51, 0.48, 0.01 &     0.0, 0.0, 1.0 &     1.0, 0.0, 0.0 &  1.0, 0.0, 0.0 &   0.71, 0.29, 0.0 &    1.0, 0.0, 0.0 &  1.0, 0.0, 0.0 \\
WHPLR1 &---------------&  ---------------&  ---------------&  ---------------&  ---------------&  ---------------&  ---------------&---------------&    0.0, 0.0, 1.0 &     0.0, 0.0, 1.0 &   0.0, 0.04, 0.96 &     0.0, 0.0, 1.0 &   0.0, 0.34, 0.66 &  1.0, 0.0, 0.0 &   0.0, 0.05, 0.95 &  0.98, 0.02, 0.0 &  1.0, 0.0, 0.0 \\
SGBC   &---------------&  ---------------&  ---------------&  ---------------&  ---------------&  ---------------&  ---------------&---------------&  ---------------&   0.99, 0.01, 0.0 &     1.0, 0.0, 0.0 &  0.48, 0.49, 0.02 &     1.0, 0.0, 0.0 &  1.0, 0.0, 0.0 &     1.0, 0.0, 0.0 &    1.0, 0.0, 0.0 &  1.0, 0.0, 0.0 \\
SLR1   &---------------&  ---------------&  ---------------&  ---------------&  ---------------&  ---------------&  ---------------&---------------&  ---------------&   ---------------&   0.97, 0.03, 0.0 &   0.0, 0.27, 0.73 &   0.99, 0.01, 0.0 &  1.0, 0.0, 0.0 &   0.89, 0.11, 0.0 &    1.0, 0.0, 0.0 &  1.0, 0.0, 0.0 \\
SHPLR1 &---------------&  ---------------&  ---------------&  ---------------&  ---------------&  ---------------&  ---------------&---------------&  ---------------&   ---------------&   ---------------&     0.0, 0.0, 1.0 &  0.68, 0.31, 0.01 &  1.0, 0.0, 0.0 &  0.26, 0.63, 0.11 &    1.0, 0.0, 0.0 &  1.0, 0.0, 0.0 \\
RGBG   &---------------&  ---------------&  ---------------&  ---------------&  ---------------&  ---------------&  ---------------&---------------&  ---------------&   ---------------&   ---------------&   ---------------&     1.0, 0.0, 0.0 &  1.0, 0.0, 0.0 &     1.0, 0.0, 0.0 &    1.0, 0.0, 0.0 &  1.0, 0.0, 0.0 \\
MGBC   &---------------&  ---------------&  ---------------&  ---------------&  ---------------&  ---------------&  ---------------&---------------&  ---------------&   ---------------&   ---------------&   ---------------&   ---------------&  1.0, 0.0, 0.0 &   0.0, 0.41, 0.59 &    1.0, 0.0, 0.0 &  1.0, 0.0, 0.0 \\
MLR1   &---------------&  ---------------&  ---------------&  ---------------&  ---------------&  ---------------&  ---------------&---------------&  ---------------&   ---------------&   ---------------&   ---------------&   ---------------&---------------&     0.0, 0.0, 1.0 &  0.0, 0.58, 0.42 &  1.0, 0.0, 0.0 \\
LSTM   &---------------&  ---------------&  ---------------&  ---------------&  ---------------&  ---------------&  ---------------&---------------&  ---------------&   ---------------&   ---------------&   ---------------&   ---------------&---------------&   ---------------&    1.0, 0.0, 0.0 &  1.0, 0.0, 0.0 \\
CLR    &---------------&  ---------------&  ---------------&  ---------------&  ---------------&  ---------------&  ---------------&---------------&  ---------------&   ---------------&   ---------------&   ---------------&   ---------------&---------------&   ---------------&  ---------------&  1.0, 0.0, 0.0 \\
NGBC   &---------------&  ---------------&  ---------------&  ---------------&  ---------------&  ---------------&  ---------------&---------------&  ---------------&   ---------------&   ---------------&   ---------------&   ---------------&---------------&   ---------------&  ---------------&---------------\\
\bottomrule
\end{tabular}

}
\end{adjustbox}
\end{minipage}
\hfill
\begin{minipage}[t]{\tw\textwidth}
\begin{adjustbox}{angle=-90}
\tiny
\resizebox{\lenbox}{\heightbox}{
\begin{tabular}{llllllllllllllllll}
\toprule
{Brier} &              LR1 &             ALR1 &           RLR1 &            HPLR1 &           RHPLR1 &             WGBC &             WLR1 &         WHPLR1 &              SGBC &              SLR1 &           SHPLR1 &             RGBG &             MGBC &              MLR1 &             LSTM &               CLR &             NGBC \\
\midrule
GBC    &  0.0, 0.05, 0.95 &  0.0, 0.23, 0.77 &  0.0, 0.0, 1.0 &    0.0, 0.0, 1.0 &    0.0, 0.0, 1.0 &    0.0, 1.0, 0.0 &    0.0, 0.0, 1.0 &  0.0, 0.0, 1.0 &     0.0, 0.0, 1.0 &     0.0, 0.0, 1.0 &    0.0, 0.0, 1.0 &    0.0, 1.0, 0.0 &    0.0, 0.0, 1.0 &     0.0, 0.0, 1.0 &  0.0, 0.13, 0.87 &     0.0, 0.0, 1.0 &    0.0, 0.0, 1.0 \\
LR1    &  --------------- &    0.0, 1.0, 0.0 &  0.0, 1.0, 0.0 &  0.0, 0.19, 0.81 &  0.0, 0.97, 0.03 &    0.6, 0.4, 0.0 &    0.0, 1.0, 0.0 &  0.0, 0.0, 1.0 &   0.0, 0.03, 0.97 &     0.0, 0.0, 1.0 &    0.0, 0.0, 1.0 &  0.66, 0.34, 0.0 &  0.0, 0.19, 0.81 &     0.0, 0.0, 1.0 &    0.0, 1.0, 0.0 &     0.0, 0.0, 1.0 &    0.0, 0.0, 1.0 \\
ALR1   &  --------------- &  --------------- &  0.0, 1.0, 0.0 &  0.0, 0.06, 0.94 &  0.0, 0.79, 0.21 &    0.3, 0.7, 0.0 &    0.0, 1.0, 0.0 &  0.0, 0.0, 1.0 &   0.0, 0.02, 0.98 &     0.0, 0.0, 1.0 &    0.0, 0.0, 1.0 &  0.38, 0.62, 0.0 &  0.0, 0.07, 0.93 &     0.0, 0.0, 1.0 &    0.0, 1.0, 0.0 &     0.0, 0.0, 1.0 &    0.0, 0.0, 1.0 \\
RLR1   &  --------------- &  --------------- &--------------- &    0.0, 0.7, 0.3 &    0.0, 1.0, 0.0 &  0.96, 0.04, 0.0 &    0.0, 1.0, 0.0 &  0.0, 0.0, 1.0 &   0.0, 0.05, 0.95 &     0.0, 0.0, 1.0 &    0.0, 0.0, 1.0 &  0.96, 0.04, 0.0 &  0.0, 0.61, 0.39 &     0.0, 0.0, 1.0 &  0.01, 0.99, 0.0 &     0.0, 0.0, 1.0 &    0.0, 0.0, 1.0 \\
HPLR1  &  --------------- &  --------------- &--------------- &  --------------- &    0.0, 1.0, 0.0 &    1.0, 0.0, 0.0 &  0.27, 0.73, 0.0 &  0.0, 0.0, 1.0 &   0.0, 0.24, 0.76 &   0.0, 0.01, 0.99 &    0.0, 0.0, 1.0 &    1.0, 0.0, 0.0 &    0.0, 1.0, 0.0 &     0.0, 0.0, 1.0 &  0.88, 0.12, 0.0 &     0.0, 0.0, 1.0 &    0.0, 0.0, 1.0 \\
RHPLR1 &  --------------- &  --------------- &--------------- &  --------------- &  --------------- &    1.0, 0.0, 0.0 &    0.0, 1.0, 0.0 &  0.0, 0.0, 1.0 &     0.0, 0.1, 0.9 &     0.0, 0.0, 1.0 &    0.0, 0.0, 1.0 &    1.0, 0.0, 0.0 &  0.0, 0.95, 0.05 &     0.0, 0.0, 1.0 &  0.18, 0.82, 0.0 &     0.0, 0.0, 1.0 &    0.0, 0.0, 1.0 \\
WGBC   &  --------------- &  --------------- &--------------- &  --------------- &  --------------- &  --------------- &  0.0, 0.04, 0.96 &  0.0, 0.0, 1.0 &     0.0, 0.0, 1.0 &     0.0, 0.0, 1.0 &    0.0, 0.0, 1.0 &    0.0, 1.0, 0.0 &    0.0, 0.0, 1.0 &     0.0, 0.0, 1.0 &  0.0, 0.59, 0.41 &     0.0, 0.0, 1.0 &    0.0, 0.0, 1.0 \\
WLR1   &  --------------- &  --------------- &--------------- &  --------------- &  --------------- &  --------------- &  --------------- &  0.0, 0.0, 1.0 &   0.0, 0.05, 0.95 &     0.0, 0.0, 1.0 &    0.0, 0.0, 1.0 &  0.96, 0.04, 0.0 &  0.0, 0.63, 0.37 &     0.0, 0.0, 1.0 &  0.01, 0.99, 0.0 &     0.0, 0.0, 1.0 &    0.0, 0.0, 1.0 \\
WHPLR1 &  --------------- &  --------------- &--------------- &  --------------- &  --------------- &  --------------- &  --------------- &--------------- &  0.36, 0.58, 0.06 &  0.03, 0.47, 0.49 &  0.0, 0.01, 0.99 &    1.0, 0.0, 0.0 &    1.0, 0.0, 0.0 &     0.0, 0.0, 1.0 &    1.0, 0.0, 0.0 &   0.05, 0.95, 0.0 &    0.0, 0.0, 1.0 \\
SGBC   &  --------------- &  --------------- &--------------- &  --------------- &  --------------- &  --------------- &  --------------- &--------------- &   --------------- &   0.0, 0.03, 0.97 &    0.0, 0.0, 1.0 &    1.0, 0.0, 0.0 &  0.76, 0.24, 0.0 &   0.0, 0.02, 0.98 &  0.98, 0.02, 0.0 &   0.1, 0.64, 0.25 &    0.0, 0.0, 1.0 \\
SLR1   &  --------------- &  --------------- &--------------- &  --------------- &  --------------- &  --------------- &  --------------- &--------------- &   --------------- &   --------------- &    0.0, 0.0, 1.0 &    1.0, 0.0, 0.0 &  0.99, 0.01, 0.0 &    0.01, 0.3, 0.7 &    1.0, 0.0, 0.0 &  0.61, 0.38, 0.02 &  0.0, 0.02, 0.98 \\
SHPLR1 &  --------------- &  --------------- &--------------- &  --------------- &  --------------- &  --------------- &  --------------- &--------------- &   --------------- &   --------------- &  --------------- &    1.0, 0.0, 0.0 &    1.0, 0.0, 0.0 &  0.55, 0.42, 0.03 &    1.0, 0.0, 0.0 &     1.0, 0.0, 0.0 &  0.08, 0.43, 0.5 \\
RGBG   &  --------------- &  --------------- &--------------- &  --------------- &  --------------- &  --------------- &  --------------- &--------------- &   --------------- &   --------------- &  --------------- &  --------------- &    0.0, 0.0, 1.0 &     0.0, 0.0, 1.0 &  0.0, 0.51, 0.49 &     0.0, 0.0, 1.0 &    0.0, 0.0, 1.0 \\
MGBC   &  --------------- &  --------------- &--------------- &  --------------- &  --------------- &  --------------- &  --------------- &--------------- &   --------------- &   --------------- &  --------------- &  --------------- &  --------------- &     0.0, 0.0, 1.0 &  0.91, 0.09, 0.0 &     0.0, 0.0, 1.0 &    0.0, 0.0, 1.0 \\
MLR1   &  --------------- &  --------------- &--------------- &  --------------- &  --------------- &  --------------- &  --------------- &--------------- &   --------------- &   --------------- &  --------------- &  --------------- &  --------------- &   --------------- &    1.0, 0.0, 0.0 &     1.0, 0.0, 0.0 &  0.01, 0.19, 0.8 \\
LSTM   &  --------------- &  --------------- &--------------- &  --------------- &  --------------- &  --------------- &  --------------- &--------------- &   --------------- &   --------------- &  --------------- &  --------------- &  --------------- &   --------------- &  --------------- &     0.0, 0.0, 1.0 &    0.0, 0.0, 1.0 \\
CLR    &  --------------- &  --------------- &--------------- &  --------------- &  --------------- &  --------------- &  --------------- &--------------- &   --------------- &   --------------- &  --------------- &  --------------- &  --------------- &   --------------- &  --------------- &   --------------- &    0.0, 0.0, 1.0 \\
NGBC   &  --------------- &  --------------- &--------------- &  --------------- &  --------------- &  --------------- &  --------------- &--------------- &   --------------- &   --------------- &  --------------- &  --------------- &  --------------- &   --------------- &  --------------- &   --------------- &  --------------- \\
\bottomrule
\end{tabular}
}
\end{adjustbox}
\end{minipage}
\hfill
\begin{minipage}[t]{\tw\textwidth}
\begin{adjustbox}{angle=-90}
\tiny
\resizebox{\lenbox}{\heightbox}{

\begin{tabular}{llllllllllllllllll}
\toprule
{ROC} &              LR1 &           ALR1 &             RLR1 &            HPLR1 &           RHPLR1 &             WGBC &             WLR1 &         WHPLR1 &             SGBC &             SLR1 &           SHPLR1 &             RGBG &           MGBC &           MLR1 &             LSTM &              CLR &           NGBC \\
\midrule
GBC    &  0.01, 0.99, 0.0 &  0.0, 1.0, 0.0 &    1.0, 0.0, 0.0 &    1.0, 0.0, 0.0 &    1.0, 0.0, 0.0 &    0.0, 1.0, 0.0 &    1.0, 0.0, 0.0 &  1.0, 0.0, 0.0 &  0.25, 0.75, 0.0 &    1.0, 0.0, 0.0 &    1.0, 0.0, 0.0 &    0.0, 1.0, 0.0 &  1.0, 0.0, 0.0 &  1.0, 0.0, 0.0 &  0.49, 0.51, 0.0 &    1.0, 0.0, 0.0 &  1.0, 0.0, 0.0 \\
LR1    &  --------------- &  0.0, 1.0, 0.0 &    0.0, 1.0, 0.0 &    1.0, 0.0, 0.0 &  0.91, 0.09, 0.0 &    0.0, 1.0, 0.0 &  0.66, 0.34, 0.0 &  1.0, 0.0, 0.0 &    0.0, 1.0, 0.0 &  0.77, 0.23, 0.0 &    1.0, 0.0, 0.0 &    0.0, 1.0, 0.0 &  1.0, 0.0, 0.0 &  1.0, 0.0, 0.0 &    0.0, 1.0, 0.0 &    1.0, 0.0, 0.0 &  1.0, 0.0, 0.0 \\
ALR1   &  --------------- &--------------- &  0.16, 0.84, 0.0 &    1.0, 0.0, 0.0 &  0.99, 0.01, 0.0 &    0.0, 1.0, 0.0 &  0.95, 0.05, 0.0 &  1.0, 0.0, 0.0 &  0.01, 0.99, 0.0 &  0.89, 0.11, 0.0 &    1.0, 0.0, 0.0 &    0.0, 1.0, 0.0 &  1.0, 0.0, 0.0 &  1.0, 0.0, 0.0 &    0.0, 1.0, 0.0 &    1.0, 0.0, 0.0 &  1.0, 0.0, 0.0 \\
RLR1   &  --------------- &--------------- &  --------------- &  0.03, 0.97, 0.0 &    0.0, 1.0, 0.0 &  0.0, 0.45, 0.55 &    0.0, 1.0, 0.0 &  1.0, 0.0, 0.0 &  0.0, 0.85, 0.15 &    0.1, 0.9, 0.0 &    1.0, 0.0, 0.0 &  0.0, 0.51, 0.49 &  1.0, 0.0, 0.0 &  1.0, 0.0, 0.0 &    0.0, 1.0, 0.0 &    1.0, 0.0, 0.0 &  1.0, 0.0, 0.0 \\
HPLR1  &  --------------- &--------------- &  --------------- &  --------------- &    0.0, 1.0, 0.0 &    0.0, 0.0, 1.0 &    0.0, 1.0, 0.0 &  1.0, 0.0, 0.0 &  0.0, 0.11, 0.89 &  0.0, 0.99, 0.01 &    1.0, 0.0, 0.0 &    0.0, 0.0, 1.0 &  1.0, 0.0, 0.0 &  1.0, 0.0, 0.0 &  0.0, 0.21, 0.79 &    1.0, 0.0, 0.0 &  1.0, 0.0, 0.0 \\
RHPLR1 &  --------------- &--------------- &  --------------- &  --------------- &  --------------- &    0.0, 0.0, 1.0 &    0.0, 1.0, 0.0 &  1.0, 0.0, 0.0 &  0.0, 0.32, 0.68 &   0.0, 0.99, 0.0 &    1.0, 0.0, 0.0 &  0.0, 0.01, 0.99 &  1.0, 0.0, 0.0 &  1.0, 0.0, 0.0 &  0.0, 0.59, 0.41 &    1.0, 0.0, 0.0 &  1.0, 0.0, 0.0 \\
WGBC   &  --------------- &--------------- &  --------------- &  --------------- &  --------------- &  --------------- &    1.0, 0.0, 0.0 &  1.0, 0.0, 0.0 &  0.03, 0.97, 0.0 &  0.94, 0.06, 0.0 &    1.0, 0.0, 0.0 &    0.0, 1.0, 0.0 &  1.0, 0.0, 0.0 &  1.0, 0.0, 0.0 &  0.02, 0.98, 0.0 &    1.0, 0.0, 0.0 &  1.0, 0.0, 0.0 \\
WLR1   &  --------------- &--------------- &  --------------- &  --------------- &  --------------- &  --------------- &  --------------- &  1.0, 0.0, 0.0 &    0.0, 0.5, 0.5 &  0.01, 0.99, 0.0 &    1.0, 0.0, 0.0 &  0.0, 0.03, 0.97 &  1.0, 0.0, 0.0 &  1.0, 0.0, 0.0 &  0.0, 0.83, 0.17 &    1.0, 0.0, 0.0 &  1.0, 0.0, 0.0 \\
WHPLR1 &  --------------- &--------------- &  --------------- &  --------------- &  --------------- &  --------------- &  --------------- &--------------- &    0.0, 0.0, 1.0 &    0.0, 0.0, 1.0 &  0.42, 0.58, 0.0 &    0.0, 0.0, 1.0 &  0.0, 0.0, 1.0 &  0.0, 0.0, 1.0 &    0.0, 0.0, 1.0 &  0.0, 0.02, 0.98 &  1.0, 0.0, 0.0 \\
SGBC   &  --------------- &--------------- &  --------------- &  --------------- &  --------------- &  --------------- &  --------------- &--------------- &  --------------- &  0.87, 0.13, 0.0 &    1.0, 0.0, 0.0 &  0.0, 0.97, 0.03 &  1.0, 0.0, 0.0 &  1.0, 0.0, 0.0 &  0.02, 0.98, 0.0 &    1.0, 0.0, 0.0 &  1.0, 0.0, 0.0 \\
SLR1   &  --------------- &--------------- &  --------------- &  --------------- &  --------------- &  --------------- &  --------------- &--------------- &  --------------- &  --------------- &    1.0, 0.0, 0.0 &  0.0, 0.07, 0.93 &  1.0, 0.0, 0.0 &  1.0, 0.0, 0.0 &  0.0, 0.51, 0.49 &    1.0, 0.0, 0.0 &  1.0, 0.0, 0.0 \\
SHPLR1 &  --------------- &--------------- &  --------------- &  --------------- &  --------------- &  --------------- &  --------------- &--------------- &  --------------- &  --------------- &  --------------- &    0.0, 0.0, 1.0 &  0.0, 0.0, 1.0 &  0.0, 0.0, 1.0 &    0.0, 0.0, 1.0 &    0.0, 0.0, 1.0 &  1.0, 0.0, 0.0 \\
RGBG   &  --------------- &--------------- &  --------------- &  --------------- &  --------------- &  --------------- &  --------------- &--------------- &  --------------- &  --------------- &  --------------- &  --------------- &  1.0, 0.0, 0.0 &  1.0, 0.0, 0.0 &  0.01, 0.99, 0.0 &    1.0, 0.0, 0.0 &  1.0, 0.0, 0.0 \\
MGBC   &  --------------- &--------------- &  --------------- &  --------------- &  --------------- &  --------------- &  --------------- &--------------- &  --------------- &  --------------- &  --------------- &  --------------- &--------------- &  1.0, 0.0, 0.0 &    0.0, 0.0, 1.0 &    1.0, 0.0, 0.0 &  1.0, 0.0, 0.0 \\
MLR1   &  --------------- &--------------- &  --------------- &  --------------- &  --------------- &  --------------- &  --------------- &--------------- &  --------------- &  --------------- &  --------------- &  --------------- &--------------- &--------------- &    0.0, 0.0, 1.0 &  0.11, 0.89, 0.0 &  1.0, 0.0, 0.0 \\
LSTM   &  --------------- &--------------- &  --------------- &  --------------- &  --------------- &  --------------- &  --------------- &--------------- &  --------------- &  --------------- &  --------------- &  --------------- &--------------- &--------------- &  --------------- &    1.0, 0.0, 0.0 &  1.0, 0.0, 0.0 \\
CLR    &  --------------- &--------------- &  --------------- &  --------------- &  --------------- &  --------------- &  --------------- &--------------- &  --------------- &  --------------- &  --------------- &  --------------- &--------------- &--------------- &  --------------- &  --------------- &  1.0, 0.0, 0.0 \\
NGBC   &  --------------- &--------------- &  --------------- &  --------------- &  --------------- &  --------------- &  --------------- &--------------- &  --------------- &  --------------- &  --------------- &  --------------- &--------------- &--------------- &  --------------- &  --------------- &--------------- \\
\bottomrule
\end{tabular}
}
\end{adjustbox}
\end{minipage}
\label{comp}
\end{table}

Performance curves for LR1, ALR1, and RLR1 are shown in Supplement~\nameref{lr1_ev}, Supplement~\nameref{alr1_ev}, and Supplement~\nameref{rlr1_ev}.  Performance curves for HPLR1 and RHPLR1 are shown in Supplement~\nameref{lr1_pen} and Supplement~\nameref{rlr1_pen}.  Stability selection included more variables, perhaps since it was less influenced by colinearity.  The performance difference between full LR1 and reduced HPLR1 suggests that adjusting for more variables improves, but also increase the variance, of the calibration curves.  Performance curves for weighted WGBC are shown in Supplement~\nameref{per_wgbc}.  When weighting, averaged calibration curves appear to be slightly closer to identity (Supplement~\nameref{cal_dist_WGBC}). When weighting, performance by utilization, shown in Supplement~\nameref{wgbc_ut}, appears unchanged.  Performance curves for WLR1 and WHPLR1 are shown in Supplement~\nameref{per_wlr1} and Supplement~\nameref{per_whplr1}, respectively.  Performance curves for sampled SGBC, SLR1, and SHPLR1 are shown in Supplement~\nameref{per_sgbc}, Supplement~\nameref{per_slr1}, and Supplement~\nameref{per_shplr1}.  Sampling leads to reduced sample size, and therefore performance appears to generally be worse, but the change is not drastic.  Notably, however, PR AUC increases.  Performance curves for RGBC, which takes into account only the most recent hospitalization, are shown in Supplement~\nameref{memoryless}.  It is evident that most predictive power is contained in the most recent hospitalization, but a small gain is achieved by including more distant hospitalizations (GBC appears slightly better than RGBC, but the difference is in the region of practical equivalence).  Notably, in GBC there are virtually no sums over hospitalizations, only means; when $G$ aggregates sequences of hospitalizations of variable length, sums have much higher variance (perhaps why GBC vastly favors labs, which can be converted to less volatile means, while diagnoses are mostly counts).  However perhaps when $G$ is the identity, such as with RGBC (features not shown), counts are just as well as means.  The medication-based performance curves for MGBC and MLR1 are shown respectively in Supplement~\nameref{med_GBC} and Supplement~\nameref{med_LR1}. Performance of CLR is shown in Supplement~\nameref{per_CLR}.  This system depends mostly on codes rather than continuous values, probably explaining its reduced performance.  Performance of LSTM is shown in Supplement~\nameref{per_LSTM}.  This system was not exhaustively optimized, so performance is not as strong, but it has the obvious benefit of requiring less feature engineering. Results for NGBC performance, a utilization analysis, and the STD with respect to error are shown in, respectively, Supplement~\nameref{n_GBC_per}, Supplement~\nameref{n_GBC_ut}, and Supplement~\nameref{n_GBC_std}.  NGBC just predicted 0.06 for every sample.

\begin{figure}[!ht]
\centering
\includegraphics[width=0.8\linewidth]{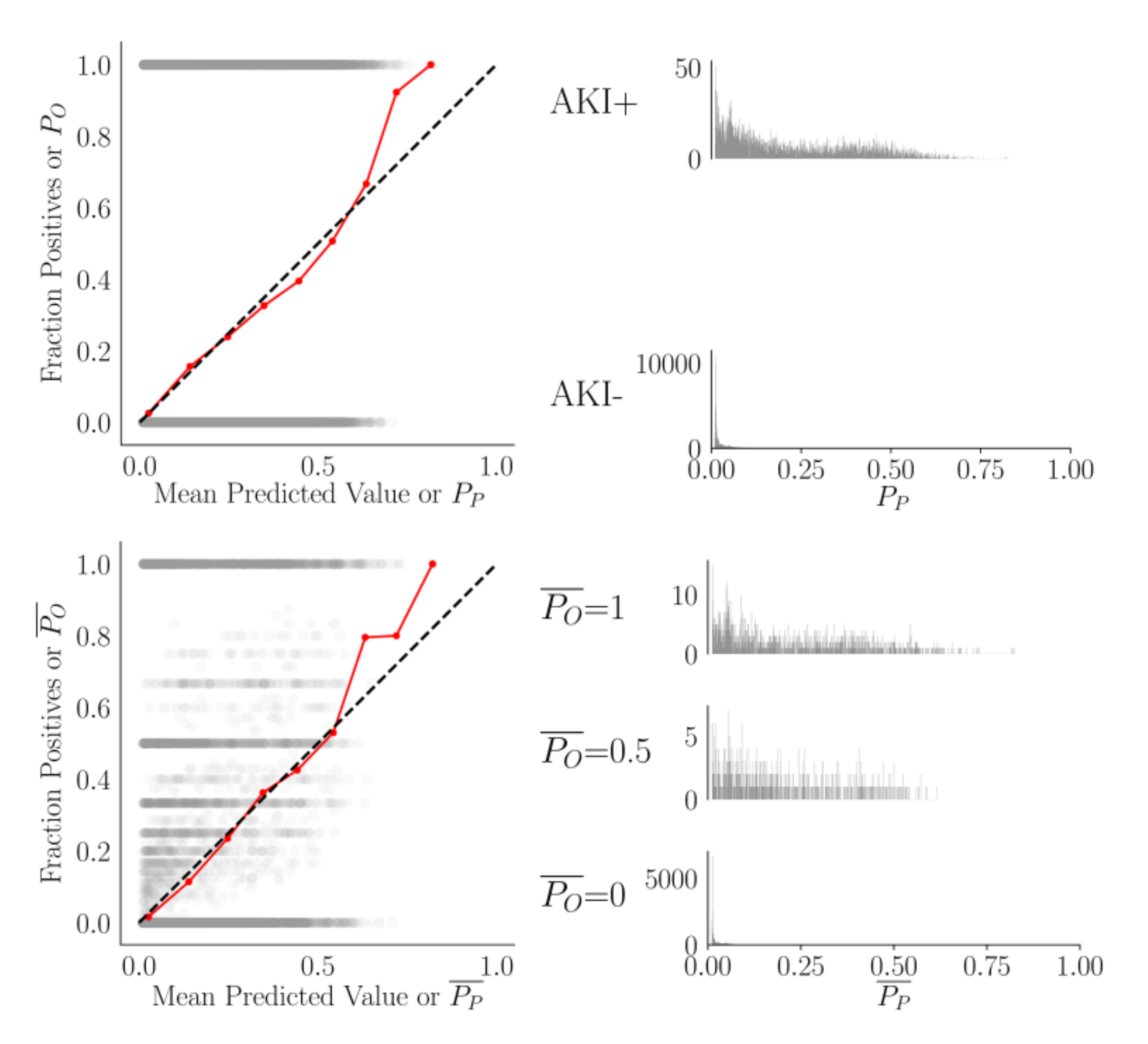}
\vfill
\caption{\textbf{GBC hospitalization- and patient-specific risk distributions.}  Observed hospitalization-level risk is plotted against predicted risk (top row) and patient-level mean observed risk against mean predicted risk (bottom row).  Distributions of predictions $P_P$ are shown at the hospitalization and patient level.  At patient level, distributions that are difficult to discern from the scatter plot are shown. In the scatter plots, alpha level is 0.05 and the red calibration curve corresponds to all hospitalizations or to patients who had either mean risk over hospitalizations of 1 or 0.  The calibration curves are computed according to the macro-averaged predicted output per hospitalization or patient over the 50 iterations of 5 fold CV (over 250 total folds).  Ideal calibration is the dotted black diagonal.  Histograms have 1000 bins to give necessary resolution.  $P_O$ = observed risk per hospitalization, $P_P$ = predicted risk per hospitalization, $\overline{P_O}$ = mean observed risk over hospitalizations, $\overline{P_P}$ = mean predicted risk over hospitalizations.}  
\label{predicted_prob}
\end{figure}

Fig~\ref{predicted_prob} shows the distributions of the probability estimates per hospitalization alongside the same per patient, where the risk is averaged over hospitalizations.  Although GBC was not optimized for patient-level prediction, aggregate calibration (averaged over CV folds and trials) appears to be good at the patient level.  The calibration curve consisting of averaged predictions is much better than the individual calibration curves per fold.  This may be related to the difficulty in sampling each fold at the patient level when there is such a wide variety of hospitalizations per patient.  More could be done on characterizing the distributions of the calibration curves.  It is apparent from these plots that it is more difficult to predict cases than controls; the distributions of predictions for cases are quite broad and appear almost bimodal.

\begin{figure}[!ht]
\centering
 \includegraphics[width=0.8\linewidth]{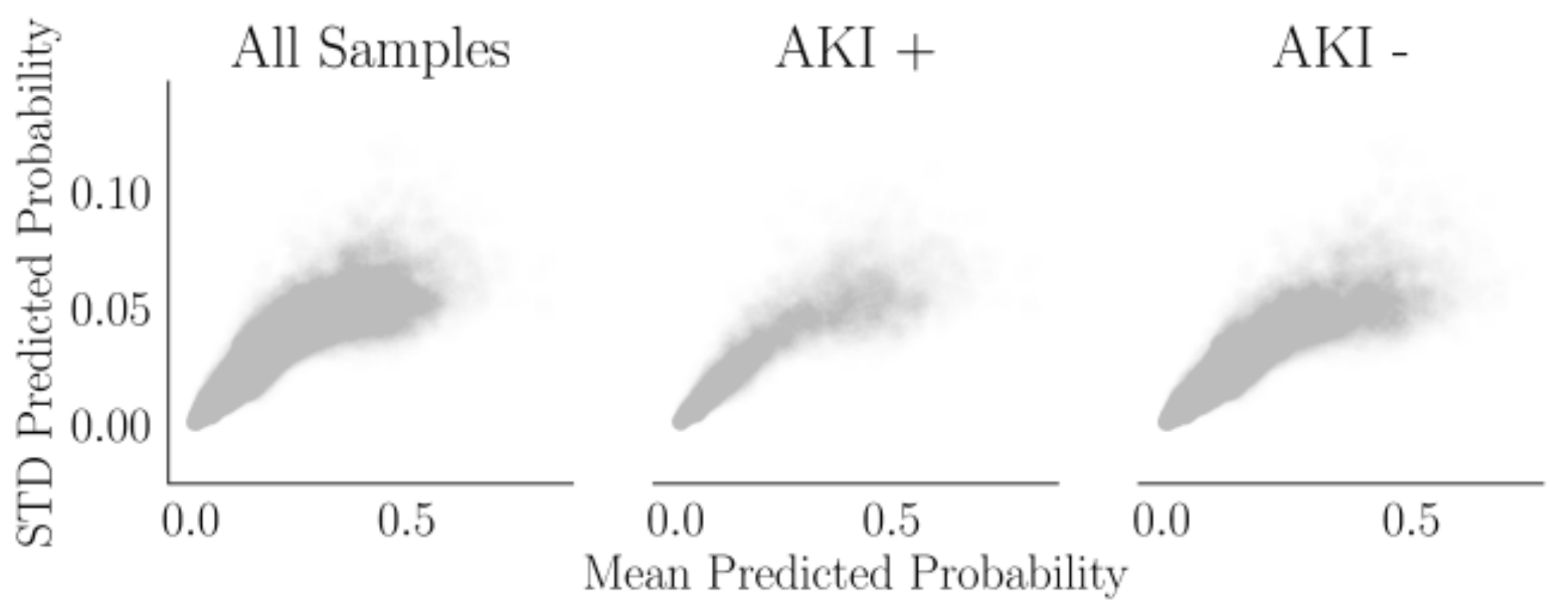}
\caption{\textbf{GBC prediction variance.} The mean and standard deviation of predicted probabilities are plotted over iterations (per hospitalization). Alpha=0.01 for all plots.}
\label{variance}
\end{figure}

Uncertainty of predictions appears to increase with increasing predicted risk, even when stratifying by outcome, as shown in Fig~\ref{variance}.  Although the range is fairly small (0-0.10), the distributions in Fig~\ref{predicted_prob} show that many of the high risk cases have low predicted risk, so the uncertainty is meaningful.  We highlight the necessity of (at least empirical) prediction intervals for GBC, if ever considered for deployment.

\subsection*{Error analysis of GBC}
We sought to identify specific subgroups which are easily recognized by a provider and for which the best performing GBC might make errors.  To show that the linear regressions used for this purpose were well fit, train, validation, and test MSE from the fold used to set HP is provided in Supplement~\nameref{error_train_val}.  HP are provided as well, chosen manually as in the main study by making the train-validation difference in a single fold small, although usually a penalty was not necessary.  The mean and standard deviations of nonzero coefficients are shown in Table~\ref{tab_error}.  The diagnoses associated with increased error in the cases (failure to predict AKI when it occurred) are assigned to patients who  were hospitalized for reasons not directly related to the kidney (substance abuse).  Conversely, the diagnoses associated with small errors are obviously associated with AKI (e.g., we see especially accurate predictions of high AKI risk in hospitalizations preceded by frequent AKI or CKD).  The diagnoses associated with increased error in the controls (failure to rule out AKI when it did not occur) were, expectedly, AKI, CKD, and anemia.  This is not as revelatory as the cases; GBC has learned that prior kidney disease is associated with future kidney disease, which is a well known phenomenon.  These may correspond to cases in which interventions occurred for high risk patients (the label flipping mentioned in Assumption (1)).  

There were no detected relationships to the error (all coefficients were 0) for different races (American Indian, Asian, Black, Black/American Indian, Declined, Other, Unknown, and White). This is very comforting, although it is difficult to make a general conclusion for the rare races (see Fig~\ref{cohort_dem} for frequencies).  Gender also, favorably, showed no relationship to error.  As shown in Table~\ref{tab_error}, increasing age leads to lower error in the cases and higher error in the controls.  Hence errors occur because predicted risk is sometimes too high in older patients when they are healthy and too low in younger patients when they are not.  Age is a particularly well-recorded variable; it is unclear what variable could be adjusted for to remove this bias, but it is likely that explicit stratification might be in order.  In this large sample, the healthy young simply overwhelm the high risk young and opposite for the older patients. A plot of error by age is shown in Supplement~\nameref{er_age} to complement the findings in Table~\ref{tab_error}.  It is likely that, at least in part, the correlation of the errors with features indicates slight underfitting; had higher capacity HP been permitted, these patterns might have been detected (at the risk of overfitting in other ways).  As described above, bias was prioritized above variance in order to avoid overfitting, but now this error analysis gives some insight into who might suffer from poor predictions as a result.  

\begin{table}[!ht]
\caption{{\bf Coefficients of features associated with error.} For diagnoses, features correspond to the count assigned in prior hospitalizations.  Note that age and diagnosis were fit in separate regressions despite being displayed in the same table.}
\tiny
\begin{tabular}{p{9cm}|p{3cm}}
\toprule
\textbf{Cases (AKI +)} \\
\toprule
{Lasso (+)} & Mean (95\% CI) \\
\midrule
``Non-present'' Dx                                                            &  0.0071 (0.0069, 0.0073) \\
Non-dependent abuse of drugs                                                  &   0.0030 (0.0028, 0.0032) \\
\end{tabular}
\\
\\
\begin{tabular}{p{9cm}|p{3cm}}
\toprule
{Lasso (-)} &            Mean (95\% CI) \\
\midrule
AKI                                              &  -0.0476 (-0.0479, -0.0473) \\
CKD                                              &  -0.0301 (-0.0304, -0.0297) \\
Other and unspecified anemias                    &  -0.0066 (-0.0068, -0.0063) \\
Convalescence and palliative care                &  -0.0039 (-0.0042, -0.0036) \\
Hypertensive chronic kidney disease              &  -0.0006 (-0.0008, -0.0004) \\
Heart failure                                    &  -0.0003 (-0.0003, -0.0002) \\
Cardiac dysrhythmias                             &  -0.0001 (-0.0002, -0.0001) \\

\midrule
Age &  -0.0271 (-0.0272, -0.0269) \\
\midrule
\textbf{Controls (AKI -)} \\
\toprule
{Lasso (+)} &            Mean (95\% CI) \\
\midrule
AKI                           &  0.0166 (0.0164, 0.0167) \\
CKD                           &  0.0112 (0.0111, 0.0113) \\
Other and unspecified anemias &  0.0014 (0.0014, 0.0015) \\
\bottomrule
Age &  0.0236 (0.0235, 0.0236) \\

\bottomrule
\end{tabular}

\label{tab_error}
\end{table}

Error and STD of predicted probability against utilization is shown in Fig~\ref{Nhosp_error}.  The average error for controls decreases with the utilization.  For cases, the pattern is not clear, but it also appears to decrease.  Hence predictions are better for patients with many hospitalizations.  STD however appears to increase with utilization for cases, unlike for controls.  Since this dataset is a time-window sample, high utilizers are overrepresented (recall that a patient with multiple hospitalizations appears multiple times in the dataset).  This is common in medical prediction problems (e.g.,~\cite{CroninVetStrat} had a final analysis cohort with roughly 1.6 million admissions generated by roughly 600,000 patients; a readmission study~\cite{Futoma_comp_readmis} had roughly 3.3 million admissions generated by 1.3 million patients).  We hypothesize that high utilizers have strong influence over parameters.  Consider two patients without AKI; one is hospitalized 10 times and each time merely visits the emergency department and another is hospitalized twice for heart failure exacerbation.  The patient with 10 hospitalizations generates 9 training examples while the one with heart failure exacerbation generates only one. The former will have much stronger influence over coefficients.  

The impact of each patient on the coefficient vector of HPLR1 is shown in Supplement~\nameref{Nhosp_coef}.  There are patients who are relatively high and relatively low utilizers who have substantial impact on the coefficients.  Since there are many low utilizers, perhaps there is greater probability that there is a very different patient that might influence coefficients more.  However, extreme influencers seem to be relatively high utilizers. 
  Although this optimizes hospitalization-level performance (a prediction error for the patient who generates 10 hospitalizations may lead to 10 errors, while an error for the patient who generates 1 will only result in one, all else being equal) this might not be fair.  In WGBC, we have downweighted hospitalizations from high utilizers such that the 9 training examples have a net effect on the coefficients equal to the 1 training example.  Making patient influence over coefficients more equal optimizes performance on the level of patients rather than hospitalizations, but we do not see a clear change in the utilization analysis.  Another option for future work might be mixed effects approaches.

\newcommand{\errbyut}{The mean and STD absolute error is shown as a function of the number of hospitalizations.  Patients were binned based on the number of hospitalizations in the dataset and then, over bins, the mean error and STD of the predictions were computed.  Stratification by outcome is performed since it was earlier established that the hospitalization:patient ratio is higher in cases than in controls.}

\begin{figure}[!ht]
\centering
\includegraphics[width=0.8\linewidth]{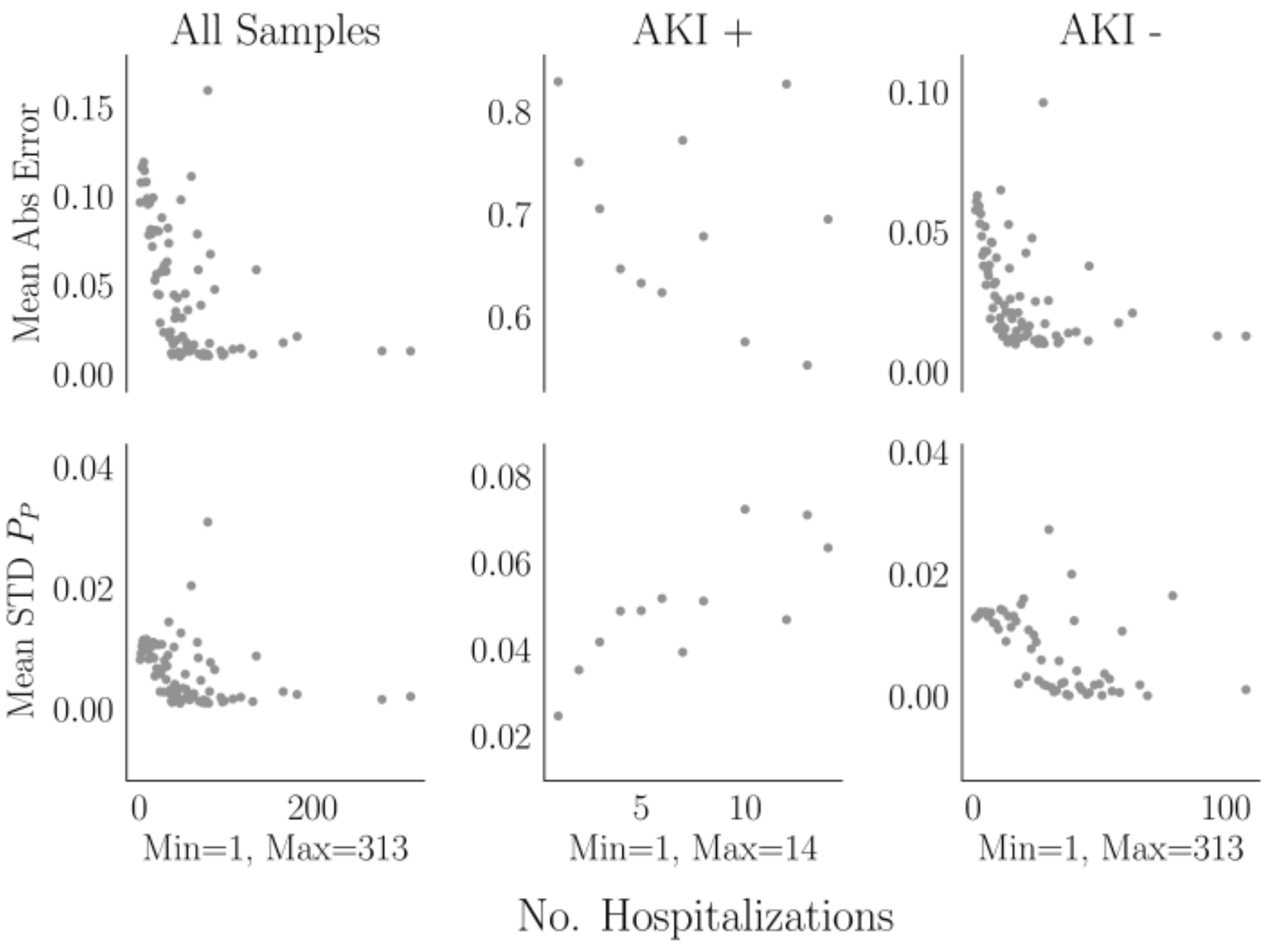}
\caption{\textbf{GBC error by utilization.} \errbyut
}

\label{Nhosp_error}
\end{figure}

\subsection*{Features}

Predictors specific to rehospitalized patients are described.  Note that these should be considered predictors, not risk factors, since causality is never established.  Further, many of the features are correlated, so it is important to note variance.  It is possible that some features in the ensuing tables might have correlated counterparts that could just as well have been selected in their places.  We still maintain that these tables are useful (1) to demonstrate that the systems depend on reasonable predictors and (2) to report potential predictors for specification of a more parsimonious system that could be validated on a different dataset.  Both of these objectives can be met despite the correlated nature of the features.  Also, note that the relationships between these features and AKI are associative, not necessarily causal.  The distribution of feature importances/coefficients was very skewed and we believe the interesting ones are adequately contained in the top 40, but this cutoff is still arbitrary. With these caveats in mind, we discuss some interesting findings. We reported 95\% bootstrap~\cite{efron1992bootstrap} confidence intervals (10,000 iterations using~\cite{bootstrapped}) instead of standard deviation as we had with the metrics because it was difficult to check each coefficient's distribution for symmetry.  Importance scores were computed via scikit-learn according to the Gini importance definition in~\cite{breiman2017classification}.  

\subsubsection*{GBC and LR1}

 The 40 features with the highest  micro-averaged GBC importance scores and largest absolute LR1 coefficients are shown in Table~\ref{tab_feat} and RLR1 coefficients are shown in Supplement~\nameref{RLR1_tab_feat}.  Some features were comprised of sub-features (e.g., diagnosis codes contained many sub-diagnoses).  For display in tables, these were succinctly renamed via a representative term (e.g., diuretics or CKD), most frequent sub-features, or general group names from~\cite{icd9com}.

\begin{table}[!ht]
\tiny
\caption{{\bf Feature importances/coefficients for GBC and LR1. }For laboratory results, the first function is $G$, aggregation over hospitalizations, and the second is $F$, aggregation within a hospitalization; e.g., ``mean max sCr'' is the mean over hospitalizations of the maximum sCr of each hospitalization. }
\begin{tabular}{p{9cm}|p{3cm}}
\toprule
{GBC} &            Mean (95\% CI) \\
\midrule
Age                                                               &  0.0409 (0.0404, 0.0414) \\
Mean count abnormally high urea nitrogen                          &  0.0351 (0.0338, 0.0365) \\
Count ``non-present'' DRGs                                        &  0.0295 (0.0289, 0.0301) \\
Count Dx: AKI                                                     &  0.0168 (0.0161, 0.0175) \\
Mean count abnormally low hemoglobin                              &   0.0160 (0.0152, 0.0168) \\
Mean sum urea nitrogen                                            &  0.0153 (0.0144, 0.0162) \\
Mean sum sCr                                                      &  0.0143 (0.0135, 0.0151) \\
Mean min direct bilirubin                                         &   0.0139 (0.013, 0.0149) \\
Mean max albumin                                                  &  0.0128 (0.0117, 0.0139) \\
Count immunosuppressant medications                               &  0.0124 (0.0116, 0.0132) \\
Max mean urea nitrogen                                            &  0.0118 (0.0109, 0.0127) \\
Mean count abnormally high sCr                                    &   0.0113 (0.0106, 0.012) \\
Count pharm subclass: Loop diuretics                              &   0.0104 (0.0098, 0.011) \\
Count pharm subclass: K-sparing diuretics                         &  0.0099 (0.0089, 0.0108) \\
Min min direct bilirubin                                          &  0.0098 (0.0088, 0.0107) \\
Mean count abnormal glomerular filtration rate-caucasian          &  0.0088 (0.0082, 0.0094) \\
Count ``non-present'' Dx                                          &   0.0084 (0.0078, 0.009) \\
Min mean chloride                                                 &   0.0080 (0.0074, 0.0087) \\
Count ``non-present'' CPT4 Px                                     &  0.0078 (0.0072, 0.0084) \\
Max max sCr                                                       &  0.0078 (0.0071, 0.0084) \\
Mean max urea nitrogen                                            &   0.0073 (0.0065, 0.008) \\
Mean mean hemoglobin                                              &  0.0071 (0.0065, 0.0077) \\
Count Px: injection of glucagon, haloperidol, heparin, enoxaparin &   0.0070 (0.0065, 0.0076) \\
Spironolactone                                                    &  0.0067 (0.0058, 0.0076) \\
Count discharges to Hospice/Medical Facility                      &  0.0066 (0.0062, 0.0071) \\
Count Dx: artificial opening status (e.g., tracheostomy)          &  0.0065 (0.0058, 0.0072) \\
Min max albumin                                                   &  0.0065 (0.0058, 0.0071) \\
Var count abnormally high urea nitrogen                           &  0.0057 (0.0049, 0.0065) \\
Count allopurinol                                                 &  0.0056 (0.0049, 0.0062) \\
Min min Glomerular filtration rate-Black         &  0.0055 (0.0049, 0.0061) \\
Count carbapenems                                &  0.0055 (0.0048, 0.0062) \\
Spinal procedures w/o CC/MCC                                      &  0.0053 (0.0044, 0.0061) \\
Sum max glomerular filtration rate-Black         &  0.0052 (0.0047, 0.0057) \\
Max min sCr                                      &  0.0052 (0.0047, 0.0058) \\
Count Dx: Disorders of fluid electrolyte and acid-base balance    &  0.0052 (0.0047, 0.0057) \\
Mean sum glomerular filtration rate-Black                         &  0.0052 (0.0045, 0.0059) \\
Count Dx: Diabetes Mellitus                                       &  0.0051 (0.0046, 0.0056) \\
Max max urea nitrogen                                        &  0.0051 (0.0045, 0.0058) \\
Max max hemoglobin                                                &  0.0049 (0.0043, 0.0055) \\
Sum max hemoglobin                                               &  0.0048 (0.0042, 0.0054) \\
\end{tabular}
\\
\\
\tiny
\begin{tabular}{p{9cm}|p{3cm}}
\toprule
{LR1 (+)} &            Mean (95\% CI) \\
\midrule
Age                                                          &  0.5846 (0.5825, 0.5867) \\
Mean mean urea nitrogen                                      &   0.1127 (0.108, 0.1175) \\
Count Dx: AKI                                                &  0.0974 (0.0956, 0.0993) \\
Mean max glucose                                             &  0.0905 (0.0882, 0.0927) \\
Mean mean sCr                                                &  0.0587 (0.0532, 0.0641) \\
Gender: Male                                                 &  0.0505 (0.0455, 0.0554) \\
Mean var glomerular filtration rate-Caucasian                &   0.0442 (0.0424, 0.046) \\
Max mean sCr                                                 &  0.0386 (0.0337, 0.0433) \\
Count discharges with home health organization care services &   0.0290 (0.0271, 0.0309) \\
Max mean urea nitrogen                                       &  0.0251 (0.0215, 0.0286) \\
Count Dx:  Chronic pulmonary heart disease                   &  0.0241 (0.0228, 0.0254) \\
Min min direct bilirubin                                     &  0.0241 (0.0221, 0.0261) \\
Mean min direct bilirubin                                    &  0.0227 (0.0208, 0.0245) \\
Count DRG: Hepatobiliary diagnostic procedures with MCC      &  0.0213 (0.0196, 0.0229) \\
Count Px: Pathology consult                                  &   0.0195 (0.0181, 0.021) \\
Count Px: Assay of blood lipoprotein or of magnesium         &   0.0186 (0.0161, 0.021) \\
Mean max urea nitrogen                                       &   0.0182 (0.015, 0.0213) \\
Count Px: Assay of urine sodium                              &  0.0179 (0.0163, 0.0194) \\
Min min sCr                                                  &  0.0173 (0.0144, 0.0201) \\
Last primary insurance: other                                &  0.0171 (0.0158, 0.0185) \\
\end{tabular}
\\
\\
\begin{tabular}{p{9cm}|p{3cm}}
\toprule
{LR1 (-)} &               Mean (95\% CI) \\
\midrule
Count ``non-present'' DRGs                        &  -0.2126 (-0.2294, -0.1964) \\
Mean min albumin                                  &  -0.2122 (-0.2171, -0.2076) \\
Mean min albumin                                  &  -0.0623 (-0.0671, -0.0575) \\
Gender: Female                                    &   -0.0610 (-0.0659, -0.0561) \\
Count Location: High risk labor and delivery unit &  -0.0585 (-0.0613, -0.0558) \\
Count location: Emergency Department              &  -0.0526 (-0.0606, -0.0443) \\
Min min glomerular filtration rate-Caucasian	  &  -0.0409 (-0.0446, -0.0373) \\
Mean min chloride                                 &  -0.0393 (-0.0418, -0.0369) \\
Count Dx: Traumatic injuries                      &   -0.0331 (-0.036, -0.0301) \\
Mean mean albumin                                 &  -0.0307 (-0.0347, -0.0263) \\
Count Admission on Tuesday                        &   -0.0288 (-0.033, -0.0246) \\
Count Admission on Saturday                       &  -0.0265 (-0.0307, -0.0221) \\
Count Dx: Injury from athletics                   &  -0.0252 (-0.0284, -0.0219) \\
Count discharges on Sunday                        &  -0.0202 (-0.0237, -0.0166) \\
Mean min glomerular filtration rate-Caucasian     &   -0.0195 (-0.0229, -0.016) \\
Max min hemoglobin                                &  -0.0166 (-0.0202, -0.0128) \\
Min min albumin                                   &  -0.0158 (-0.0187, -0.0129) \\
Max min bicarbonate                                &  -0.0109 (-0.0126, -0.0091) \\
Mean max albumin                             &  -0.0108 (-0.0128, -0.0088) \\
Mean min bicarbonate                              &  -0.0105 (-0.0122, -0.0087) \\
\bottomrule
\end{tabular}
\\
\\
\label{tab_feat}
\end{table}

For GBC, many features correspond to known indicators of acute or chronic kidney dysfunction (e.g. diagnosis of AKI, sCr, UN, GFR).  As our features are gleaned from prior hospitalizations, they suggest that prior acute or chronic kidney disease increases the probability of AKI.  Age is associated with declining kidney function in general, as well as a higher incidence of CKD and other conditions strongly associated with renal disease.  Thus it is not surprising that age is the strongest predictor of AKI in both GBC and LR1.  Another constellation of highly ranked features carries strong secondary association with underlying kidney disease.  These include medications used to treat consequences of decreased kidney function such as allopurinol, used to treat elevated uric acid levels, and loop diuretics, used to reduce fluid retention, edema, and hypertension. Highly ranked features associated with the presence of liver disease (bilirubin) and associated treatment for both liver and heart disease (spironolactone) were also identified.  Moderate to advanced liver and heart disease are associated with hepatorenal and cardiorenal syndromes, respectively, with resulting AKI (we even see hepatobiliary diagnostic procedures associated with increased risk in LR1).  Hemoglobin is also identified,  likely as an indicator of anemia resulting from renal pathology.  Interestingly, UN is often slightly preferred to sCr here, perhaps reflecting loss of muscle mass due to catabolism during illness, with an associated lower creatinine production blunting rise in sCr.  UN is generally correlated with sCr, probably explaining the high STD in the importances of both. 

The LR1 coefficients reveal the sign of predictors.  A number of features were associated with lower probability of AKI by LR1, especially those generally associated with populations having a lower incidence of kidney disease, including locations (labor and delivery, emergency department), and diagnoses (injuries from trauma and athletics).  Interestingly, although with small coefficients, timing of discharge and admission was identified as predictive.  For example, prior Sunday discharge was associated with a lower probability of AKI.  This may be due to the common practice in nursing homes and rehabilitation facilities to not accept weekend transfers, giving complicated patients with a higher likelihood of AKI lower probability of Sunday discharge.  In contrast, weekend hospital admissions (Saturday admission) have a higher number of traumatic injuries\cite{ERvisit2014} and thus a lower number of conditions associated with AKI (we see that diagnosis of traumatic injury is also present as a negative predictor).  In both GBC and LR1, ``non-present'' diagnoses and procedures were highly ranked since history of few diagnoses and procedures reflect robust health.

Although UN and sCr would likely have been chosen to predict AKI, many of the features studied here are novel representations.  For example, rather than just a recent UN, we include the number of abnormal lab flags for UN; rather than just an at-admission sCr, we include the mean over hospitalizations of the sum of sCr per hospitalization; rather than just presence of a loop diuretic on a medication list, we include the actual number of administrations.  Features of this form would probably not have been collected \textit{a priori} for AKI prediction, and their components are generally hidden to providers.  Many highly ranked features further depend on \textit{the behavior of providers}.  This might suggest that to optimize EHR data it is important to capture features that showcase provider behavior--such as testing or prescribing frequency.  Commonly used features such as ``does this patient have comorbidity X'' might be better reformulated as ``how many times in this patient's history has a provider assigned a code for comorbidity X''.  The features are further enhanced by EHR-based analyses (abnormal lab flagging).  

Interestingly, features associated with AKI in prior studies that analyzed only data available at admission were \textit{not} necessarily detected as the best predictors here in \textit{rehospitalized} patients.  For example, laboratory values dominated diagnosis codes, with the exception of diagnoses related to CKD or AKI.  We hypothesize that this may be due to our focus on longitudinal measurements, inclusion of more candidate features, the sparsity of ICD-9 codes, or perhaps correlation of diagnoses with laboratory predictors (the latter provide more predictive information, being continuous-valued and reliably collected variables).  Laboratory features may also have been boosted by the basis functions $F$, while the codes were generally just counted.

\subsubsection*{HPLR1}

All features with nonzero coefficients for HPLR1 are shown in Table~\ref{tab_pen} and the same for RHPLR1 is shown in Supplement~\nameref{tab_rpen}.  HPLR1 is especially interpretable.  UN has a large positive coefficient (note that there are two that are likely correlated and hence have high STD).  High glucose (endocrine or metabolic disorders) and potassium (renal dysfunction) are also predictive along with discharge with assisted care (Home Health Org.).  Negative coefficients are on mean \textit{minimum} hemoglobin, albumin, and calcium (all resounding laboratory indicators of strong health and robust kidney function).  Note that every positive laboratory coefficient contains a maximum and every negative a minimum.

\begin{table}[!ht]
\caption{{\bf Coefficients of HPLR1.} }
\tiny
\begin{tabular}{p{9cm}|p{3cm}}
\toprule
{HPLR1 (+)} &            Mean (95\% CI) \\
\midrule
Age                                                          &   0.2304 (0.229, 0.2318) \\
Max max urea nitrogen                                        &  0.1752 (0.1695, 0.1811) \\
Mean max urea nitrogen                                       &  0.1297 (0.1242, 0.1352) \\
Count Dx: AKI                                                &   0.0248 (0.0236, 0.026) \\
Mean max glucose                                           &    0.0001 (-0.0, 0.0002) \\
Mean max potassium                                            &    0.0001 (-0.0, 0.0002) \\
\end{tabular}
\\
\\
\begin{tabular}{p{9cm}|p{3cm}}
\toprule
{HPLR1 (-)} &               Mean (95\% CI) \\
\midrule
Mean min hemoglobin &  -0.0931 (-0.0949, -0.0912) \\
Mean min albumin    &  -0.0557 (-0.0573, -0.0541) \\
Mean min calcium    &      -0.0001 (-0.0002, 0.0) \\
\bottomrule
\end{tabular}
\label{tab_pen}
\end{table}

\subsection*{Comparison with features from Cronin et al.~\cite{CroninVetStrat}}
We can compare our features to those in Cronin et al.~\cite{CroninVetStrat}, where a random forest was used to predict AKI stage 1+ (KDIGO stages 1, 2, or 3).   
In Cronin et al, we see strong dependence on renal indicators (e.g., GFR, UN), labs indirectly associated with renal function (Hemoglobin), heart failure, diuretics (loop, thiazides), and anti-hypertensives such as angiotensin-converting enzyme inhibitors (ACEi), which is also reflected in our findings.  Although it is difficult to test rigorously, our study might suggest an opportunity to more extensively incorporate laboratory values from the past as predictors; Cronin et al. only used diagnoses and body mass index further than 365 days back and medications and temperature further than 90 days back.   

We can also compare our LR1 with lasso results from Cronin et al. High odds ratios in Cronin et al. were present in patients on antihypertensives (ACEi, angiotensin II receptor blockers, thiazides, $\beta$-blockers), diagnoses associated with AKI (diabetes, anemia, hyper and hypotension, peripheral vascular disease, HIV, cancer, and rheumatoid arthritis), labs associated with renal function or injury (calcium, hemoglobin, GFR, troponin, bilirubin), and antiobiotics (Sulfa). 
Again, we see many features associated with renal function, renal medications, sepsis, or cardiovascular dysfunction, which is also reflected in our findings.  In our features, but not in Cronin et al., we see discharge to home with outpatient care provided by a home health care organization (e.g. visiting nurse, home physical therapy, home health aide), lab values involving glucose, presence in the high risk labor and delivery unit or in the emergency department, injury from athletics, assay of urine sodium, discharge with organization care services, and marital status (possibly a proxy for age).
\subsubsection*{MGBC \& MLR1}

A substantial percentage of AKI is due to, or exacerbated by, medications.  We were thus interested in examining the medications in prior hospitalizations that might be associated with AKI in subsequent hospitalizations.  There were 927 medications analyzed.  The most important medication predictors are shown in Table~\ref{tab_med}.  Here again, the combination of GBC and LR1 results is useful to put the identified features in context.  Medications used to treat chronic obstructive pulmonary disease, such as albuterol and betamethasone, psychiatric conditions (respiridone, trazadone, aripiprazole), or obstetric therapies (magnesium, pre-natal vitamins) had a negative association with AKI.  Our aim was to detect potentially modifiable risk factors, but it is very difficult to disentangle confounders (e.g., Heparin might be associated with thrombotic event prophylaxis, dextrose with diabetic ketoacidosis and malnutrition).  Most medications associated with high risk were actually given to protect the kidney and most medications associated with low risk were given in the context of robust kidney health.  This analysis might be enhanced by somehow incorporating predictors from the current hospitalization.  We re-emphasize that no causal inference can be performed in this study, but interesting findings include tacrolimus (known nephrotoxicity~\cite{naesens2009calcineurin}), midazolam (this association has been shown relative to propofol~\cite{leite2015renal}), and oxycodone (opioid nephrotoxicity is currently researched~\cite{atici2004opioid}).  It is worth highlighting the counter-intuitive finding that ibuprofen administration in prior hospitalizations is a negative predictor for AKI.  Probably this is because non-steroidal anti-inflammatory medications (i.e. ibuprofen, ketorolac) are contraindicated in patients with elevated AKI risk, and thus administration during a prior hospitalization is a clinical indicator for low AKI risk.  However, patients with extensive histories of ibuprofen use, given its potentially deleterious effect on the kidney, should be monitored \textit{more closely} for AKI.  Here however we analyze \textit{administrations}, which, unlike use, reflect provider behavior.

\begin{table}[!ht]
\caption{{\bf Feature importances/coefficients for MGBC and MLR1.} Each feature corresponds to the count of administrations of the medication over prior hospitalizations. }
\tiny

\begin{tabular}{p{9cm}|p{3cm}}
\toprule
{MGBC} &            Mean (95\% CI) \\
\midrule
Furosemide                     &  0.0718 (0.0706, 0.0731) \\
Ibuprofen                      &   0.0356 (0.035, 0.0361) \\
Sodium Chloride                &  0.0306 (0.0299, 0.0313) \\
Allopurinol                    &   0.0245 (0.0239, 0.025) \\
Amlodipine Besylate            &  0.0237 (0.0232, 0.0242) \\
Oxycodone-Acetaminophen        &   0.0237 (0.023, 0.0243) \\
Spironolactone                 &   0.0232 (0.0224, 0.024) \\
Heparin Sodium                 &  0.0229 (0.0223, 0.0235) \\
Tacrolimus                     &  0.0217 (0.0212, 0.0223) \\
Enoxaparin Sodium              &   0.0210 (0.0203, 0.0217) \\
Torsemide                      &  0.0189 (0.0183, 0.0195) \\
Aspirin                        &  0.0187 (0.0179, 0.0195) \\
Fentanyl Citrate               &  0.0183 (0.0171, 0.0194) \\
Dextrose                       &   0.0180 (0.0175, 0.0186) \\
Levothyroxine Sodium           &  0.0154 (0.0148, 0.0161) \\
Piperacillin-Tazobactam In D   &   0.0145 (0.014, 0.0151) \\
Epoetin Alfa                   &  0.0144 (0.0139, 0.0149) \\
Carvedilol                     &  0.0133 (0.0125, 0.0141) \\
Hydralazine HCL                &  0.0128 (0.0121, 0.0135) \\
Sevelamer Carbonate            &  0.0119 (0.0111, 0.0126) \\
Metoprolol Tartrate            &  0.0113 (0.0107, 0.0118) \\
Docusate Sodium                &  0.0111 (0.0103, 0.0119) \\
Ceftriaxone Sodium In Dextrose  &  0.0104 (0.0097, 0.0111) \\
Pantoprazole Sodium             &  0.0103 (0.0095, 0.0111) \\
Metformin Hcl                   &  0.0102 (0.0096, 0.0109) \\
Albuterol Sulfate Hfa           &  0.0099 (0.0092, 0.0105) \\
Vancomycin Hcl In Dextrose	&  0.0099 (0.0092, 0.0105) \\
Nephro-Vite                     &   0.0090 (0.0084, 0.0096) \\
Magnesium Sulfate               &  0.0083 (0.0077, 0.0088) \\
Midazolam (Versed)              &  0.0083 (0.0076, 0.0091) \\
Glycopyrrolate                  &  0.0081 (0.0072, 0.0088) \\
Paricalcitol                    &  0.0071 (0.0062, 0.0081) \\
Cyclosporine Modified           &   0.0070 (0.0061, 0.0079) \\
Acetaminophen                   &   0.0070 (0.0063, 0.0077) \\
Benazepril HCL                 &   0.0068 (0.006, 0.0076) \\
Diltiazem HCL Er Beads         &  0.0067 (0.0057, 0.0076) \\
Labetalol HCL                  &   0.0067 (0.006, 0.0073) \\
Losartan Potassium             &  0.0064 (0.0058, 0.0071) \\
Warfarin Sodium                &   0.0064 (0.0058, 0.007) \\
Albumin Human                  &  0.0063 (0.0056, 0.0071) \\
\end{tabular}
\\
\\
\begin{tabular}{p{9cm}|p{3cm}}
\toprule
{MLR1 (+)} &            Mean (95\% CI) \\
\midrule
Furosemide                     &   0.1643 (0.161, 0.1677) \\
Heparin Sodium                 &  0.1349 (0.1325, 0.1372) \\
Allopurinol                    &  0.0947 (0.0912, 0.0982) \\
Enoxaparin Sodium              &  0.0883 (0.0857, 0.0911) \\
Piperacillin-Tazobactam In D   &   0.0809 (0.079, 0.0828) \\
Dextrose                       &  0.0785 (0.0725, 0.0844) \\
Tacrolimus                     &  0.0727 (0.0707, 0.0746) \\
Metoprolol Tartrate            &  0.0706 (0.0683, 0.0729) \\
Hydralazine HCL                &  0.0668 (0.0647, 0.0689) \\
Torsemide                      &  0.0609 (0.0583, 0.0637) \\
Glucagon HCL (Rdna)            &  0.0545 (0.0483, 0.0606) \\
Ceftriaxone Sodium In Dextrose &   0.0536 (0.0512, 0.056) \\
Epoetin Alfa                   &   0.0531 (0.051, 0.0553) \\
Spironolactone                 &    0.0530 (0.0501, 0.056) \\
Metoprolol Succinate           &    0.0500 (0.0483, 0.0517) \\
Sodium Chloride                &  0.0436 (0.0398, 0.0474) \\
Moxifloxacin HCL               &  0.0358 (0.0341, 0.0375) \\
Ciprofloxacin HCL              &  0.0347 (0.0326, 0.0368) \\
Fish Oil                       &  0.0284 (0.0269, 0.0299) \\
Oxycodone HCL                  &   0.0281 (0.026, 0.0303) \\
\end{tabular}
\\
\\
\begin{tabular}{p{9cm}|p{3cm}}
\toprule
{MLR1 (-)} &               Mean (95\% CI) \\
\midrule
Ibuprofen                      &  -0.2765 (-0.2807, -0.2722) \\
Oxycodone-Acetaminophen        &  -0.1575 (-0.1607, -0.1544) \\
Promethazine HCL               &  -0.0914 (-0.0949, -0.0879) \\
Ondansetron                    &   -0.0791 (-0.082, -0.0761) \\
Hydroxyzine Pamoate            &  -0.0707 (-0.0733, -0.0681) \\
Albuterol Sulfate Hfa          &  -0.0576 (-0.0609, -0.0543) \\
Nicotine Polacrilex            &  -0.0468 (-0.0501, -0.0434) \\
Tetanus-Diphth-Acell Pert      &  -0.0393 (-0.0415, -0.0372) \\
Cyclobenzaprine HCL            &  -0.0313 (-0.0335, -0.0292) \\
Classic Prenatal Vitamin              &  -0.0312 (-0.0334, -0.0289) \\
Trazodone HCL                  &    -0.0300 (-0.0331, -0.0268) \\
Oxytocin                       &  -0.0288 (-0.0309, -0.0267) \\
Risperidone Microspheres       &   -0.0264 (-0.0289, -0.024) \\
Risperidone                    &  -0.0231 (-0.0255, -0.0207) \\
Lorazepam (Ativan)             &  -0.0207 (-0.0227, -0.0186) \\
Ketorolac Tromethamine         &  -0.0207 (-0.0229, -0.0183) \\
Betamethasone Acetate \& Sodium Phosphate &  -0.0119 (-0.0134, -0.0103) \\
Prenavite Protein Coated       &  -0.0091 (-0.0116, -0.0064) \\
Aripiprazole                        &   -0.0067 (-0.008, -0.0053) \\
Etomidate                      &  -0.0065 (-0.0077, -0.0053) \\
\bottomrule
\end{tabular}

\label{tab_med}
\end{table}

\section*{Discussion}

In this study, we investigated the feasibility of using prior hospitalizations to estimate AKI risk at hospital re-entry.  The general objective was to extract and compress high-dimensional EHR information into a probability estimate specifically for rehospitalized patients.  Performance was assessed at the patient as well as hospitalization level.  Errors were also carefully analyzed to uncover gaps in predictive performance, with comprehensive analysis of diagnosis, race, gender, age, utilization, and method of AKI diagnosis.  Increasing the $l1$ penalty produced a parsimonious and interpretable HPLR1 whose features correspond to a striking physiological fingerprint for AKI risk.  Stability selection was performed to reinforce the results given the colinearity of features.  Other interesting predictors for AKI in rehospitalized patients were found, including medications, which may enhance specification of statistical AKI models and new investigations into modifiable risk factors.  While such predictive systems require extensive validation before clinical deployment, this work is a step toward creating automated AKI predictions, specifically for rehospitalized patients.   

With respect to generalizeability, we stress that we do not present a ``model'' for AKI, but instead a mapping from input features to AKI probabilities.  We reference a distinction made in Schmueli, et al.~\cite{shmueli2010explain} between \textit{explaining} and \textit{predicting}.  Here, we do the latter.  We also reference a distinction made in Breiman, et al.~\cite{Breiman_alg_v_mod} between \textit{models} and \textit{algorithms}.  Here, we use the latter.  An \textit{explanatory model} would require different methods, especially with regard to model specification and dependencies in the data.  We also recommend that parameters be retuned on different data for use elsewhere (``train locally'') as is commonly advised~\cite{Futoma_comp_readmis,CroninVetStrat,callahan2018machine}.  Thus, the systems presented here are only valid in the population from which the training data were sampled, and even there would require out-of-sample validation.

\subsubsection*{Comparison to other AKI and EHR prediction studies}
The state of the art in AKI prediction is  the work of Cronin, et al.~\cite{CroninVetStrat}. Direct comparison of performance with their models is challenging for several reasons.  First, they provide predictions at a different time.  We provide an at-entry risk score while Cronin et al. provides a risk score 48 hours \textit{post-admission}.   
We therefore use only features from prior hospitalizations while Cronin et al. uses features from the current hospitalization (from the 48 hours between admission and prediction time) as well as prior history.  Specifically, Cronin et al. used preadmission body mass index and preadmission diagnoses from -365 days to -24 hours and preadmission medications and temperatures from -90 days to -24 hours.  We did not have access to body mass index or temperature, and the feature engineering required to extract other variables such as medications was labor intensive, so even a comparison of our system with their pre-admission system was not possible.  Second, Cronin et al. focused specifically on hospital-acquired AKI while we focused on hospital and community acquired AKI.
Third, we analyzed different cohorts. In Cronin et al., since prediction was made at 48 hours, all hospitalizations with duration less than 48 hours were excluded (roughly 1.9 million hospitalizations). In contrast, our study, in which a prediction is made at hospital re-entry, applies to any patient regardless of length of stay.  We, however also excluded patients without prior hospitalizations (although we could give a prediction for these patients with no information by simply using the baseline prevalence of AKI). Therefore, in the space of all patients still present after 48 hours, the system in Cronin et al. is more general; in the space of all rehospitalized patients, our system is more general.  Also, in Cronin et al., data was from Veterans Affairs hospitals and included outpatient data; we only used inpatient data from a single hospital network, not just veterans.  Another similar study Kate et al.~\cite{Kate2016}, analyzed strictly patients 65 years of age and older, also making comparison difficult. 

Outside of AKI, the state of the art in EHR prediction has generally been achieved with RNN~\cite{rajkomar2018scalable,lipton2015learning,futoma2017learning} or variations~\cite{yoon2018deep}.  Here, we implemented an LSTM for sake of comparison.  The LSTM implemented here was not well optimized compared to those in other studies, so it did not outperform the other systems.  Nevertheless, LSTM has the clear advantage of reducing dependence on feature engineering.

\subsubsection*{Interpretability}

We do not recommend GBC, LR1, or LSTM for deployment because they are opaque.   These systems make the best predictions.  However, GBC, LSTM, and LR1 analyze thousands of features.  In principle, a user must understand and check each of these features in order to truly explain a prediction.  Otherwise, GBC or LR1 could infer that ibuprofen lowers AKI risk in an older patient with arthritis.  Or, given so many candidate predictors, GBC or LR1 might rely heavily on a feature whose relationship to the response is borne of pure chance \textit{throughout the dataset and undetectable by internal validation}~\cite{BengioReg}.  Some studies~\cite{sutherland2016utilizingAKIADQI} have  recommended that tools like GBC or LR1 only be used for feature discovery, and rather that a tool similar to HPLR1 be deployed, even with some reduction in predictive performance.  The user, on whom the onus falls to separate prediction from action~\cite{doi:10.1001/jama.2017.19198}, can more easily interpret HPLR1.  Using fewer features especially facilitates tracing an aberrant prediction back to, for example, a data entry error.  A parsimonious statistical model might even enable much needed closed-form expressions for prediction intervals (e.g., since prediction variance increases with risk).  Thus, insights from this study can be used for \textit{specification} of such a model. 

We note, however, that only taking into account a few features potentially results in a system that does not adjust for variables when it should.  Further, 
a human provider cannot analyze 1,500 features. Many of the features we analyze here are \textit{hidden} from the EHR user.  A learning algorithm that analyzes a large amount of---sometimes hidden---EHR data might thus be a useful complement.  However, we cannot ignore the benefits of parsimony, so recommend that both GBC (or LR1) and HPLR1 be used \textit{in concert} to give two separate risk scores.

\subsubsection*{Limitations \& future directions}
The major limitation of this study is difficulty in validating the assumptions outlined in the methods, especially the first assumption regarding interventions that flip labels.  Dependence on this assumption could be reduced by predicting sCr directly; since an intervening provider is responding to sCr, the algorithm could stay one step ahead, or by modifying the cost function to account for uncertainty in AKI status~\cite{christopher2016pattern}.  The last assumption is also difficult to validate and might lead to a system that \textit{favors} high utilizers~\cite{mullainathan2017does}.  These difficulties arise from the fact that EHR data is not collected explicitly for predictive modeling.  We also list some methodological limitations and future directions:  the HP search space and the HP themselves were not conceived of independently in each fold of nested CV, but instead set manually.  Bias was preferred to variance in choice of HP (and it was required that the test performance of the fold used to select HP not be optimistic relative to the other folds, a constraint much more easily fulfilled with higher bias HP).  By doing so, however, the data were slightly underfit, as evidenced by the error analysis, which essentially revealed undetected patterns.  This is especially apparent with respect to age.  We strongly suspect that had an ideal parameter search been achieved, or had HP that allowed higher variance been permitted, the GBC could have detected most of these patterns, and the error analysis would not have revealed such biases.  This, however, might have increased risk of overfitting.  At least bias is possible to detect (as we have done) whereas overfitting can be elusive.  Given the high number of predictors (especially relative to the cases), GBC and LR1 are likely overfit (not in the traditional sense, which can be detected via internal validation) but to peculiarities of the entire dataset, impossible to determine with internal validation alone.  We however, via domain-expertise-guided evaluation of features, consider this study to still contain insights of value to this prediction problem and cohort.  

Administrative codes are problematic predictors.  Although codes may be embedded or otherwise optimized as features~\cite{singhLeveHier,choi2016multi}, such approaches are not straightforward to implement in a pipeline.  Also, past AKI is a good predictor of future AKI.  Numerous reports suggest that codes have low sensitivity for AKI.  Therefore, using code-based AKI as a predictor is not ideal. AKI as a target was supplemented with sCr; AKI as a predictor was not supplemented with sCr, however, as this would necessitate extensive preprocessing of sCr trajectories in real time if deployed (time series models could take care of this for free, however).  For missing data imputation, more careful classification of missingness and more sophisticated methods such as matrix completion should be explored in the future. For laboratory values, Gaussian processes have also shown good performance~\cite{futoma2017learning}. 

Codes are also problematic as targets.  Although sCr-based diagnoses were used to supplement codes, we noted high discrepancy between the two.  Visual inspection suggests that sCr for hospitalizations diagnosed by code but not sCr usually began above normal and then decreased during the hospital stay, suggesting that an outpatient reading, or even a high initial measurement, prompted code assignment.  Without these cases, our findings align with previous reports that codes are specific but not sensitive for AKI. It was also apparent that errors were slightly higher in the cases diagnosed by sCr but not by code.  Another difficulty with diagnosis codes as labels is that they are often assigned at the end of the hospitalization and therefore not time stamped.  It is impossible therefore to know when the AKI occurred during the hospitalization (i.e., we do not distinguish between hospital- and community-acquired AKI).  On a similar note, because the majority of AKI codes were of ``unspecified'' severity, it was not possible to distinguish severities of AKI.  
This issue could be alleviated by predicting sCr directly in future work. Also relevant but not assessed is the performance of the systems as a function of time as analyzed in~\cite{calDavis}.  For example, certain medications might wane in popularity or diseases might be seasonal.  We hope to asssess this in the future and analyze the effect of online training.
\section*{Conclusion}

This study gives insight into the EHR-based AKI prediction problem in rehospitalized patients.  
Our objective was to investigate the feasibility of predicting AKI in this cohort as well as to analyze some interesting predictors.  We trained several learning algorithms and perform an in-depth error analysis, looking for specific patient groups for which predictions might be poor.  We also revealed novel predictors that could be used for specification of a statistical model.  We further focused on pharmaceutical predictors that may be worth further exploration as modifiable risk factors.  
We consider this work a step towards an automated, locally-trained tool that leverages sometimes hidden, longitudinal EHR data to estimate AKI risk in rehospitalized patients without manual ordering of tests, data collection, or data entry.  Such an estimate could provide a prior probability at the time of hospital re-entry to be used by an admitting provider or another predictive algorithm.

\section*{Supporting Information}

\paragraph*{S1 File.}
\label{alg_spec} 
{\bf Algorithm specifications.}

GBC~\cite{gbcSK}: loss=deviance, learning rate=0.1, n estimators=100, 
subsample=1.0, criterion=friedman mse, min samples split=150,
min samples leaf=100, min weight fraction leaf=0.0, 
max depth=2, min impurity split=1e-07, init=None, class weight=balanced
random state=random state, max features=None, 
verbose=0, max leaf nodes=None, 
warm start=False, presort=auto\\

LR1~\cite{LR1SK}: penalty=l1, dual=False, tol=0.0001, C=2e-3 (LR1) or 2e-4 (HPLR1), 
fit intercept=True, intercept scaling=1, class weight=balanced, 
random state=random state, solver=liblinear (uses coordinate descent), max iter=100, 
multi class=ovr, verbose=0, warm start=False, n jobs=5\\

Lasso~\cite{LassoSK}: alpha=for diagnoses cases 0.015, for controls 10e-5; for other stratifiers, 0, fit intercept=True, normalize=False, precompute=False, copy X=True, max iter=1000, tol=0.0001, warm start=False, positive=False, selection=cyclic

Randomized Logistic Regression~\cite{rlr}: C=0.5 for RLR1 and 0.2 for RHPLR1, sample fraction=0.74, n resampling=50 pipe to lasso with C=1, class weights=balanced

LSTM~\cite{keras}: optimizer=adam, epochs=4, batch size=500, layer 1 hidden units=30, layer 2 hidden units=20, dropout=random(0.25, 0.50, 0.75), iterations random search=3, score random search=log loss

\newpage
\paragraph*{S2 Fig.}
\label{met_dist}
{\bf Metric distributions. Metric distributions over the 250 inner folds are shown.}
\begin{figure}[!ht]
\centering
\includegraphics[width=\linewidth]{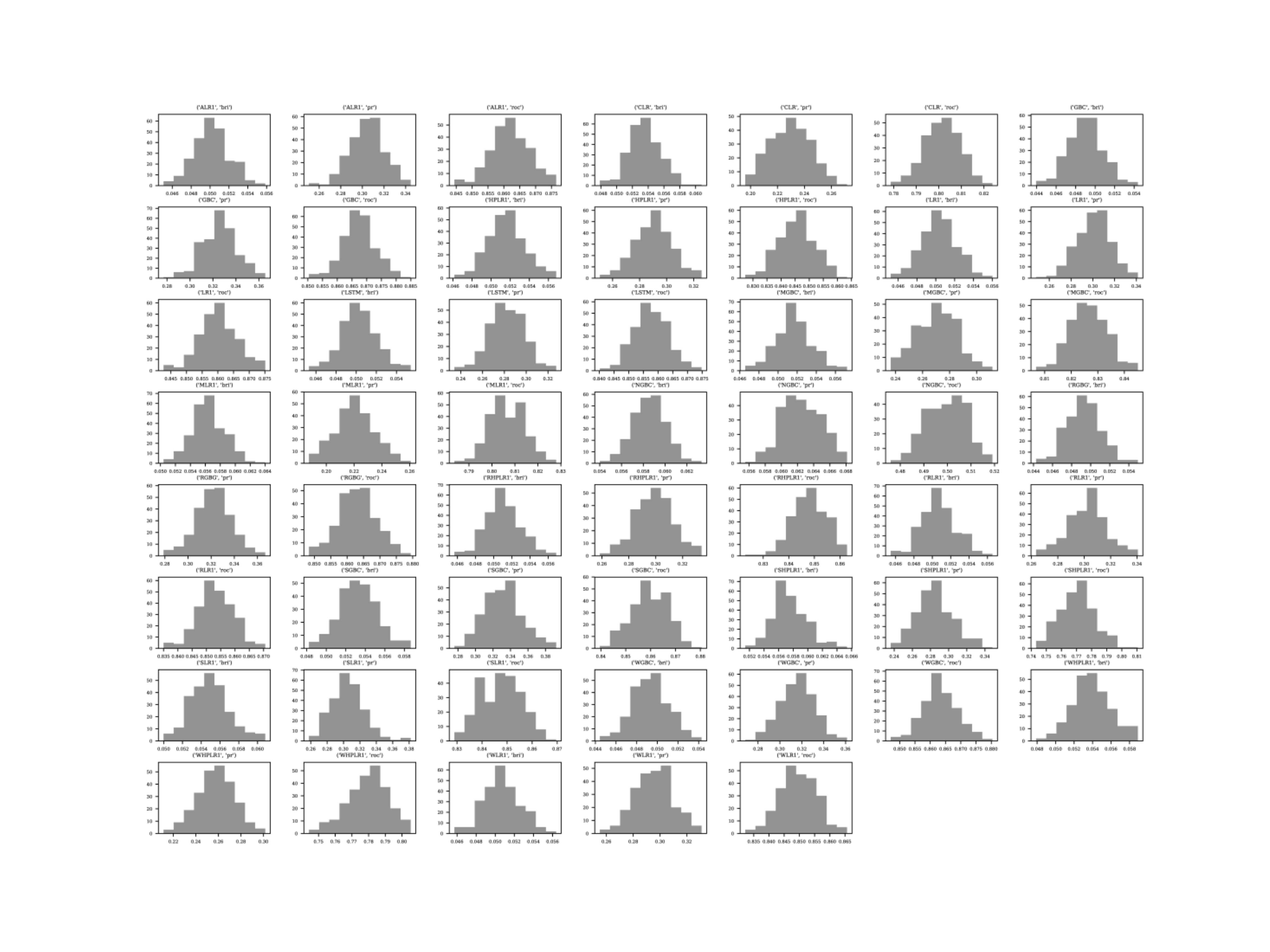}
\end{figure}

\paragraph*{S3 Fig.}
{\bf Error distributions by diagnosis method.} We show the distributions of error, $|\hat{y}-y|$ where $y$ is a binary label and $\hat{y}$ is the probability estimate, by diagnosis method. ``$\lor$'' corresponds to cases where diagnosis was made either by code or sCr; ``$\land$'' corresponds to cases in which diagnosis was made by both code and sCr; ``-'' indicates a set difference.  Histograms have 1000 bins.
\label{error_by_dx}
\begin{figure}[!ht]
\centering
  \includegraphics[width=0.8\linewidth]{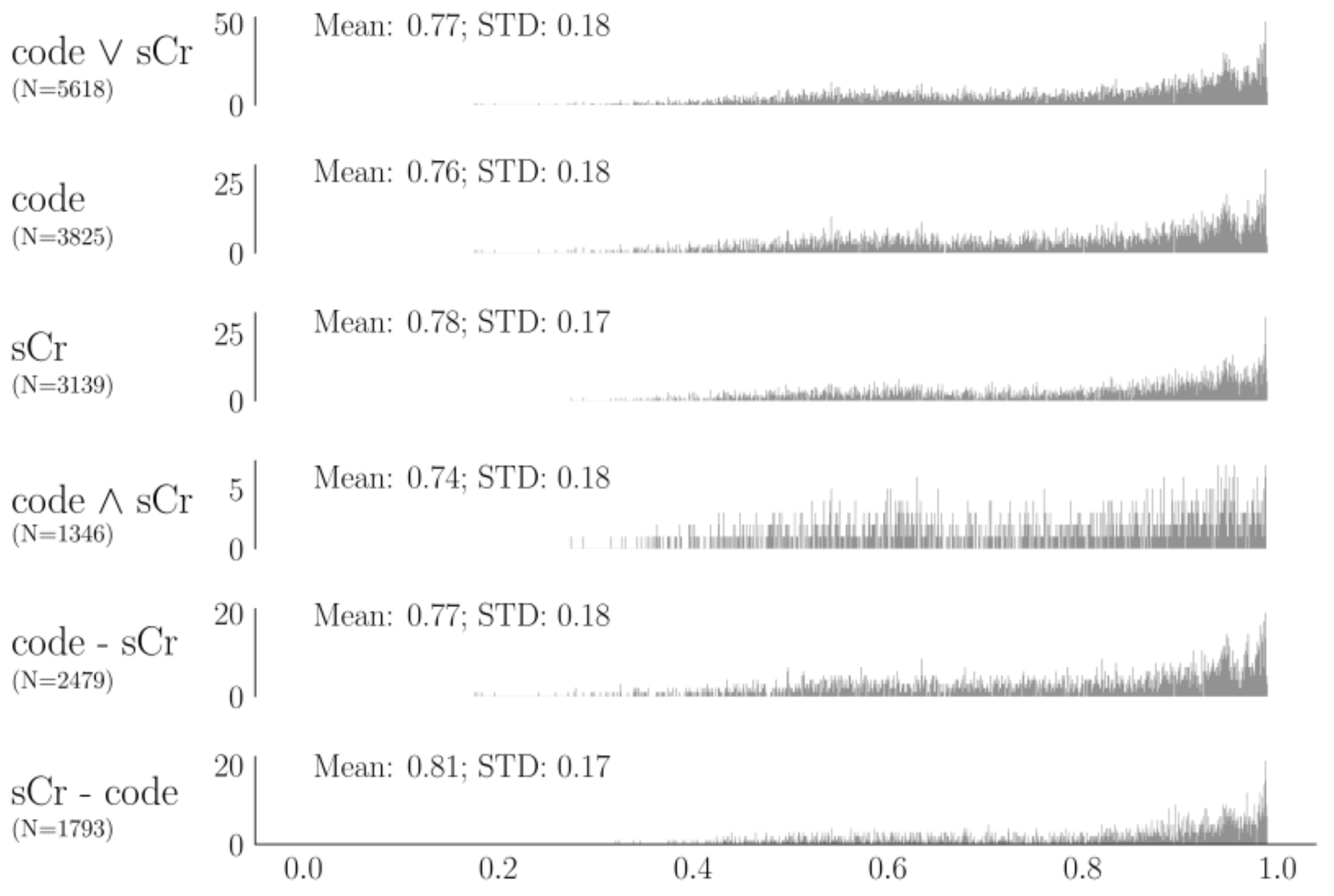}
\end{figure}
\newpage

\paragraph*{S4 Fig.}
\label{lr1_ev}
{\bf LR1 evaluation.} ROC, Calibration, and PR curves for 50 iterations of 5-fold CV for LR1.

\begin{figure}[!ht]
\includegraphics[width=\linewidth]{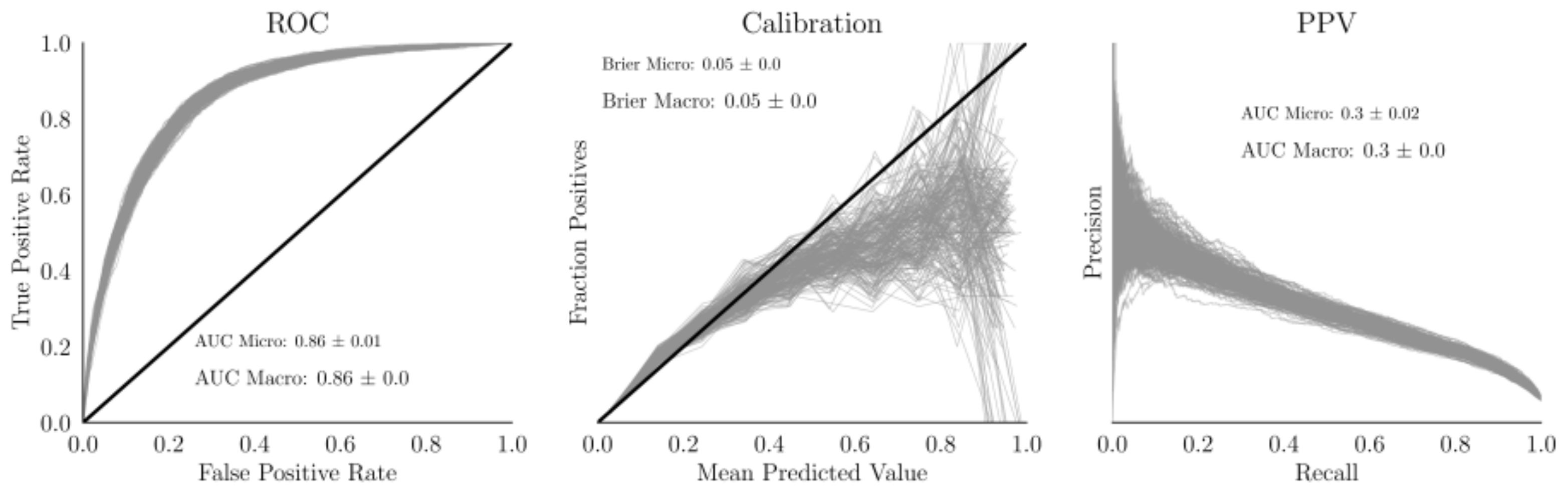}
\end{figure}

\paragraph*{S5 Fig.}
\label{alr1_ev}
{\bf ALR1 evaluation.} ROC, Calibration, and PR curves for 50 iterations of 5-fold CV for the Anscombe LR1.

\begin{figure}[!ht]
\includegraphics[width=\linewidth]{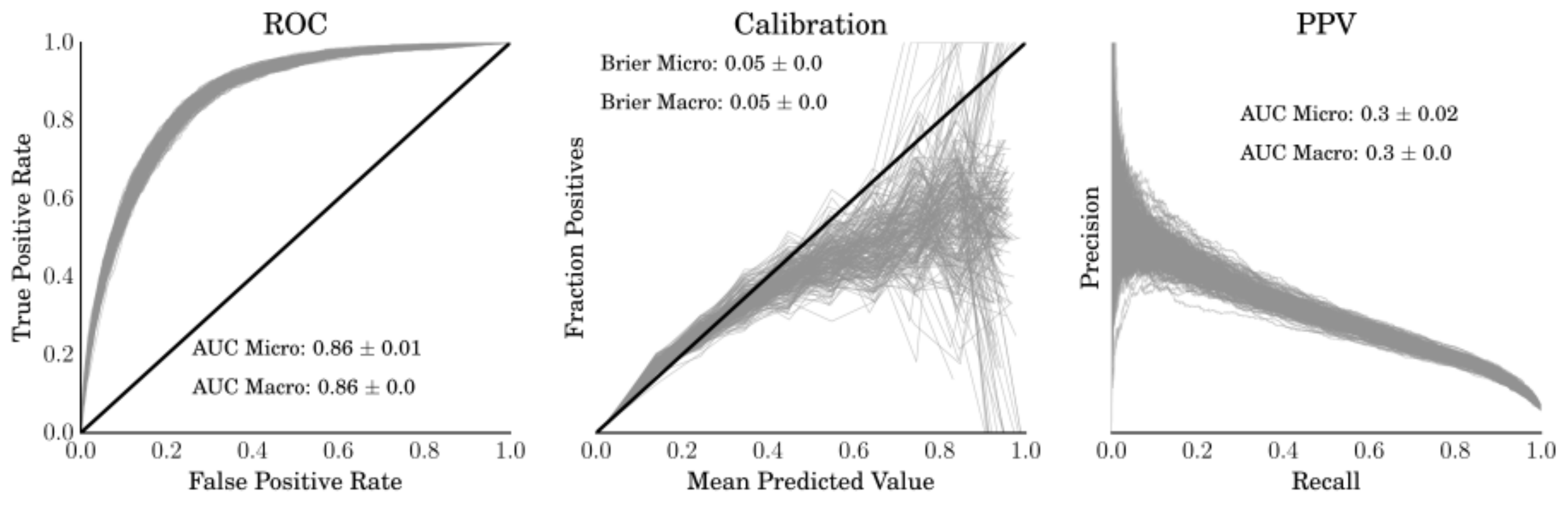}
\end{figure}

\paragraph*{S6 Fig.}
\label{rlr1_ev}
{\bf RLR1 evaluation.} ROC, Calibration, and PR curves for 50 iterations of 5-fold CV for the randomized LR1.

\begin{figure}[!ht]
\includegraphics[width=\linewidth]{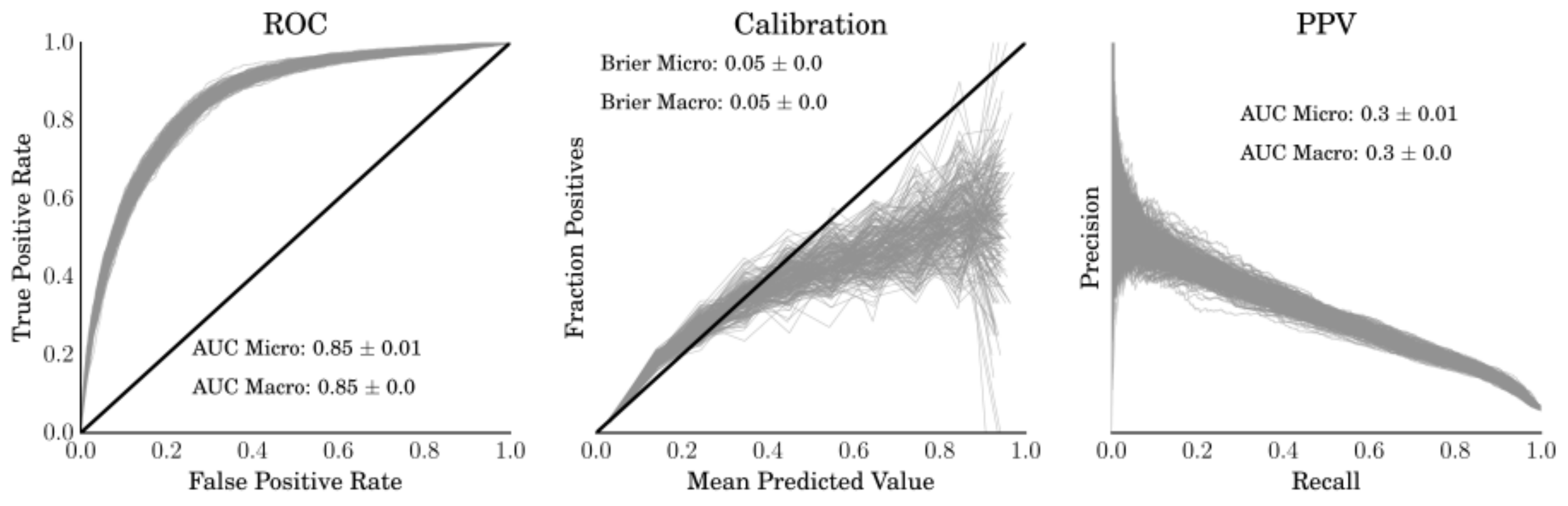}
\end{figure}

\newpage

\paragraph*{S7 Fig.}
\label{lr1_pen}
{\bf HPLR1 evaluation.} ROC, Calibration, and PR curves for 50 iterations of 5-fold CV for the highly penalized LR1.

\begin{figure}[!ht]
\centering
\includegraphics[width=\linewidth]{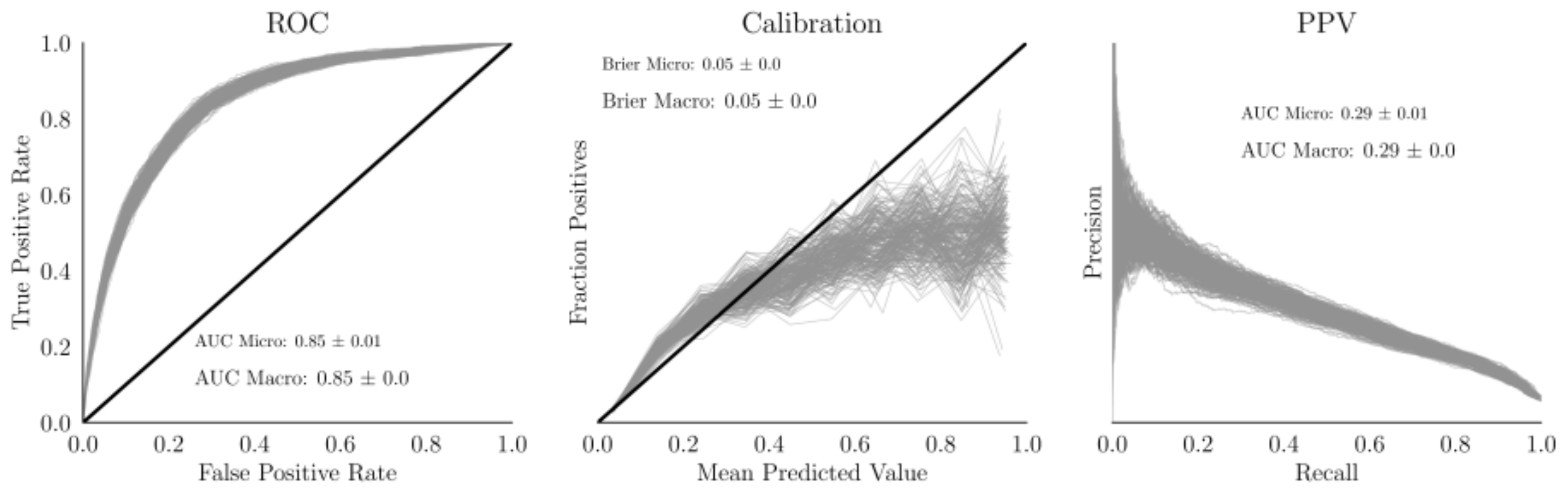}
\end{figure}

\paragraph*{S8 Fig.}
\label{rlr1_pen}
{\bf RHPLR1 evaluation.} ROC, Calibration, and PR curves for 50 iterations of 5-fold CV for the randomized highly penalized LR1.

\begin{figure}[!ht]
\centering
\includegraphics[width=\linewidth]{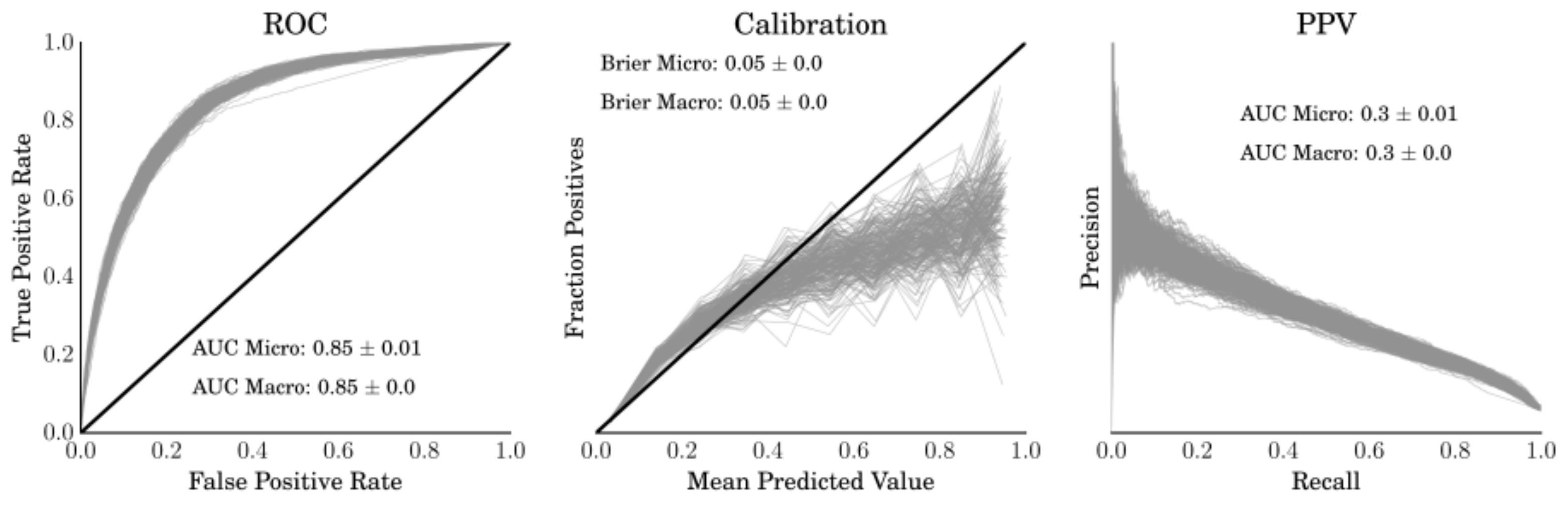}
\end{figure}

\paragraph*{S9 Fig.}
\label{per_wgbc}
{\bf WGBC evaluation.} ROC, Calibration, and PR curves for 50 iterations of 5-fold CV for weighted GBC.
\begin{figure}[!ht]
\centering
\includegraphics[width=\linewidth]{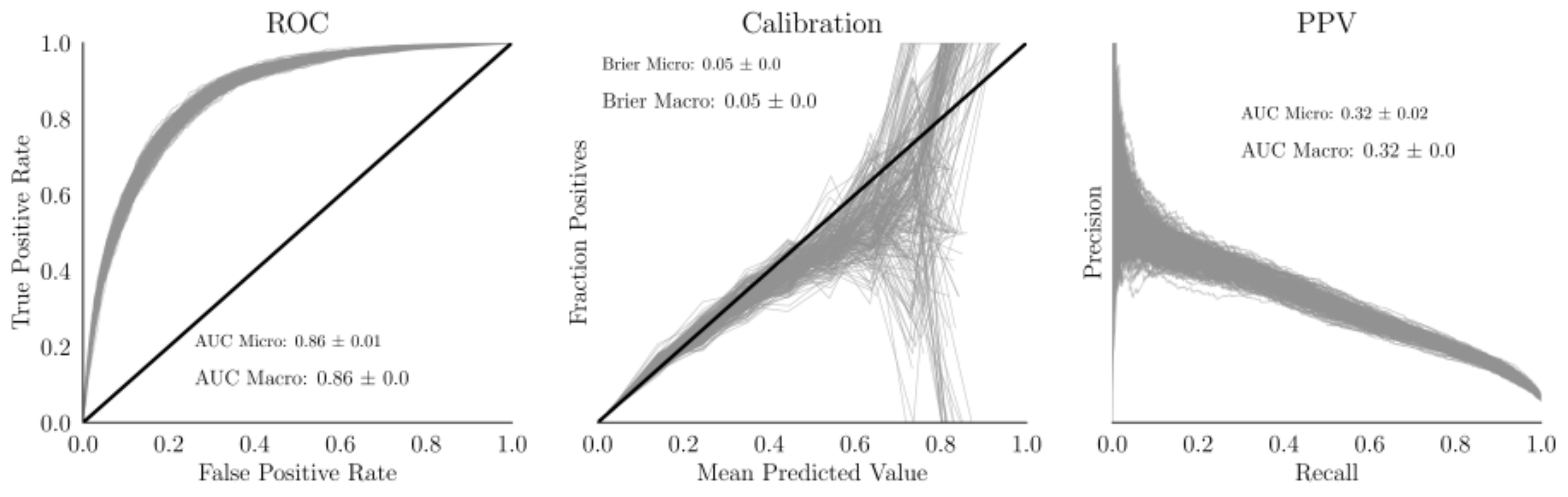}
\end{figure}

\newpage 

\paragraph*{S10 Fig.}
\label{cal_dist_WGBC}
{\bf WGBC calibration.} Observed hospitalization-level risk is plotted against predicted risk (top) and patient-level mean observed risk against mean predicted risk (bottom).  In the scatter plots, alpha level is 0.05 and the red calibration curve corresponds to all hospitalizations or to patients who had either mean risk over hospitalizations of 1 or 0.  The calibration curves are computed according to the macro-averaged predicted output per hospitalization or patient over the 50 iterations of 5 fold CV (over 250 total folds).  Ideal calibration is the dotted black diagonal.  $P_O$ = observed risk per hospitalization, $P_P$ = predicted risk per hospitalization, $\overline{P_O}$ = mean observed risk over hospitalizations, $\overline{P_P}$ = mean predicted risk over hospitalizations.
\begin{figure}[!ht]
\centering
\includegraphics[width=0.25\linewidth]{cal_dist_WGBC.pdf}

\end{figure}

\paragraph*{S11 Fig.}
\label{wgbc_ut}
{\bf WGBC utilization.} \errbyut
\begin{figure}[!ht]
\centering
\includegraphics[width=0.7\linewidth]{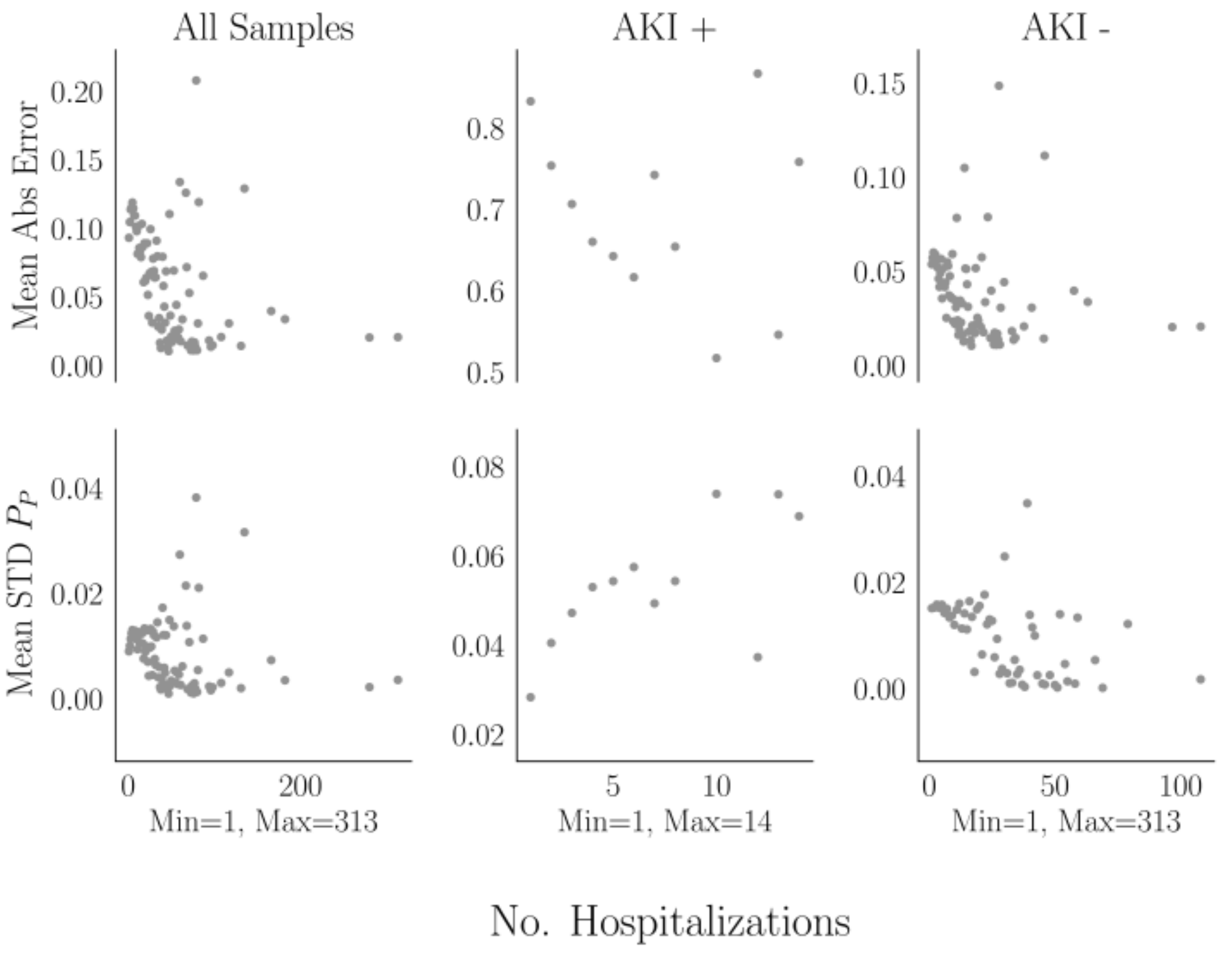}
\end{figure}

\newpage
\paragraph*{S12 Fig.}
\label{per_wlr1}
{\bf WLR1 evaluation.} ROC, Calibration, and PR curves for 50 iterations of 5-fold CV for weighted LR1.
\begin{figure}[!ht]
\centering
\includegraphics[width=\linewidth]{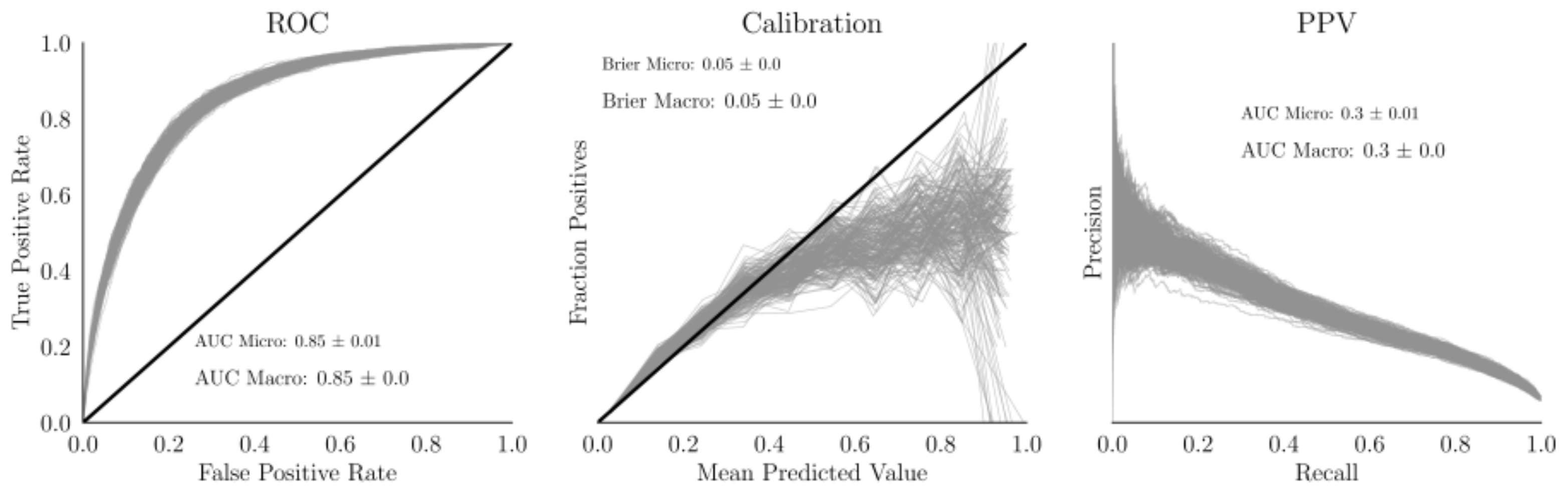}
\end{figure}

\paragraph*{S13 Fig.}
\label{per_whplr1}
{\bf WHPLR1 evaluation.} ROC, Calibration, and PR curves for 50 iterations of 5-fold CV for weighted HPLR1.
\begin{figure}[!ht]
\centering
\includegraphics[width=\linewidth]{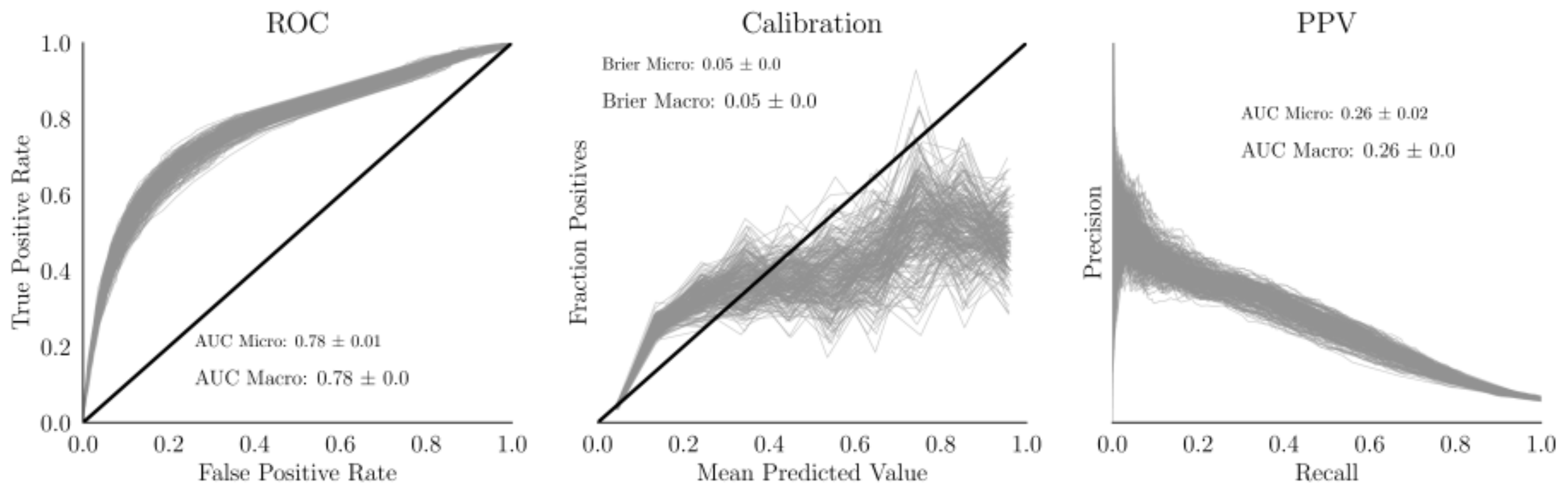}
\end{figure}

\paragraph*{S14 Fig.}
\label{per_sgbc}
{\bf SGBC evaluation.} ROC, Calibration, and PR curves for 50 iterations of 5-fold CV for sampled GBC.
\begin{figure}[!ht]
\centering
\includegraphics[width=\linewidth]{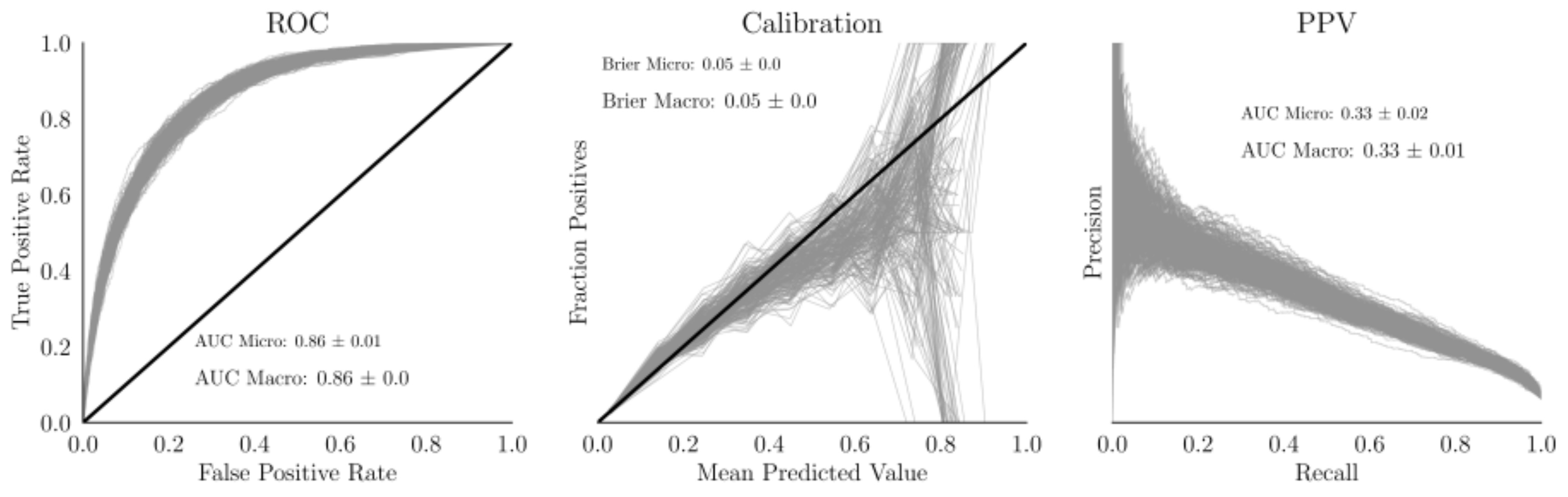}
\end{figure}

\newpage
\paragraph*{S15 Fig.}
\label{per_slr1}
{\bf SLR1 evaluation.} ROC, Calibration, and PR curves for 50 iterations of 5-fold CV for sampled LR1.
\begin{figure}[!ht]
\centering
\includegraphics[width=\linewidth]{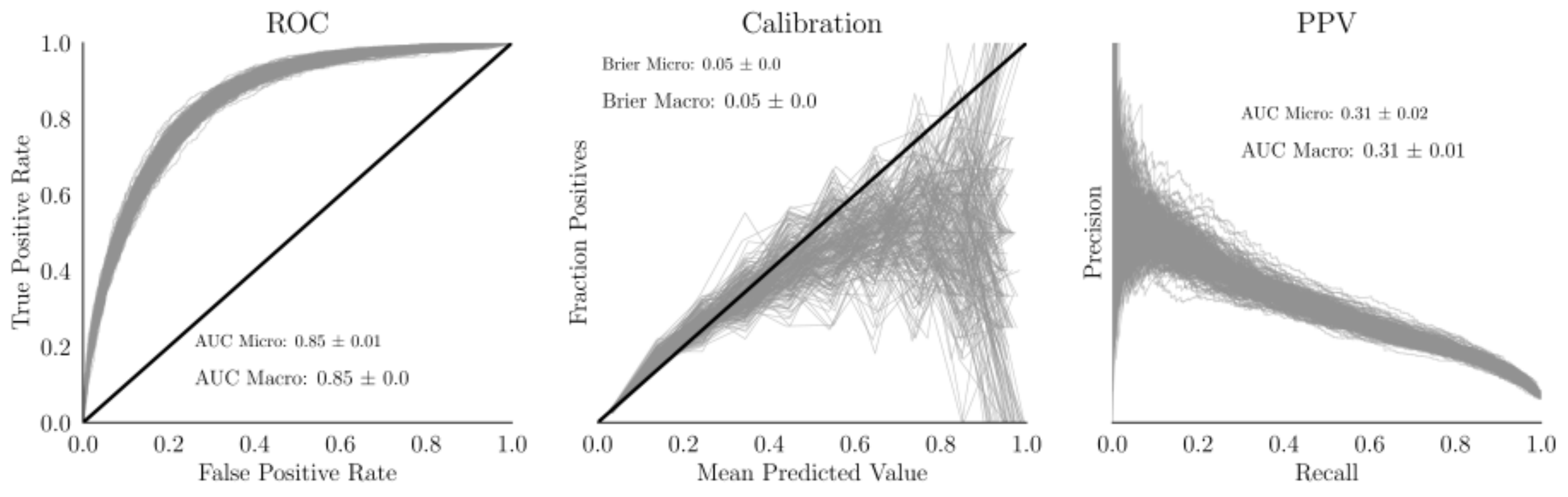}
\end{figure}

\paragraph*{S16 Fig.}
\label{per_shplr1}
{\bf SHPLR1 evaluation.} ROC, Calibration, and PR curves for 50 iterations of 5-fold CV for sampled HPLR1.
\begin{figure}[!ht]
\centering
\includegraphics[width=\linewidth]{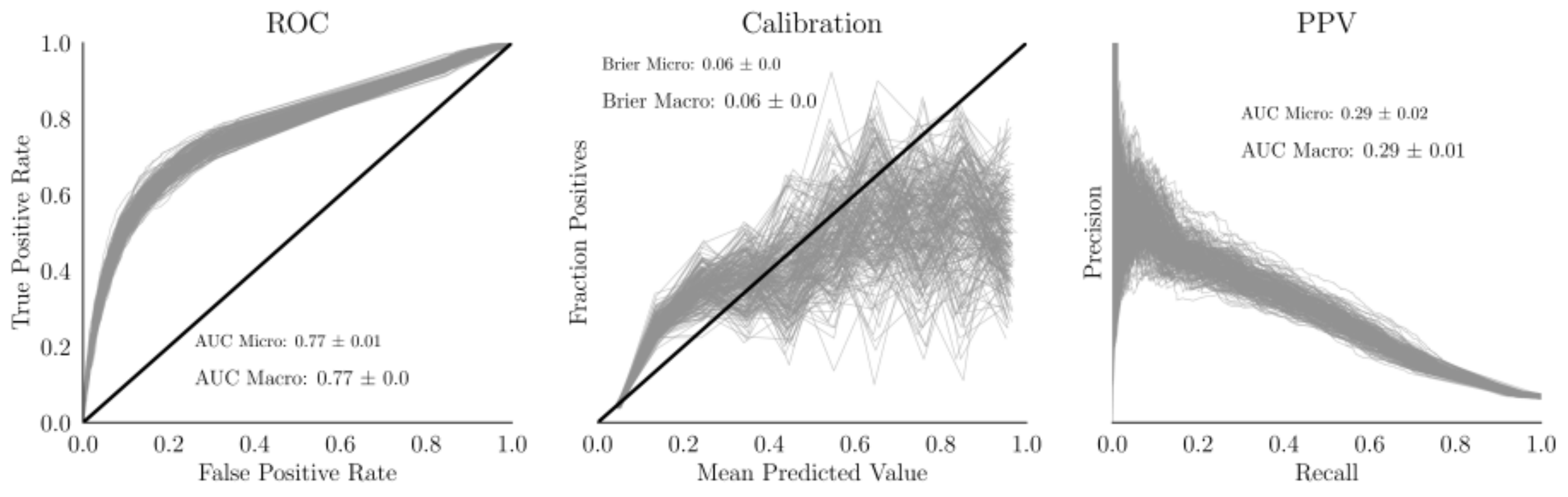}
\end{figure}

\paragraph*{S17 Fig.}
\label{memoryless}
{\bf RGBC evaluation.} ROC, Calibration, and PR curves for 50 iterations of 5-fold CV for the RGBC using features from only the most recent hospitalization rather than all available prior hospitalizations.
\begin{figure}[!ht]
\centering
\includegraphics[width=\linewidth]{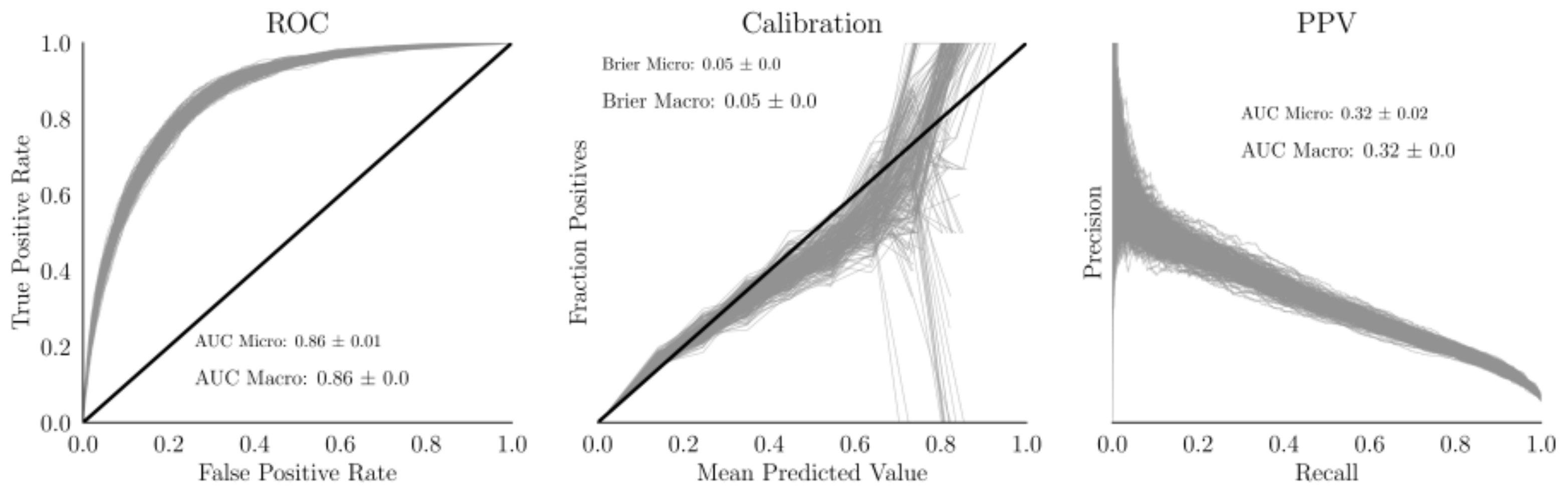}
\end{figure}

\newpage

\paragraph*{S18 Fig.}
\label{med_GBC}
{\bf MGBC evaluation.} ROC, Calibration, and PR curves for 50 iterations of 5-fold CV for the MGBC trained only on medications.
\begin{figure}[!ht]
\centering
\includegraphics[width=\linewidth]{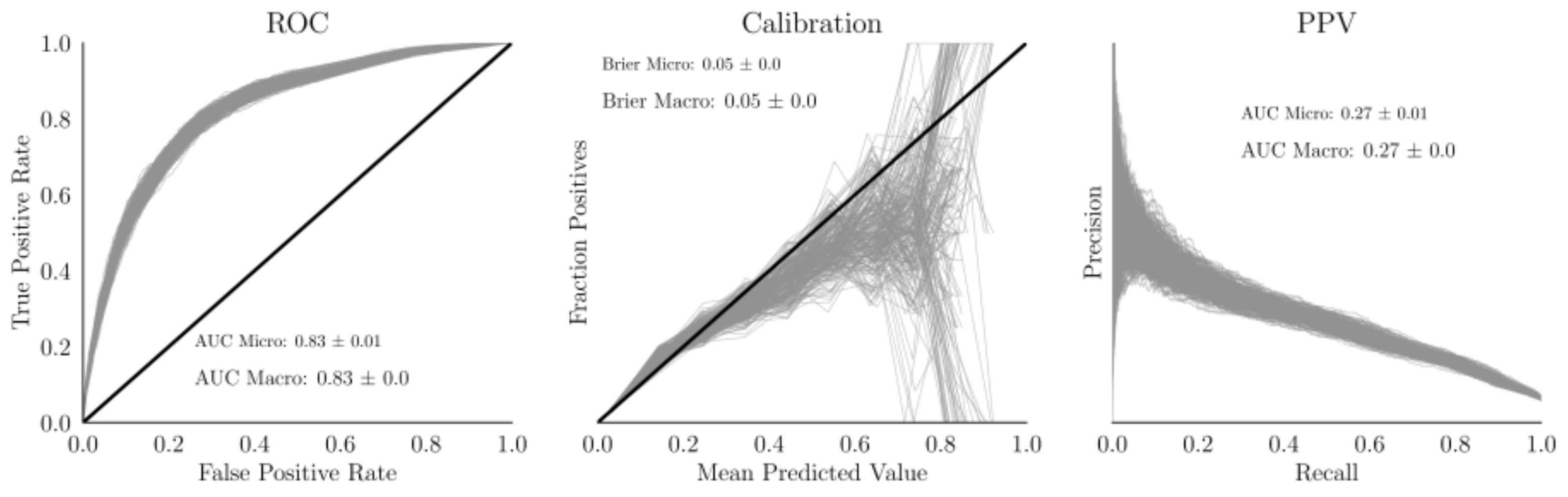}
\end{figure}

\paragraph*{S19 Fig.}
\label{med_LR1}
{\bf MLR1 evaluation.} ROC, Calibration, and PR curves for 50 iterations of 5-fold CV for the MLR1 trained only on medications.
\begin{figure}[!ht]
\centering
\includegraphics[width=\linewidth]{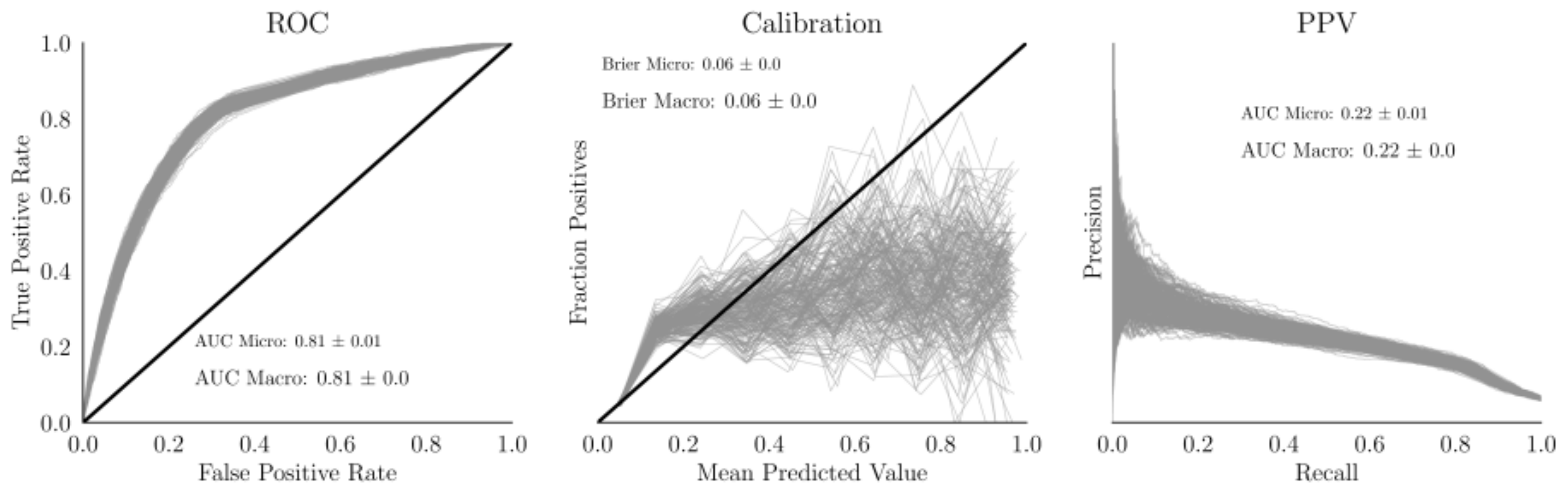}
\end{figure}

\paragraph*{S20 Fig.}
\label{per_CLR}
{\bf CLR evaluation.} ROC, Calibration, and PR curves for 50 iterations of 5-fold CV for clinical LR.
\begin{figure}[!ht]
\centering
\includegraphics[width=\linewidth]{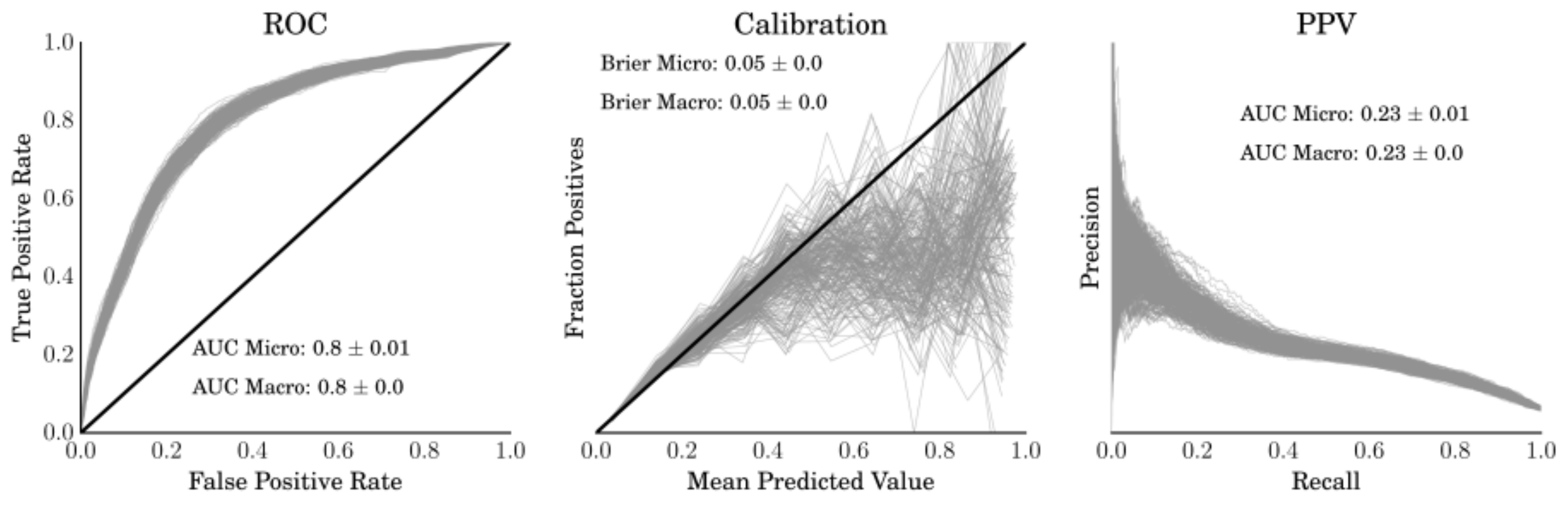}
\end{figure}

\newpage
\paragraph*{S21 Fig.}
\label{per_LSTM}
{\bf LSTM evaluation.} ROC, Calibration, and PR curves for 50 iterations of 5-fold CV for LSTM.
\begin{figure}[!ht]
\centering
\includegraphics[width=\linewidth]{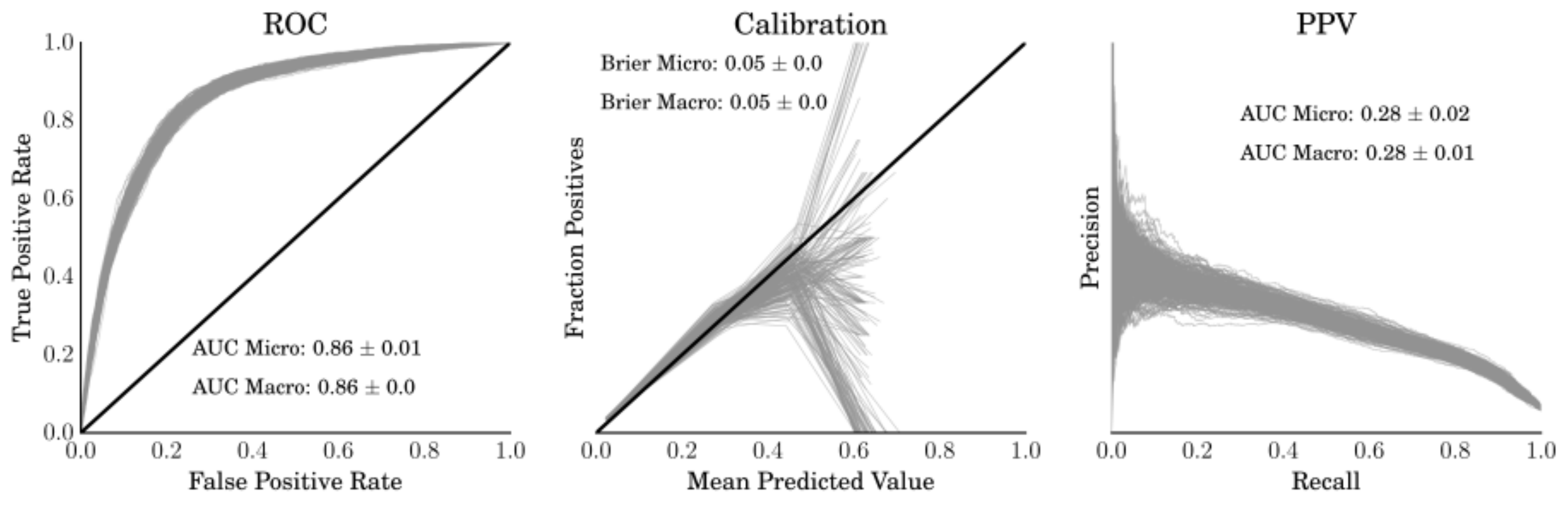}
\end{figure}

\paragraph*{S22 Fig.}
\label{n_GBC_per}
{\bf NGBC evaluation.} ROC, Calibration, and PR curves for 50 iterations of 5-fold CV for GBC trained on permuted response.  The identity for the calibration curve was hidden and the alpha value set to 1 for better visualization.
\begin{figure}[!ht]
\centering
\includegraphics[width=\linewidth]{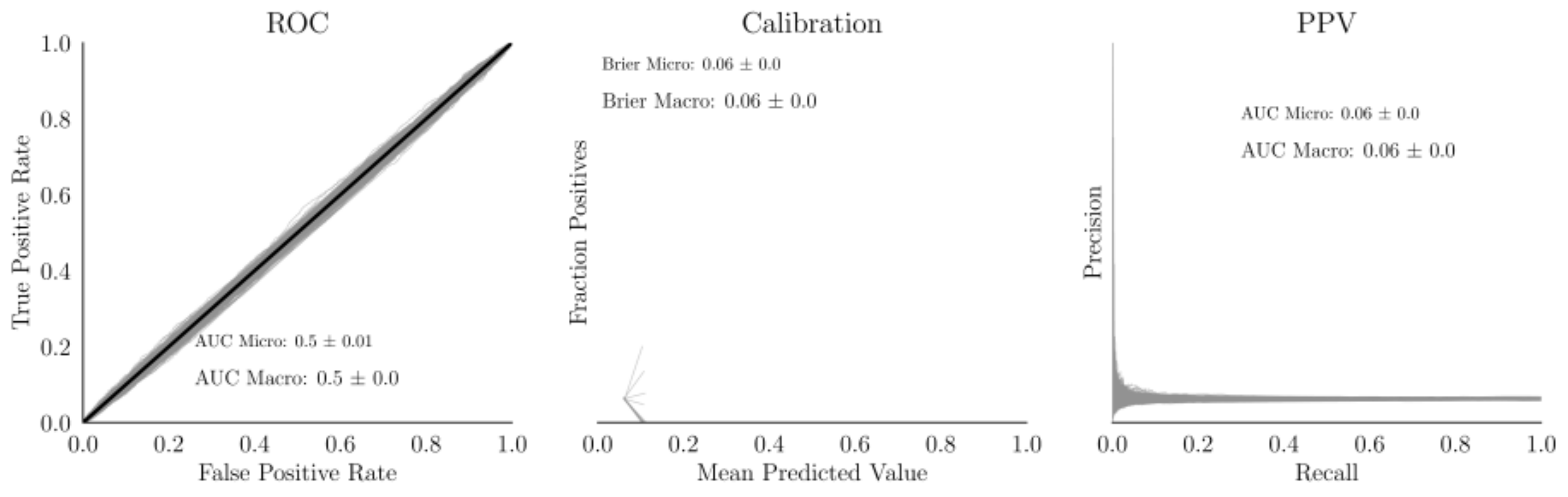}
\end{figure}

\newpage

\paragraph*{S23 Fig.}
\label{n_GBC_ut}
{\bf NGBC utilization.} \errbyut
\begin{figure}[!ht]
\centering
\includegraphics[width=0.8\linewidth]{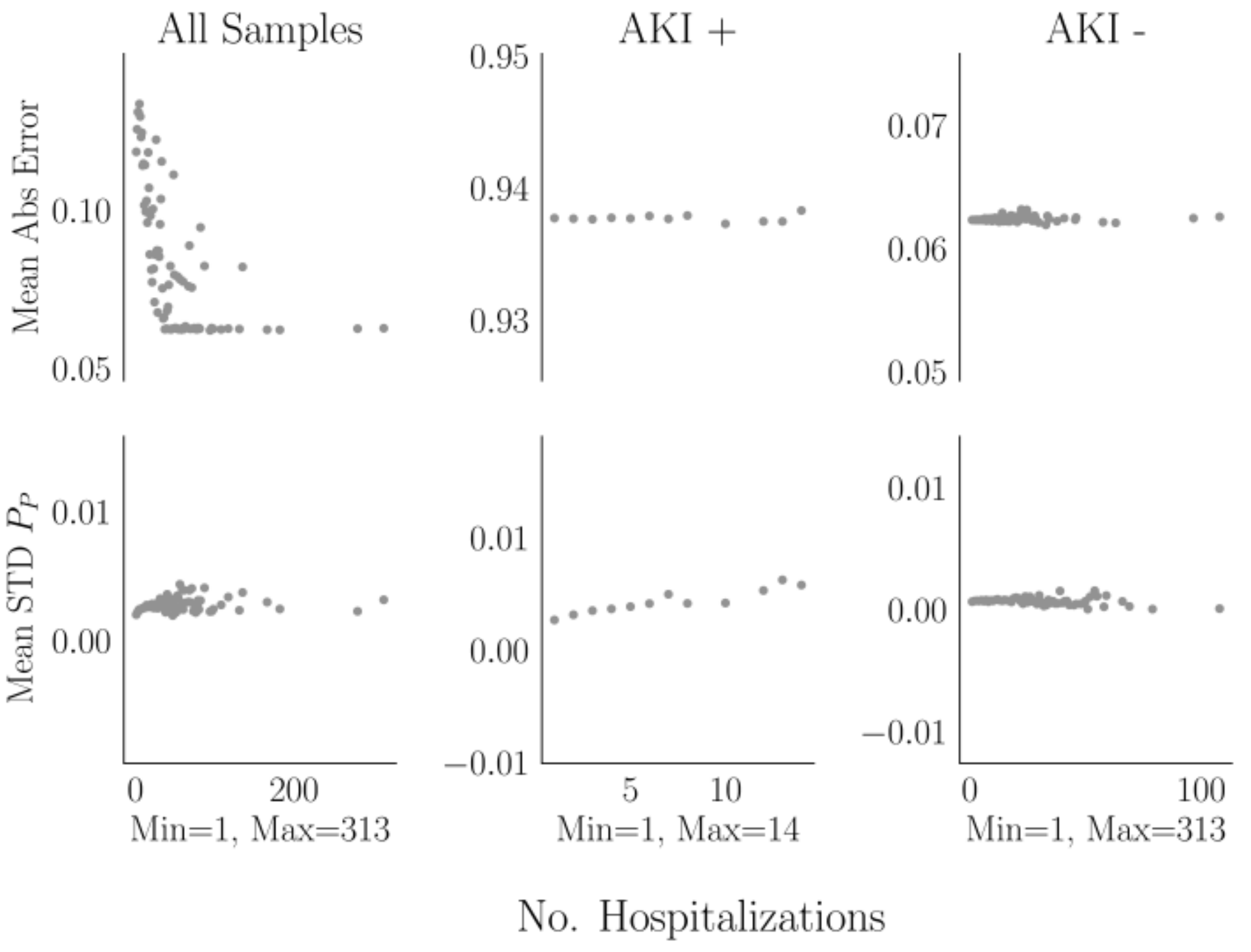}
\end{figure}
\newpage
\paragraph*{S24 Fig.}
\label{n_GBC_std}
{\bf NGBC prediction variance.} The mean and standard deviation of predicted probabilities are plotted over iterations (per hospitalization). Alpha=0.01 for all plots.
\begin{figure}[!ht]
\centering
\includegraphics[width=0.8\linewidth]{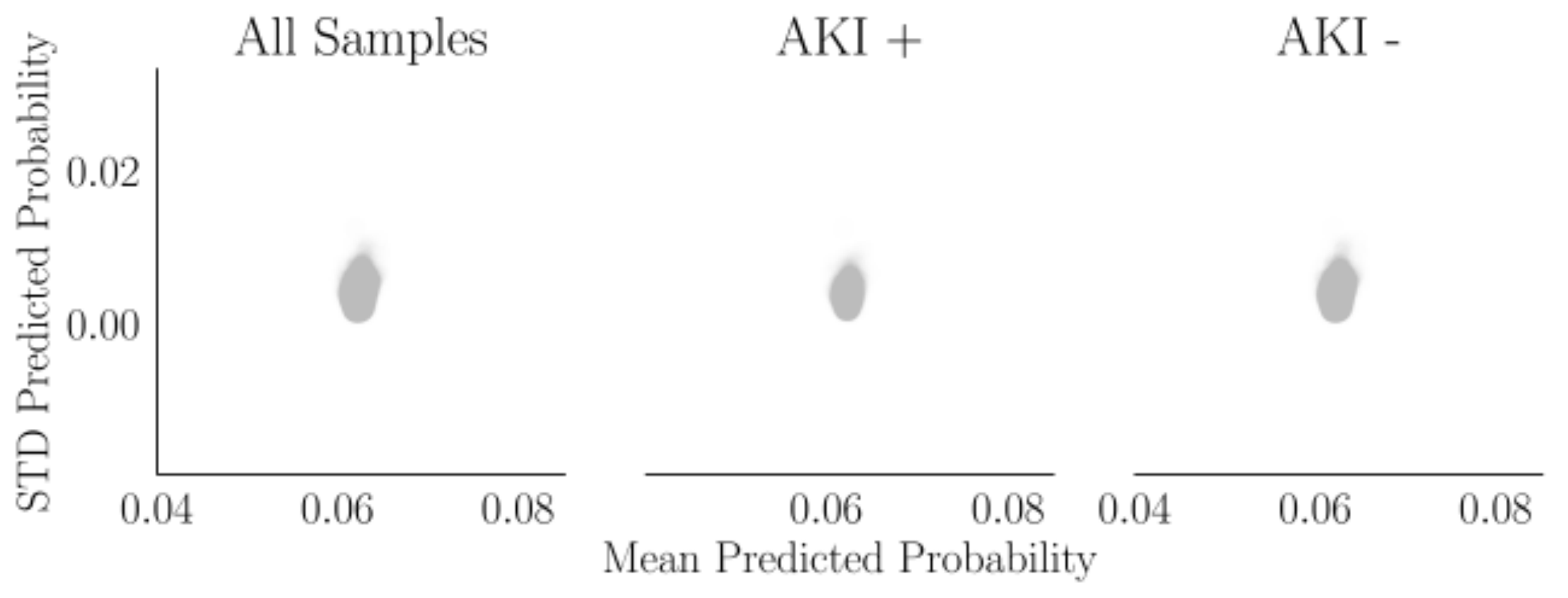}
\end{figure}

\newpage
\paragraph*{S25 Table.}
\label{error_train_val}
{\bf Regression results for error analysis.} Shown is the choice of HP Alpha and the train, validation, and test mean squared error (MSE) of the regression from the fold in which Alpha was chosen.
\begin{table}[!ht]
\tiny
\centering
\begin{tabular}{p{2cm}p{1cm}|p{1cm}p{1cm}p{1cm}p{1cm}p{1cm}}
\toprule
&                    &            & Diagnosis               & Race              & Gender               & Age\\
\midrule

AKI + (N=5,618)&Alpha&            & 0.015                   & 0                      &  0                        & 0\\                    
                    
&MSE			     & Train      & 0.021                   & 0.32                    &  0.032                    & 0.030\\
&                    & Validation & 0.022                   & 0.31                    &  0.031                    & 0.028\\
&                    & Test 	  & 0.021                   & 0.31                    &  0.031                    & 0.030\\

\midrule

AKI - (N=84,395)&Alpha&            & 1e-5                    & 0                       &  0                      & 0\\                       
&MSE                 & Train      & 0.005                   & 0.007                   &  0.007                  &0.006 \\
&                    & Validation & 0.005                   & 0.007                   &  0.007                  &0.006 \\
&                    & Test       & 0.005                   & 0.008                   &  0.008                  &0.006 \\

\bottomrule          
\end{tabular}
\end{table}

\paragraph*{S26 Fig.}
{\textbf{GBC error by age.}  Alpha=0.01. The top (in red, lighter) are the cases and the bottom (in blue, darker) are the controls.}
\label{er_age}
\begin{figure}[!ht]
\centering
 \includegraphics[width=0.4\linewidth]{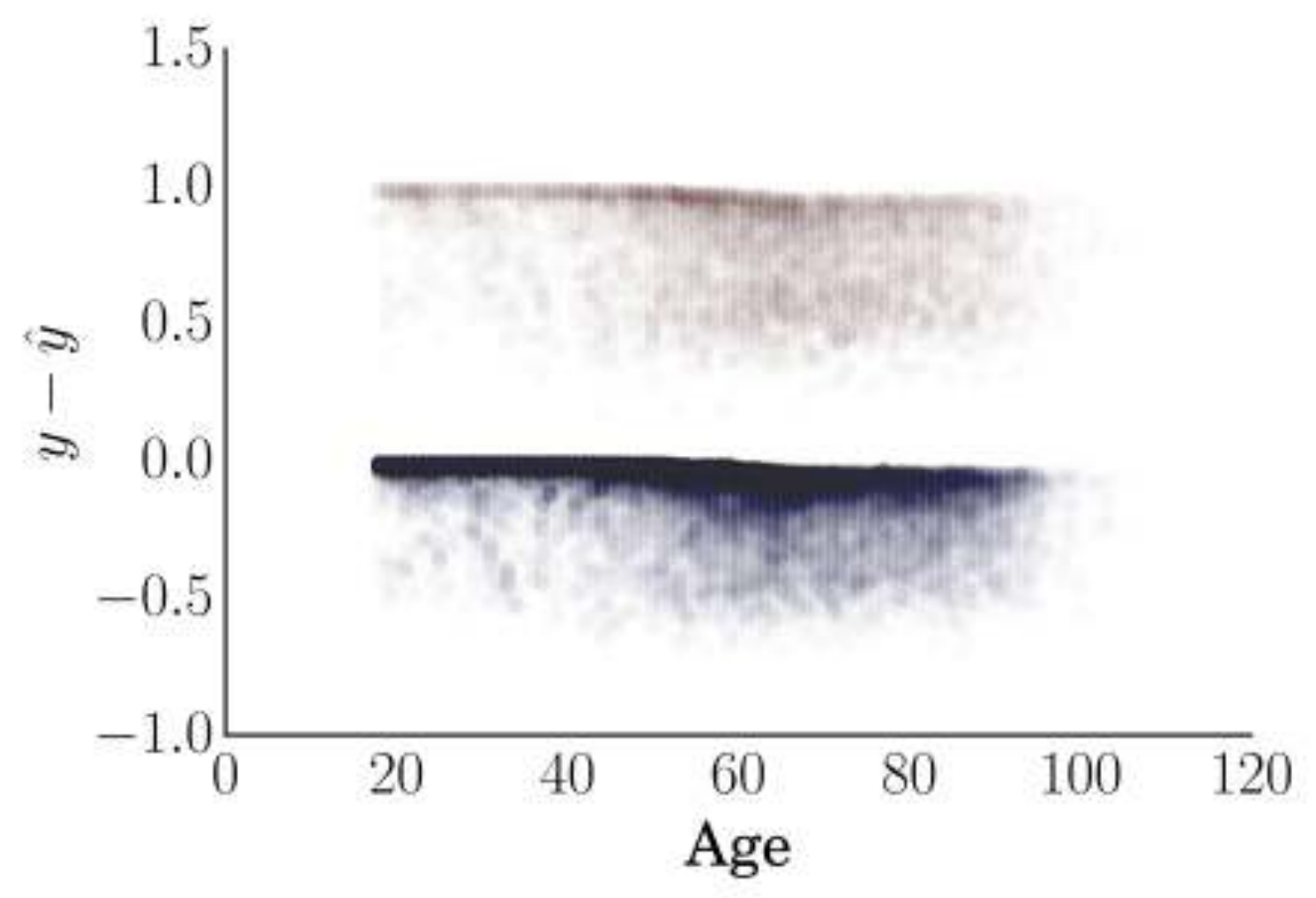}

\end{figure}

\paragraph*{S27 Fig.}
{\textbf{HPLR1 coefficient perturbation by utilization.}     Influence over coefficients of HPLR1 vs. utilization is shown for each patient with two or more hospitalizations.  Distance between coefficient vectors was computed using the l1 norm.}
\label{Nhosp_coef}
\begin{figure}[!ht]
\centering
 \includegraphics[width=0.6\linewidth]{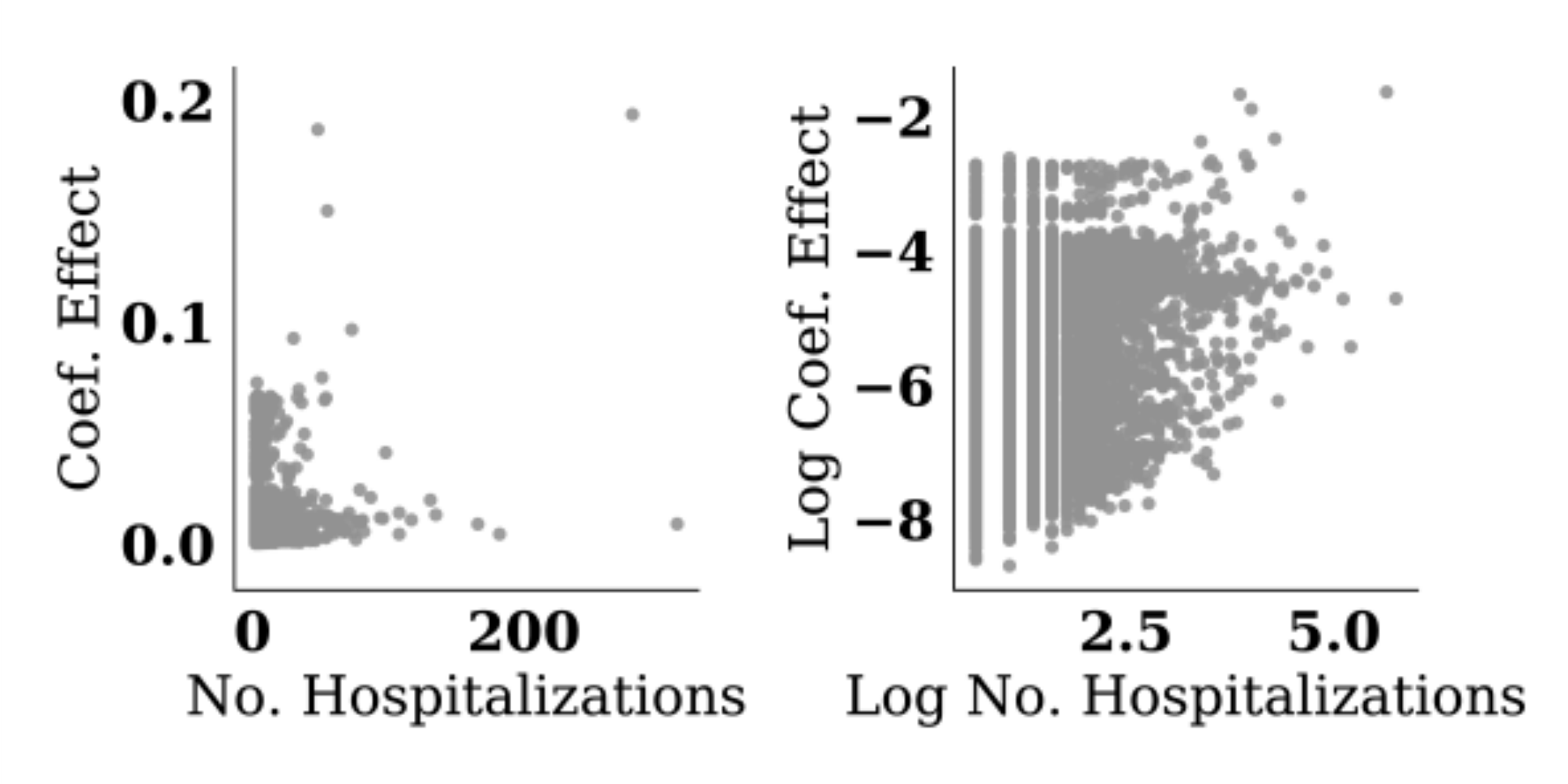}
\end{figure}

\newpage

\paragraph*{S28 Table.}
\label{RLR1_tab_feat}
{{\bf Feature importances/coefficients for RLR1. }For laboratory results, the first function is $G$, aggregation over hospitalizations, and the second is $F$, aggregation within a hospitalization; e.g., ``mean max sCr'' is the mean over hospitalizations of the maximum sCr of each hospitalization.}
\begin{table}[!ht]
\tiny

\begin{tabular}{p{9cm}|p{3cm}}
\toprule
{RLR1 (+)} &            Mean (95\% CI) \\
\midrule
Age                                                   &  0.3533 (0.3511, 0.3555) \\
Count Dx: AKI                                         &   0.083 (0.0816, 0.0845) \\
Count discharges with home organization care services &  0.0774 (0.0761, 0.0787) \\
Mean max glucose                                      &  0.0741 (0.0722, 0.0761) \\
Mean mean urea nitrogen                               &    0.0682 (0.0665, 0.07) \\
Mean max urea nitrogen                                &  0.0612 (0.0597, 0.0627) \\
Count Dx: CKD                                         &   0.0448 (0.043, 0.0467) \\
Mean max potassium                                    &  0.0429 (0.0415, 0.0443) \\
Mean min direct bilirubin                             &  0.0394 (0.0367, 0.0421) \\
Max mean urea nitrogen                                &  0.0381 (0.0371, 0.0391) \\
Count Px: Assay of urine sodium                       &  0.0369 (0.0352, 0.0387) \\
Mean max sCr                                          &  0.0325 (0.0299, 0.0352) \\
Max max urea nitrogen                                 &  0.0317 (0.0308, 0.0327) \\
Max mean sCr                                          &  0.0258 (0.0233, 0.0284) \\
Max max sCr                                           &  0.0253 (0.0237, 0.0268) \\
Min min direct bilirubin                              &  0.0231 (0.0201, 0.0262) \\
Mean mean sCr                                         &   0.0218 (0.0186, 0.025) \\
Min mean urea nitrogen                                &  0.0194 (0.0156, 0.0231) \\
Min max urea nitrogen                                 &   0.019 (0.0155, 0.0224) \\
Max max glucose                                       &    0.0177 (0.0155, 0.02) \\
\end{tabular}
\\
\\
\begin{tabular}{p{9cm}|p{3cm}}
\toprule
{RLR1 (-)} &               Mean (95\% CI) \\
\midrule
Mean min hemoglobin                           &  -0.0855 (-0.0882, -0.0826) \\
Mean min glomerular filtration rate-caucasian &  -0.0819 (-0.0846, -0.0793) \\
Max min hemoglobin                            &   -0.0737 (-0.0785, -0.069) \\
Mean min albumin                              &   -0.064 (-0.0653, -0.0627) \\
Mean mean albumin                             &  -0.0617 (-0.0631, -0.0603) \\
Mean min chloride                             &  -0.0579 (-0.0603, -0.0555) \\
Mean min calcium                              &   -0.0502 (-0.0545, -0.046) \\
Min min glomerular filtration rate-caucasian  &  -0.0463 (-0.0477, -0.0449) \\
Marital status: single                        &  -0.0341 (-0.0394, -0.0287) \\
Min min hemoglobin                            &  -0.0239 (-0.0265, -0.0212) \\
Min min albumin                               &  -0.0181 (-0.0204, -0.0156) \\
Min min chloride                              &  -0.0177 (-0.0197, -0.0157) \\
Mean mean hemoglobin                          &   -0.0096 (-0.012, -0.0071) \\
Sum count abnormally high sCr                 &  -0.0045 (-0.0054, -0.0036) \\
Max mean albumin                              &   -0.003 (-0.0045, -0.0013) \\
Max max potassium                             &  -0.0027 (-0.0035, -0.0019) \\
Mean max albumin                              &   -0.0022 (-0.0033, -0.001) \\
Max min calcium                               &  -0.0021 (-0.0035, -0.0004) \\
Min mean albumin                              &  -0.0019 (-0.0028, -0.0011) \\
Count ``non-present'' DRGs                    &   -0.0015 (-0.0031, 0.0008) \\
\bottomrule

\end{tabular} 
\\
\end{table}

\newpage
\paragraph*{S29 Table.}
{{\bf Coefficients of RHPLR1.} }
\label{tab_rpen}
\begin{table}[!ht]
\tiny
\begin{tabular}{p{9cm}|p{3cm}}
\toprule
{RHPLR1 (+)} &             Mean (95\% CI) \\
\midrule
Age                                                            &   0.3968 (0.3943, 0.4006) \\
Count Dx: AKI                                                  &    0.132 (0.1302, 0.1339) \\
Max mean urea nitrogen                                         &   0.1181 (0.1153, 0.1208) \\
Max max urea nitrogen                                          &   0.0952 (0.0934, 0.0971) \\
Mean mean urea nitrogen                                        &   0.0837 (0.0797, 0.0879) \\
Max mean urea nitrogen                                         &   0.0754 (0.0737, 0.0771) \\
Mean max potassium                                             &   0.0565 (0.0501, 0.0628) \\
Sum count abnormally high sCr                                  &   0.0501 (0.0476, 0.0526) \\
Min mean urea nitrogen                                         &   0.0257 (0.0206, 0.0307) \\
Max max sCr                                                    &   0.0198 (0.0151, 0.0243) \\
Count Dx: CKD                                                  &    0.012 (0.0085, 0.0153) \\
Mean max sCr                                                   &   0.0064 (0.0034, 0.0092) \\
Count Px: Assay of urine sodium                                &   0.0052 (0.0027, 0.0075) \\
Sum count abnormally high glomerular filtration rate-caucasian &   0.0047 (0.0036, 0.0059) \\
Mean max glucose                                               &   0.0033 (0.0005, 0.0055) \\
Count discharges with home organization care services          &   0.0027 (0.0004, 0.0045) \\
Count Px: Injection of furosemide or levetiracetam             &   0.0025 (0.0005, 0.0043) \\
Max max potassium                                              &      0.0012 (0.0, 0.0022) \\
Max count abnormally high sCr                                  &  0.0008 (-0.0001, 0.0014) \\
Sum count abnormally high glucose                              &  0.0004 (-0.0004, 0.0007) \\
\end{tabular}
\\
\\
\begin{tabular}{p{9cm}|p{3cm}}
\toprule
{RHPLR1 (-)} &               Mean (95\% CI) \\
\midrule
Mean min hemoglobin &  -0.1627 (-0.1675, -0.1577) \\
Mean min albumin    &   -0.1286 (-0.1322, -0.125) \\
Mean mean albumin   &  -0.0638 (-0.0695, -0.0582) \\
Max min hemoglobin  &   -0.0341 (-0.0404, -0.028) \\
Min min hemoglobin  &   -0.0307 (-0.036, -0.0254) \\
Mean min calcium    &   -0.005 (-0.0077, -0.0019) \\
Min min albumin     &  -0.0029 (-0.0046, -0.0009) \\
\bottomrule
\end{tabular}
\end{table}

 \section*{Acknowledgments}
The authors thank staff at the Center for Integrated Research Computing at the University of Rochester and Information Systems Division at Rochester.  The authors thank Robert White for obtaining the data and helpful discussions.  This work was partially funded by the University of Rochester Clinical and Translational Science Institute grants UL1 TR002001 and TL1 TR002000 from the National Center for Advancing Translational Sciences (NCATS) a component of the National Institutes of Health (NIH).  The content is solely the responsibility of the authors and does not necessarily represent the official views of the National Institutes of Health. 

\nolinenumbers


\begin{thebibliography}{10}

\bibitem{RIFLE2004}
Bellomo R, Ronco C, Kellum JA, Mehta RL, Palevsky P, Acute Dialysis
  Quality~Initiative w.
\newblock Acute renal failure - definition, outcome measures, animal models,
  fluid therapy and information technology needs: the Second International
  Consensus Conference of the Acute Dialysis Quality Initiative (ADQI) Group
  [Journal Article].
\newblock Crit Care. 2004;8(4):R204--12.
\newblock Bellomo, Rinaldo Ronco, Claudio Kellum, John A Mehta, Ravindra L
  Palevsky, Paul eng Consensus Development Conference Guideline Practice
  Guideline Review England 2004/08/18 05:00 Crit Care. 2004 Aug;8(4):R204-12.
  doi: 10.1186/cc2872. Epub 2004 May 24.
\newblock Available from: \url{https://www.ncbi.nlm.nih.gov/pubmed/15312219}.

\bibitem{Venkataraman2007}
Venkataraman R, Kellum JA.
\newblock Defining acute renal failure: the RIFLE criteria [Journal Article].
\newblock J Intensive Care Med. 2007;22(4):187--93.
\newblock Venkataraman, Ramesh Kellum, John A eng Review 2007/08/23 09:00 J
  Intensive Care Med. 2007 Jul-Aug;22(4):187-93. doi: 10.1177/0885066607299510.
\newblock Available from: \url{https://www.ncbi.nlm.nih.gov/pubmed/17712054}.

\bibitem{chertow2005MortLOSCost}
Chertow GM, Burdick E, Honour M, Bonventre JV, Bates DW.
\newblock Acute kidney injury, mortality, length of stay, and costs in
  hospitalized patients [Journal Article].
\newblock J Am Soc Nephrol. 2005;16(11):3365--70.
\newblock Chertow, Glenn M Burdick, Elisabeth Honour, Melissa Bonventre, Joseph
  V Bates, David W eng 2005/09/24 09:00 J Am Soc Nephrol. 2005
  Nov;16(11):3365-70. doi: 10.1681/ASN.2004090740. Epub 2005 Sep 21.
\newblock Available from: \url{https://www.ncbi.nlm.nih.gov/pubmed/16177006}.

\bibitem{bellomo2012acute}
Bellomo R, Kellum JA, Ronco C.
\newblock Acute kidney injury [Journal Article].
\newblock Lancet. 2012;380(9843):756--66.
\newblock Bellomo, Rinaldo Kellum, John A Ronco, Claudio eng Review England
  2012/05/24 06:00 Lancet. 2012 Aug 25;380(9843):756-66. doi:
  10.1016/S0140-6736(11)61454-2. Epub 2012 May 21.
\newblock Available from: \url{https://www.ncbi.nlm.nih.gov/pubmed/22617274}.

\bibitem{AKIN}
Mehta RL, Kellum JA, Shah SV, Molitoris BA, Ronco C, Warnock DG, et~al.
\newblock Acute Kidney Injury Network: report of an initiative to improve
  outcomes in acute kidney injury [Journal Article].
\newblock Crit Care. 2007;11(2):R31.
\newblock Mehta, Ravindra L Kellum, John A Shah, Sudhir V Molitoris, Bruce A
  Ronco, Claudio Warnock, David G Levin, Adeera eng Congresses Research
  Support, Non-U.S. Gov't England 2007/03/03 09:00 Crit Care. 2007;11(2):R31.
  doi: 10.1186/cc5713.
\newblock Available from: \url{https://www.ncbi.nlm.nih.gov/pubmed/17331245}.

\bibitem{KDIGO-AKIdef2012}
Group KAKIW.
\newblock Section 2: AKI Definition [Journal Article].
\newblock Kidney Int Suppl (2011). 2012;2(1):19--36.
\newblock Eng 2012/03/01 00:00 Kidney Int Suppl (2011). 2012 Mar;2(1):19-36.
  doi: 10.1038/kisup.2011.32.
\newblock Available from: \url{https://www.ncbi.nlm.nih.gov/pubmed/25018918}.

\bibitem{ERBPpositionState}
Ad-hoc working~group of E, Fliser D, Laville M, Covic A, Fouque D, Vanholder R,
  et~al.
\newblock A European Renal Best Practice (ERBP) position statement on the
  Kidney Disease Improving Global Outcomes (KDIGO) clinical practice guidelines
  on acute kidney injury: part 1: definitions, conservative management and
  contrast-induced nephropathy [Journal Article].
\newblock Nephrol Dial Transplant. 2012;27(12):4263--72.
\newblock Fliser, Danilo Laville, Maurice Covic, Adrian Fouque, Denis
  Vanholder, Raymond Juillard, Laurent Van Biesen, Wim eng Practice Guideline
  England 2012/10/10 06:00 Nephrol Dial Transplant. 2012 Dec;27(12):4263-72.
  doi: 10.1093/ndt/gfs375. Epub 2012 Oct 8.
\newblock Available from: \url{https://www.ncbi.nlm.nih.gov/pubmed/23045432}.

\bibitem{sutherland2016utilizingAKIADQI}
Sutherland SM, Chawla LS, Kane-Gill SL, Hsu RK, Kramer AA, Goldstein SL, et~al.
\newblock Utilizing electronic health records to predict acute kidney injury
  risk and outcomes: workgroup statements from the 15(th) ADQI Consensus
  Conference [Journal Article].
\newblock Can J Kidney Health Dis. 2016;3:11.
\newblock Sutherland, Scott M Chawla, Lakhmir S Kane-Gill, Sandra L Hsu,
  Raymond K Kramer, Andrew A Goldstein, Stuart L Kellum, John A Ronco, Claudio
  Bagshaw, Sean M eng Review England 2016/03/01 06:00 Can J Kidney Health Dis.
  2016 Feb 26;3:11. doi: 10.1186/s40697-016-0099-4. eCollection 2016.
\newblock Available from: \url{https://www.ncbi.nlm.nih.gov/pubmed/26925247}.

\bibitem{lameire2008prevention}
Lameire N, van Biesen W, Hoste E, Vanholder R.
\newblock The prevention of acute kidney injury an in-depth narrative review:
  Part 2: Drugs in the prevention of acute kidney injury [Journal Article].
\newblock NDT Plus. 2009;2(1):1--10.
\newblock Lameire, Norbert van Biesen, Wim Hoste, Eric Vanholder, Raymond eng
  England 2009/02/01 00:00 NDT Plus. 2009 Feb;2(1):1-10. doi:
  10.1093/ndtplus/sfn199.
\newblock Available from: \url{https://www.ncbi.nlm.nih.gov/pubmed/25949275}.

\bibitem{TepelPreventContrastRedRenalFunc}
Tepel M, van~der Giet M, Schwarzfeld C, Laufer U, Liermann D, Zidek W.
\newblock Prevention of radiographic-contrast-agent-induced reductions in renal
  function by acetylcysteine [Journal Article].
\newblock N Engl J Med. 2000;343(3):180--4.
\newblock Tepel, M van der Giet, M Schwarzfeld, C Laufer, U Liermann, D Zidek,
  W eng Clinical Trial Randomized Controlled Trial 2000/07/20 11:00 N Engl J
  Med. 2000 Jul 20;343(3):180-4. doi: 10.1056/NEJM200007203430304.
\newblock Available from: \url{https://www.ncbi.nlm.nih.gov/pubmed/10900277}.

\bibitem{SolomonRadioCon}
Solomon R, Werner C, Mann D, D'Elia J, Silva P.
\newblock Effects of saline, mannitol, and furosemide on acute decreases in
  renal function induced by radiocontrast agents [Journal Article].
\newblock N Engl J Med. 1994;331(21):1416--20.
\newblock Solomon, R Werner, C Mann, D D'Elia, J Silva, P eng Clinical Trial
  Comparative Study Randomized Controlled Trial Research Support, Non-U.S.
  Gov't 1994/11/24 00:00 N Engl J Med. 1994 Nov 24;331(21):1416-20. doi:
  10.1056/NEJM199411243312104.
\newblock Available from: \url{https://www.ncbi.nlm.nih.gov/pubmed/7969280}.

\bibitem{GuoDrugNephro}
Guo X, Nzerue C.
\newblock How to prevent, recognize, and treat drug-induced nephrotoxicity
  [Journal Article].
\newblock Cleve Clin J Med. 2002;69(4):289--90.
\newblock Available from: \url{https://www.ncbi.nlm.nih.gov/pubmed/11996200}.

\bibitem{VogelzangChemoNeph}
Vogelzang NJ.
\newblock Nephrotoxicity from chemotherapy: prevention and management [Journal
  Article].
\newblock Oncology (Williston Park). 1991;5(10):97--102, 105; disc 105,
  109--11.
\newblock Vogelzang, N J eng Review 1991/10/01 00:00 Oncology (Williston Park).
  1991 Oct;5(10):97-102, 105; disc. 105, 109-11.
\newblock Available from: \url{https://www.ncbi.nlm.nih.gov/pubmed/1838278}.

\bibitem{koyner2016development}
Koyner JL, Adhikari R, Edelson DP, Churpek MM.
\newblock Development of a Multicenter Ward-Based AKI Prediction Model [Journal
  Article].
\newblock Clin J Am Soc Nephrol. 2016;11(11):1935--1943.
\newblock Koyner, Jay L Adhikari, Richa Edelson, Dana P Churpek, Matthew M eng
  K08 HL121080/HL/NHLBI NIH HHS/ UL1 RR024999/RR/NCRR NIH HHS/ Multicenter
  Study Validation Studies 2016/09/17 06:00 Clin J Am Soc Nephrol. 2016 Nov
  7;11(11):1935-1943. doi: 10.2215/CJN.00280116. Epub 2016 Sep 15.
\newblock Available from: \url{https://www.ncbi.nlm.nih.gov/pubmed/27633727}.

\bibitem{calDavis}
Davis SE, Lasko TA, Chen G, Siew ED, Matheny ME.
\newblock Calibration drift in regression and machine learning models for acute
  kidney injury [Journal Article].
\newblock J Am Med Inform Assoc. 2017;24(6):1052--1061.
\newblock Davis, Sharon E Lasko, Thomas A Chen, Guanhua Siew, Edward D Matheny,
  Michael E eng T15 LM007450/LM/NLM NIH HHS/ England 2017/04/06 06:00 J Am Med
  Inform Assoc. 2017 Nov 1;24(6):1052-1061. doi: 10.1093/jamia/ocx030.
\newblock Available from: \url{https://www.ncbi.nlm.nih.gov/pubmed/28379439}.

\bibitem{CroninVetStrat}
Cronin RM, VanHouten JP, Siew ED, Eden SK, Fihn SD, Nielson CD, et~al.
\newblock National Veterans Health Administration inpatient risk stratification
  models for hospital-acquired acute kidney injury [Journal Article].
\newblock J Am Med Inform Assoc. 2015;22(5):1054--71.
\newblock Cronin, Robert M VanHouten, Jacob P Siew, Edward D Eden, Svetlana K
  Fihn, Stephan D Nielson, Christopher D Peterson, Josh F Baker, Clifton R
  Ikizler, T Alp Speroff, Theodore Matheny, Michael E eng
  5U01DK082185-02/DK/NIDDK NIH HHS/ I01 HX000853/HX/HSRD VA/
  5U01DK082192-02/DK/NIDDK NIH HHS/ K23DK088964-01A1/DK/NIDDK NIH HHS/
  5T15LM007450-12/LM/NLM NIH HHS/ T32 GM007347/GM/NIGMS NIH HHS/ K24
  DK62849/DK/NIDDK NIH HHS/ R01 LM009965-03/LM/NLM NIH HHS/
  5U01DK082223-02/DK/NIDDK NIH HHS/ Comparative Study Research Support, N.I.H.,
  Extramural Research Support, U.S. Gov't, Non-P.H.S. England 2015/06/25 06:00
  J Am Med Inform Assoc. 2015 Sep;22(5):1054-71. doi: 10.1093/jamia/ocv051.
  Epub 2015 Jun 23.
\newblock Available from: \url{https://www.ncbi.nlm.nih.gov/pubmed/26104740}.

\bibitem{c8}
Kristovic D, Horvatic I, Husedzinovic I, Sutlic Z, Rudez I, Baric D, et~al.
\newblock Cardiac surgery-associated acute kidney injury: risk factors analysis
  and comparison of prediction models [Journal Article].
\newblock Interact Cardiovasc Thorac Surg. 2015;21(3):366--73.
\newblock Kristovic, Darko Horvatic, Ivica Husedzinovic, Ino Sutlic, Zeljko
  Rudez, Igor Baric, Davor Unic, Daniel Blazekovic, Robert Crnogorac, Matija
  eng Comparative Study England 2015/06/21 06:00 Interact Cardiovasc Thorac
  Surg. 2015 Sep;21(3):366-73. doi: 10.1093/icvts/ivv162. Epub 2015 Jun 18.
\newblock Available from: \url{https://www.ncbi.nlm.nih.gov/pubmed/26091696}.

\bibitem{c12}
Tsai TT, Patel UD, Chang TI, Kennedy KF, Masoudi FA, Matheny ME, et~al.
\newblock Validated contemporary risk model of acute kidney injury in patients
  undergoing percutaneous coronary interventions: insights from the National
  Cardiovascular Data Registry Cath-PCI Registry [Journal Article].
\newblock J Am Heart Assoc. 2014;3(6):e001380.
\newblock Tsai, Thomas T Patel, Uptal D Chang, Tara I Kennedy, Kevin F Masoudi,
  Frederick A Matheny, Michael E Kosiborod, Mikhail Amin, Amit P Weintraub,
  William S Curtis, Jeptha P Messenger, John C Rumsfeld, John S Spertus, John A
  eng I01 HX000853/HX/HSRD VA/ Research Support, Non-U.S. Gov't Research
  Support, U.S. Gov't, Non-P.H.S. Validation Studies England 2014/12/18 06:00 J
  Am Heart Assoc. 2014 Dec;3(6):e001380. doi: 10.1161/JAHA.114.001380.
\newblock Available from: \url{https://www.ncbi.nlm.nih.gov/pubmed/25516439}.

\bibitem{c15}
Gurm HS, Seth M, Kooiman J, Share D.
\newblock A novel tool for reliable and accurate prediction of renal
  complications in patients undergoing percutaneous coronary intervention
  [Journal Article].
\newblock J Am Coll Cardiol. 2013;61(22):2242--8.
\newblock Gurm, Hitinder S Seth, Milan Kooiman, Judith Share, David eng
  Research Support, N.I.H., Extramural Research Support, Non-U.S. Gov't
  Validation Studies 2013/06/01 06:00 J Am Coll Cardiol. 2013 Jun
  4;61(22):2242-8. doi: 10.1016/j.jacc.2013.03.026.
\newblock Available from: \url{https://www.ncbi.nlm.nih.gov/pubmed/23721921}.

\bibitem{c18}
Legrand M, Pirracchio R, Rosa A, Petersen ML, Van~der Laan M, Fabiani JN,
  et~al.
\newblock Incidence, risk factors and prediction of post-operative acute kidney
  injury following cardiac surgery for active infective endocarditis: an
  observational study [Journal Article].
\newblock Crit Care. 2013;17(5):R220.
\newblock Legrand, Matthieu Pirracchio, Romain Rosa, Anne Petersen, Maya L Van
  der Laan, Mark Fabiani, Jean-Noel Fernandez-gerlinger, Marie-paule Podglajen,
  Isabelle Safran, Denis Cholley, Bernard Mainardi, Jean-Luc eng Observational
  Study England 2013/10/08 06:00 Crit Care. 2013 Oct 4;17(5):R220. doi:
  10.1186/cc13041.
\newblock Available from: \url{https://www.ncbi.nlm.nih.gov/pubmed/24093498}.

\bibitem{i1}
Chawla LS, Davison DL, Brasha-Mitchell E, Koyner JL, Arthur JM, Shaw AD, et~al.
\newblock Development and standardization of a furosemide stress test to
  predict the severity of acute kidney injury [Journal Article].
\newblock Crit Care. 2013;17(5):R207.
\newblock Chawla, Lakhmir S Davison, Danielle L Brasha-Mitchell, Ermira Koyner,
  Jay L Arthur, John M Shaw, Andrew D Tumlin, James A Trevino, Sharon A Kimmel,
  Paul L Seneff, Michael G eng K23 DK081616/DK/NIDDK NIH HHS/ L30
  DK084760/DK/NIDDK NIH HHS/ K23DK081616/DK/NIDDK NIH HHS/ R01
  DK080234/DK/NIDDK NIH HHS/ Multicenter Study Research Support, N.I.H.,
  Extramural England 2013/09/24 06:00 Crit Care. 2013 Sep 20;17(5):R207. doi:
  10.1186/cc13015.
\newblock Available from: \url{https://www.ncbi.nlm.nih.gov/pubmed/24053972}.

\bibitem{i2}
Koyner JL, Davison DL, Brasha-Mitchell E, Chalikonda DM, Arthur JM, Shaw AD,
  et~al.
\newblock Furosemide Stress Test and Biomarkers for the Prediction of AKI
  Severity [Journal Article].
\newblock J Am Soc Nephrol. 2015;26(8):2023--31.
\newblock Koyner, Jay L Davison, Danielle L Brasha-Mitchell, Ermira Chalikonda,
  Divya M Arthur, John M Shaw, Andrew D Tumlin, James A Trevino, Sharon A
  Bennett, Michael R Kimmel, Paul L Seneff, Michael G Chawla, Lakhmir S eng UL1
  TR000430/TR/NCATS NIH HHS/ R01-DK080234/DK/NIDDK NIH HHS/ K23
  DK081616/DK/NIDDK NIH HHS/ K23-DK081616/DK/NIDDK NIH HHS/ R01
  DK080234/DK/NIDDK NIH HHS/ Clinical Trial Comparative Study Research Support,
  N.I.H., Extramural 2015/02/07 06:00 J Am Soc Nephrol. 2015 Aug;26(8):2023-31.
  doi: 10.1681/ASN.2014060535. Epub 2015 Feb 5.
\newblock Available from: \url{https://www.ncbi.nlm.nih.gov/pubmed/25655065}.

\bibitem{i3}
Cruz DN, Ferrer-Nadal A, Piccinni P, Goldstein SL, Chawla LS, Alessandri E,
  et~al.
\newblock Utilization of small changes in serum creatinine with clinical risk
  factors to assess the risk of AKI in critically lll adults [Journal Article].
\newblock Clin J Am Soc Nephrol. 2014;9(4):663--72.
\newblock Cruz, Dinna N Ferrer-Nadal, Asuncion Piccinni, Pasquale Goldstein,
  Stuart L Chawla, Lakhmir S Alessandri, Elisa Belluomo Anello, Clara Bohannon,
  Will Bove, Tiziana Brienza, Nicola Carlini, Mauro Forfori, Francesco
  Garzotto, Francesco Gramaticopolo, Silvia Iannuzzi, Michele Montini, Luca
  Pelaia, Paolo Ronco, Claudio eng Multicenter Study Observational Study
  Research Support, Non-U.S. Gov't 2014/03/29 06:00 Clin J Am Soc Nephrol. 2014
  Apr;9(4):663-72. doi: 10.2215/CJN.05190513. Epub 2014 Mar 27.
\newblock Available from: \url{https://www.ncbi.nlm.nih.gov/pubmed/24677553}.

\bibitem{i4}
Forni LG, Dawes T, Sinclair H, Cheek E, Bewick V, Dennis M, et~al.
\newblock Identifying the patient at risk of acute kidney injury: a predictive
  scoring system for the development of acute kidney injury in acute medical
  patients [Journal Article].
\newblock Nephron Clin Pract. 2013;123(3-4):143--50.
\newblock Forni, Lui G Dawes, Thomas Sinclair, Hamish Cheek, Elizabeth Bewick,
  Vivien Dennis, Mark Venn, Richard eng Switzerland 2013/07/28 06:00 Nephron
  Clin Pract. 2013;123(3-4):143-50. doi: 10.1159/000351509. Epub 2013 Jul 25.
\newblock Available from: \url{https://www.ncbi.nlm.nih.gov/pubmed/23887252}.

\bibitem{i5}
Kane-Gill SL, Sileanu FE, Murugan R, Trietley GS, Handler SM, Kellum JA.
\newblock Risk factors for acute kidney injury in older adults with critical
  illness: a retrospective cohort study [Journal Article].
\newblock Am J Kidney Dis. 2015;65(6):860--9.
\newblock Kane-Gill, Sandra L Sileanu, Florentina E Murugan, Raghavan Trietley,
  Gregory S Handler, Steven M Kellum, John A eng P30 AG024827/AG/NIA NIH HHS/
  R01DK083961/DK/NIDDK NIH HHS/ R01DK070910/DK/NIDDK NIH HHS/ R01
  HS018721/HS/AHRQ HHS/ KL2 RR024154/RR/NCRR NIH HHS/ R01 DK083961/DK/NIDDK NIH
  HHS/ R01 DK070910/DK/NIDDK NIH HHS/ R01HS018721/HS/AHRQ HHS/ Research
  Support, N.I.H., Extramural 2014/12/10 06:00 Am J Kidney Dis. 2015
  Jun;65(6):860-9. doi: 10.1053/j.ajkd.2014.10.018. Epub 2014 Dec 6.
\newblock Available from: \url{https://www.ncbi.nlm.nih.gov/pubmed/25488106}.

\bibitem{i6}
Flechet M, Guiza F, Schetz M, Wouters P, Vanhorebeek I, Derese I, et~al.
\newblock AKIpredictor, an online prognostic calculator for acute kidney injury
  in adult critically ill patients: development, validation and comparison to
  serum neutrophil gelatinase-associated lipocalin [Journal Article].
\newblock Intensive Care Med. 2017;43(6):764--773.
\newblock Flechet, Marine Guiza, Fabian Schetz, Miet Wouters, Pieter
  Vanhorebeek, Ilse Derese, Inge Gunst, Jan Spriet, Isabel Casaer, Michael Van
  den Berghe, Greet Meyfroidt, Geert eng Comparative Study Validation Studies
  2017/01/29 06:00 Intensive Care Med. 2017 Jun;43(6):764-773. doi:
  10.1007/s00134-017-4678-3. Epub 2017 Jan 27.
\newblock Available from: \url{https://www.ncbi.nlm.nih.gov/pubmed/28130688}.

\bibitem{Kate2016}
Kate RJ, Perez RM, Mazumdar D, Pasupathy KS, Nilakantan V.
\newblock Prediction and detection models for acute kidney injury in
  hospitalized older adults [Journal Article].
\newblock BMC Med Inform Decis Mak. 2016;16(1):39.
\newblock Kate, Rohit J Perez, Ruth M Mazumdar, Debesh Pasupathy, Kalyan S
  Nilakantan, Vani eng England 2016/03/31 06:00 BMC Med Inform Decis Mak. 2016
  Mar 29;16:39. doi: 10.1186/s12911-016-0277-4.
\newblock Available from: \url{https://www.ncbi.nlm.nih.gov/pubmed/27025458}.

\bibitem{livt2}
Xu X, Ling Q, Wei Q, Wu J, Gao F, He ZL, et~al.
\newblock An effective model for predicting acute kidney injury after liver
  transplantation [Journal Article].
\newblock Hepatobiliary Pancreat Dis Int. 2010;9(3):259--63.
\newblock Xu, Xiao Ling, Qi Wei, Qiang Wu, Jian Gao, Feng He, Zeng-Lei Zhou,
  Lin Zheng, Shu-Sen eng Research Support, Non-U.S. Gov't Validation Studies
  Singapore 2010/06/09 06:00 Hepatobiliary Pancreat Dis Int. 2010
  Jun;9(3):259-63.
\newblock Available from: \url{https://www.ncbi.nlm.nih.gov/pubmed/20525552}.

\bibitem{lungT1}
Grimm JC, Lui C, Kilic A, Valero r V, Sciortino CM, Whitman GJ, et~al.
\newblock A risk score to predict acute renal failure in adult patients after
  lung transplantation [Journal Article].
\newblock Ann Thorac Surg. 2015;99(1):251--7.
\newblock Grimm, Joshua C Lui, Cecillia Kilic, Arman Valero, Vicente 3rd
  Sciortino, Christopher M Whitman, Glenn J R Shah, Ashish S eng Netherlands
  2014/12/03 06:00 Ann Thorac Surg. 2015 Jan;99(1):251-7. doi:
  10.1016/j.athoracsur.2014.07.073. Epub 2014 Nov 14.
\newblock Available from: \url{https://www.ncbi.nlm.nih.gov/pubmed/25440281}.

\bibitem{R1}
McMahon GM, Zeng X, Waikar SS.
\newblock A risk prediction score for kidney failure or mortality in
  rhabdomyolysis [Journal Article].
\newblock JAMA Intern Med. 2013;173(19):1821--8.
\newblock McMahon, Gearoid M Zeng, Xiaoxi Waikar, Sushrut S eng R01
  DK093574/DK/NIDDK NIH HHS/ U01 DK085660/DK/NIDDK NIH HHS/ DK085660/DK/NIDDK
  NIH HHS/ DK093574/DK/NIDDK NIH HHS/ Research Support, N.I.H., Extramural
  Research Support, Non-U.S. Gov't 2013/09/04 06:00 JAMA Intern Med. 2013 Oct
  28;173(19):1821-8. doi: 10.1001/jamainternmed.2013.9774.
\newblock Available from: \url{https://www.ncbi.nlm.nih.gov/pubmed/24000014}.

\bibitem{R2}
Rodriguez E, Soler MJ, Rap O, Barrios C, Orfila MA, Pascual J.
\newblock Risk factors for acute kidney injury in severe rhabdomyolysis
  [Journal Article].
\newblock PLoS One. 2013;8(12):e82992.
\newblock Rodriguez, Eva Soler, Maria J Rap, Oana Barrios, Clara Orfila, Maria
  A Pascual, Julio eng 2013/12/25 06:00 PLoS One. 2013 Dec 18;8(12):e82992.
  doi: 10.1371/journal.pone.0082992. eCollection 2013.
\newblock Available from: \url{https://www.ncbi.nlm.nih.gov/pubmed/24367578}.

\bibitem{Breiman_alg_v_mod}
Breiman L, et~al.
\newblock Statistical modeling: The two cultures (with comments and a rejoinder
  by the author) [Journal Article].
\newblock Statistical science. 2001;16(3):199--231.

\bibitem{friedman2001greedy}
Friedman JH.
\newblock Greedy function approximation: a gradient boosting machine [Journal
  Article].
\newblock Annals of statistics. 2001;p. 1189--1232.

\bibitem{hochreiter1997long}
Hochreiter S, Schmidhuber J.
\newblock Long short-term memory [Journal Article].
\newblock Neural Comput. 1997;9(8):1735--80.
\newblock Hochreiter, S Schmidhuber, J eng Research Support, Non-U.S. Gov't
  1997/10/23 00:00 Neural Comput. 1997 Nov 15;9(8):1735-80.
\newblock Available from: \url{https://www.ncbi.nlm.nih.gov/pubmed/9377276}.

\bibitem{tibshirani1996regression}
Tibshirani R.
\newblock Regression shrinkage and selection via the lasso [Journal Article].
\newblock Journal of the Royal Statistical Society Series B (Methodological).
  1996;p. 267--288.

\bibitem{waikar2006validity}
Waikar SS, Wald R, Chertow GM, Curhan GC, Winkelmayer WC, Liangos O, et~al.
\newblock Validity of International Classification of Diseases, Ninth Revision,
  Clinical Modification Codes for Acute Renal Failure [Journal Article].
\newblock J Am Soc Nephrol. 2006;17(6):1688--94.
\newblock Waikar, Sushrut S Wald, Ron Chertow, Glenn M Curhan, Gary C
  Winkelmayer, Wolfgang C Liangos, Orfeas Sosa, Marie-Anne Jaber, Bertrand L
  eng K23 DK065102-02/DK/NIDDK NIH HHS/ R33 DK067645/DK/NIDDK NIH HHS/ T32
  DK007791/DK/NIDDK NIH HHS/ Multicenter Study Research Support, N.I.H.,
  Extramural Research Support, Non-U.S. Gov't Research Support, U.S. Gov't,
  Non-P.H.S. 2006/04/28 09:00 J Am Soc Nephrol. 2006 Jun;17(6):1688-94. doi:
  10.1681/ASN.2006010073. Epub 2006 Apr 26.
\newblock Available from: \url{https://www.ncbi.nlm.nih.gov/pubmed/16641149}.

\bibitem{lipton2016directly}
Lipton ZC, Kale D, Wetzel R.
\newblock Directly modeling missing data in sequences with RNNs: Improved
  classification of clinical time series.
\newblock Machine Learning for Healthcare Conference;. p. 253--270.

\bibitem{singhLeveHier}
Singh A, Nadkarni G, Guttag J, Bottinger E.
\newblock Leveraging hierarchy in medical codes for predictive modeling.
\newblock Proceedings of the 5th ACM Conference on Bioinformatics,
  Computational Biology, and Health Informatics. ACM;. p. 96--103.

\bibitem{goldstein2017comparison}
Goldstein BA, Pomann GM, Winkelmayer WC, Pencina MJ.
\newblock A comparison of risk prediction methods using repeated observations:
  an application to electronic health records for hemodialysis [Journal
  Article].
\newblock Statistics in Medicine. 2017;.

\bibitem{choiHF}
Choi E, Schuetz A, Stewart WF, Sun J.
\newblock Using recurrent neural network models for early detection of heart
  failure onset [Journal Article].
\newblock J Am Med Inform Assoc. 2017;24(2):361--370.
\newblock Choi, Edward Schuetz, Andy Stewart, Walter F Sun, Jimeng eng R01
  HL116832/HL/NHLBI NIH HHS/ England 2016/08/16 06:00 J Am Med Inform Assoc.
  2017 Mar 1;24(2):361-370. doi: 10.1093/jamia/ocw112.
\newblock Available from: \url{https://www.ncbi.nlm.nih.gov/pubmed/27521897}.

\bibitem{colopy2017bayesian}
Colopy GW, Roberts S, Clifton D.
\newblock Bayesian Optimisation of Personalised Models for Patient Vital-Sign
  Monitoring [Journal Article].
\newblock IEEE J Biomed Health Inform. 2017;Colopy, Glen Wright Roberts,
  Stephen Clifton, David eng 2018/07/11 06:00 IEEE J Biomed Health Inform. 2017
  Dec 19. doi: 10.1109/JBHI.2017.2751509.
\newblock Available from: \url{https://www.ncbi.nlm.nih.gov/pubmed/29990047}.

\bibitem{alaa2016hidden}
Alaa AM, van~der Schaar M.
\newblock A Hidden Absorbing Semi-Markov Model for Informatively Censored
  Temporal Data: Learning and Inference [Journal Article].
\newblock arXiv preprint arXiv:161206007. 2016;.

\bibitem{cawley2010over}
Cawley GC, Talbot NL.
\newblock On over-fitting in model selection and subsequent selection bias in
  performance evaluation [Journal Article].
\newblock Journal of Machine Learning Research. 2010;11(Jul):2079--2107.

\bibitem{varma2006bias}
Varma S, Simon R.
\newblock Bias in error estimation when using cross-validation for model
  selection [Journal Article].
\newblock BMC bioinformatics. 2006;7(1):91.

\bibitem{bergstra2012random}
Bergstra J, Bengio Y.
\newblock Random search for hyper-parameter optimization [Journal Article].
\newblock Journal of Machine Learning Research. 2012;13(Feb):281--305.

\bibitem{hoerl1970ridge}
Hoerl AE, Kennard RW.
\newblock Ridge regression: Biased estimation for nonorthogonal problems
  [Journal Article].
\newblock Technometrics. 1970;12(1):55--67.

\bibitem{Breiman_RF}
Breiman L.
\newblock Random forests [Journal Article].
\newblock Machine Learning. 2001;45(1):5--32.
\newblock Available from: \url{<Go to ISI>://WOS:000170489900001
  https://doi.org/Doi 10.1023/A:1010933404324}.

\bibitem{rumelhart1985learning}
Rumelhart DE, Hinton GE, Williams RJ.
\newblock Learning internal representations by error propagation.
\newblock California Univ San Diego La Jolla Inst for Cognitive Science; 1985.

\bibitem{harrell2015regression}
Harrell F.
\newblock Regression modeling strategies: with applications to linear models,
  logistic and ordinal regression, and survival analysis.
\newblock Springer; 2015.

\bibitem{Platt}
Platt J, et~al.
\newblock Probabilistic outputs for support vector machines and comparisons to
  regularized likelihood methods [Journal Article].
\newblock Advances in large margin classifiers. 1999;10(3):61--74.

\bibitem{stabselect}
Meinshausen N, Bühlmann P.
\newblock Stability selection [Journal Article].
\newblock Journal of the Royal Statistical Society: Series B (Statistical
  Methodology). 2010;72(4):417--473.
\newblock Available from:
  \url{https://rss.onlinelibrary.wiley.com/doi/abs/10.1111/j.1467-9868.2010.00740.x
  https://rss.onlinelibrary.wiley.com/doi/pdf/10.1111/j.1467-9868.2010.00740.x
  https://doi.org/10.1111/j.1467-9868.2010.00740.x}.

\bibitem{LeblancRiskFact}
Leblanc M, Kellum JA, Gibney RT, Lieberthal W, Tumlin J, Mehta R.
\newblock Risk factors for acute renal failure: inherent and modifiable risks
  [Journal Article].
\newblock Curr Opin Crit Care. 2005;11(6):533--6.
\newblock Leblanc, Martine Kellum, John A Gibney, R T Noel Lieberthal, Wilfred
  Tumlin, James Mehta, Ravindra eng Review 2005/11/18 09:00 Curr Opin Crit
  Care. 2005 Dec;11(6):533-6.
\newblock Available from: \url{https://www.ncbi.nlm.nih.gov/pubmed/16292055}.

\bibitem{caruana2015intelligible}
Caruana R, Lou Y, Gehrke J, Koch P, Sturm M, Elhadad N.
\newblock Intelligible models for healthcare: Predicting pneumonia risk and
  hospital 30-day readmission.
\newblock Proceedings of the 21th ACM SIGKDD International Conference on
  Knowledge Discovery and Data Mining. ACM;. p. 1721--1730.

\bibitem{KCraw}
Crawford K.
\newblock The Trouble with Bias.
\newblock NIPS 2017, Long Beach, CA.;. Available from:
  \url{https://nips.cc/Conferences/2017/Schedule?showEvent=8742}.

\bibitem{corbett2017algorithmic}
Corbett-Davies S, Pierson E, Feller A, Goel S, Huq A.
\newblock Algorithmic decision making and the cost of fairness [Journal
  Article].
\newblock arXiv preprint arXiv:170108230. 2017;.

\bibitem{pedregosa2011scikit}
Pedregosa F, Varoquaux G, Gramfort A, Michel V, Thirion B, Grisel O, et~al.
\newblock Scikit-learn: Machine Learning in Python [Journal Article].
\newblock Journal of Machine Learning Research. 2011;12(Oct):2825--2830.
\newblock Available from: \url{<Go to ISI>://WOS:000298103200003}.

\bibitem{scipy}
Jones E, Oliphant T, Peterson P, et~al.. SciPy: Open source scientific tools
  for Python [Generic]; 2001.
\newblock Available from: \url{http://www.scipy.org/}.

\bibitem{mckinney2010pandas}
McKinney W.
\newblock Data structures for statistical computing in python.
\newblock vol. 445 of Proceedings of the 9th Python in Science Conference. van
  der Voort S, Millman J;. p. 51--56.

\bibitem{perez2007ipython}
Pérez F, Granger BE.
\newblock IPython: a system for interactive scientific computing [Journal
  Article].
\newblock Computing in Science and Engineering. 2007;9(3).

\bibitem{Hunter:2007}
Hunter JD.
\newblock Matplotlib: A 2D Graphics Environment [Journal Article].
\newblock Computing in Science and Engineering. 2007;9(3):90--95.
\newblock Available from: \url{https://doi.org/10.1109/MCSE.2007.55}.

\bibitem{behnel2010cython}
Behnel S, Bradshaw R, Citro C, Dalcin L, Seljebotn DS, Smith K.
\newblock Cython: The Best of Both Worlds [Journal Article].
\newblock Computing in Science and Engineering. 2011;13(2):31--39.
\newblock Available from: \url{https://doi.org/10.1109/MCSE.2010.118}.

\bibitem{walt2011numpy}
Walt Svd, Colbert SC, Varoquaux G.
\newblock The NumPy array: a structure for efficient numerical computation
  [Journal Article].
\newblock Computing in Science and Engineering. 2011;13(2):22--30.

\bibitem{ShustermanHospAcq}
Shusterman N, Strom BL, Murray TG, Morrison G, West SL, Maislin G.
\newblock Risk factors and outcome of hospital-acquired acute renal failure.
  Clinical epidemiologic study [Journal Article].
\newblock Am J Med. 1987;83(1):65--71.
\newblock Shusterman, N Strom, B L Murray, T G Morrison, G West, S L Maislin, G
  eng 5T32-AM-07006/AM/NIADDK NIH HHS/ Comparative Study Research Support,
  Non-U.S. Gov't Research Support, U.S. Gov't, P.H.S. 1987/07/01 00:00 Am J
  Med. 1987 Jul;83(1):65-71.
\newblock Available from: \url{https://www.ncbi.nlm.nih.gov/pubmed/3605183}.

\bibitem{chawla2014acute}
Chawla LS, Eggers PW, Star RA, Kimmel PL.
\newblock Acute kidney injury and chronic kidney disease as interconnected
  syndromes [Journal Article].
\newblock New England Journal of Medicine. 2014;371(1):58--66.

\bibitem{ronco2008cardiorenal}
Ronco C, Haapio M, House AA, Anavekar N, Bellomo R.
\newblock Cardiorenal syndrome [Journal Article].
\newblock Journal of the American College of Cardiology.
  2008;52(19):1527--1539.

\bibitem{garcia2008acute}
Garcia-Tsao G, Parikh CR, Viola A.
\newblock Acute kidney injury in cirrhosis [Journal Article].
\newblock Hepatology. 2008;48(6):2064--77.
\newblock Garcia-Tsao, Guadalupe Parikh, Chirag R Viola, Antonella eng Review
  2008/11/13 09:00 Hepatology. 2008 Dec;48(6):2064-77. doi: 10.1002/hep.22605.
\newblock Available from: \url{https://www.ncbi.nlm.nih.gov/pubmed/19003880}.

\bibitem{fede2012renal}
Fede G, D'Amico G, Arvaniti V, Tsochatzis E, Germani G, Georgiadis D, et~al.
\newblock Renal failure and cirrhosis: a systematic review of mortality and
  prognosis [Journal Article].
\newblock J Hepatol. 2012;56(4):810--8.
\newblock Fede, Giuseppe D'Amico, Gennaro Arvaniti, Vasiliki Tsochatzis,
  Emmanuel Germani, Giacomo Georgiadis, Dimosthenis Morabito, Alberto
  Burroughs, Andrew Kenneth eng Comparative Study Meta-Analysis Review
  Netherlands 2011/12/17 06:00 J Hepatol. 2012 Apr;56(4):810-8. doi:
  10.1016/j.jhep.2011.10.016. Epub 2011 Dec 13.
\newblock Available from: \url{https://www.ncbi.nlm.nih.gov/pubmed/22173162}.

\bibitem{BayesianTimeForBenavoli}
Benavoli A, Corani G, Demsar J, Zaffalon M.
\newblock Time for a change: a tutorial for comparing multiple classifiers
  through Bayesian analysis [Journal Article].
\newblock ArXiv e-prints. 2016;Available from:
  \url{http://arxiv.org/abs/1606.04316}.

\bibitem{PalevskyCommentOnKDIGONephConsult}
Palevsky PM, Liu KD, Brophy PD, Chawla LS, Parikh CR, Thakar CV, et~al.
\newblock KDOQI US commentary on the 2012 KDIGO clinical practice guideline for
  acute kidney injury [Journal Article].
\newblock Am J Kidney Dis. 2013;61(5):649--72.
\newblock Palevsky, Paul M Liu, Kathleen D Brophy, Patrick D Chawla, Lakhmir S
  Parikh, Chirag R Thakar, Charuhas V Tolwani, Ashita J Waikar, Sushrut S
  Weisbord, Steven D eng Research Support, Non-U.S. Gov't Review 2013/03/19
  06:00 Am J Kidney Dis. 2013 May;61(5):649-72. doi:
  10.1053/j.ajkd.2013.02.349. Epub 2013 Mar 15.
\newblock Available from: \url{https://www.ncbi.nlm.nih.gov/pubmed/23499048}.

\bibitem{Futoma_comp_readmis}
Futoma J, Morris J, Lucas J.
\newblock A comparison of models for predicting early hospital readmissions
  [Journal Article].
\newblock J Biomed Inform. 2015;56:229--38.
\newblock Futoma, Joseph Morris, Jonathan Lucas, Joseph eng Comparative Study
  Research Support, U.S. Gov't, Non-P.H.S. 2015/06/06 06:00 J Biomed Inform.
  2015 Aug;56:229-38. doi: 10.1016/j.jbi.2015.05.016. Epub 2015 Jun 1.
\newblock Available from: \url{https://www.ncbi.nlm.nih.gov/pubmed/26044081}.

\bibitem{efron1992bootstrap}
Efron B.
\newblock In: Bootstrap methods: another look at the jackknife. Springer; 1992.
  p. 569--593.

\bibitem{bootstrapped}
Bootstrapped [Generic];.
\newblock Available from:
  \url{https://github.com/facebookincubator/bootstrapped}.

\bibitem{breiman2017classification}
Breiman L.
\newblock Classification and regression trees.
\newblock Routledge; 2017.

\bibitem{icd9com}
ICD9Data.com [Generic];.
\newblock Available from: \url{http://www.icd9data.com}.

\bibitem{ERvisit2014}
Ryan K, Levit K, Davis PH.
\newblock Characteristics of Weekday and Weekend Hospital Admissions.
\newblock Agency for Healthcare Research and Quality; 2010.
\newblock Available from:
  \url{http://www.hcupus.ahrq.gov/reports/statbriefs/sb87.pdf}.

\bibitem{naesens2009calcineurin}
Naesens M, Kuypers DR, Sarwal M.
\newblock Calcineurin inhibitor nephrotoxicity [Journal Article].
\newblock Clin J Am Soc Nephrol. 2009;4(2):481--508.
\newblock Naesens, Maarten Kuypers, Dirk R J Sarwal, Minnie eng Review
  2009/02/17 09:00 Clin J Am Soc Nephrol. 2009 Feb;4(2):481-508. doi:
  10.2215/CJN.04800908.
\newblock Available from: \url{https://www.ncbi.nlm.nih.gov/pubmed/19218475}.

\bibitem{leite2015renal}
Leite TT, Macedo E, Martins~Ida S, Neves FM, Liborio AB.
\newblock Renal Outcomes in Critically Ill Patients Receiving Propofol or
  Midazolam [Journal Article].
\newblock Clin J Am Soc Nephrol. 2015;10(11):1937--45.
\newblock Leite, Tacyano Tavares Macedo, Etienne Martins, Izanio da Silva
  Neves, Fernanda Macedo de Oliveira Liborio, Alexandre Braga eng 2015/09/06
  06:00 Clin J Am Soc Nephrol. 2015 Nov 6;10(11):1937-45. doi:
  10.2215/CJN.02330315. Epub 2015 Sep 4.
\newblock Available from: \url{https://www.ncbi.nlm.nih.gov/pubmed/26342046}.

\bibitem{atici2004opioid}
Atici S, Cinel L, Cinel I, Doruk N, Aktekin M, Akca A, et~al.
\newblock Opioid neurotoxicity: comparison of morphine and tramadol in an
  experimental rat model [Journal Article].
\newblock Int J Neurosci. 2004;114(8):1001--11.
\newblock Atici, Sebnem Cinel, Leyla Cinel, Ismail Doruk, Nurcan Aktekin,
  Mustafa Akca, Almila Camdeviren, Handan Oral, Ugur eng Comparative Study
  England 2004/11/06 09:00 Int J Neurosci. 2004 Aug;114(8):1001-11.
\newblock Available from: \url{https://www.ncbi.nlm.nih.gov/pubmed/15527204}.

\bibitem{shmueli2010explain}
Shmueli G, et~al.
\newblock To explain or to predict? [Journal Article].
\newblock Statistical science. 2010;25(3):289--310.

\bibitem{callahan2018machine}
Callahan A, Shah NH.
\newblock In: Machine Learning in Healthcare. Elsevier; 2018. p. 279--291.

\bibitem{rajkomar2018scalable}
Rajkomar A, Oren E, Chen K, Dai AM, Hajaj N, Hardt M, et~al.
\newblock Scalable and accurate deep learning with electronic health records
  [Journal Article].
\newblock npj Digital Medicine. 2018;1(1):18.

\bibitem{lipton2015learning}
Lipton ZC, Kale DC, Elkan C, Wetzell R.
\newblock Learning to diagnose with LSTM recurrent neural networks [Journal
  Article].
\newblock arXiv preprint arXiv:151103677. 2015;.

\bibitem{futoma2017learning}
Futoma J, Hariharan S, Heller K.
\newblock Learning to detect sepsis with a multitask gaussian process rnn
  classifier [Journal Article].
\newblock arXiv preprint arXiv:170604152. 2017;.

\bibitem{yoon2018deep}
Yoon J, Zame WR, van~der Schaar M.
\newblock Deep Sensing: Active Sensing using Multi-directional Recurrent Neural
  Networks.
\newblock International Conference on Learning Representations;. Available
  from: \url{https://openreview.net/forum?id=r1SnX5xCb}.

\bibitem{BengioReg}
Jo J, Bengio Y.
\newblock Measuring the tendency of CNNs to Learn Surface Statistical
  Regularities [Journal Article].
\newblock arXiv preprint arXiv:171111561. 2017;.

\bibitem{doi:10.1001/jama.2017.19198}
Verghese A, Shah NH, Harrington RA.
\newblock What This Computer Needs Is a Physician: Humanism and Artificial
  Intelligence [Journal Article].
\newblock JAMA. 2018;319(1):19--20.
\newblock Verghese, Abraham Shah, Nigam H Harrington, Robert A eng 2017/12/21
  06:00 JAMA. 2018 Jan 2;319(1):19-20. doi: 10.1001/jama.2017.19198.
\newblock Available from: \url{https://www.ncbi.nlm.nih.gov/pubmed/29261830}.

\bibitem{christopher2016pattern}
Christopher MB.
\newblock Pattern Recognition and Machine Learning.
\newblock Springer-Verlag New York; 2016.

\bibitem{mullainathan2017does}
Mullainathan S, Obermeyer Z.
\newblock Does machine learning automate moral hazard and error?
\newblock American Economic Review. 2017;107(5):476--80.

\bibitem{choi2016multi}
Choi E, Bahadori MT, Searles E, Coffey C, Thompson M, Bost J, et~al.
\newblock Multi-layer representation learning for medical concepts.
\newblock Proceedings of the 22nd ACM SIGKDD International Conference on
  Knowledge Discovery and Data Mining. ACM;. p. 1495--1504.

\bibitem{gbcSK}
Scikit-learn Gradient Boosting Classifier [Generic];.
\newblock Available from:
  \url{http://scikit-learn.org/stable/modules/generated/sklearn.ensemble.GradientBoostingClassifier.html}.

\bibitem{LR1SK}
Scikit-learn Logistic Regression [Generic];.
\newblock Available from:
  \url{http://scikit-learn.org/stable/modules/generated/sklearn.linear_model.LogisticRegression.html}.

\bibitem{LassoSK}
Scikit-learn Lasso [Generic];.
\newblock Available from:
  \url{http://scikit-learn.org/stable/modules/generated/sklearn.linear_model.Lasso.html}.

\bibitem{rlr}
Scikit-learn Randomized Logistic Regression [Generic];.
\newblock Available from:
  \url{http://lijiancheng0614.github.io/scikit-learn/modules/generated/sklearn.linear_model.RandomizedLogisticRegression.html}.

\bibitem{keras}
Keras LSTM [Generic];.
\newblock Available from: \url{https://keras.io/layers/recurrent/}.

\end{thebibliography}
\end{document}